\documentclass[conference]{IEEEtran}
\IEEEoverridecommandlockouts
\usepackage{cite}
\usepackage{amsmath,amssymb,amsfonts}
\usepackage{algorithmic}
\usepackage{graphicx}
\usepackage{textcomp}
\usepackage{xcolor}

\usepackage{hyperref}
\usepackage[ruled, vlined]{algorithm2e}
\SetKwComment{Comment}{$\triangleright$\ }{}
\usepackage{diagbox}
\usepackage{slashbox}
\usepackage{multirow}
\usepackage[caption=false]{subfig}
\usepackage{xcolor}
\usepackage{subfiles}
\usepackage{amsmath}
\usepackage{graphicx}

\usepackage{amsthm}
\usepackage{pifont}
\theoremstyle{definition}

\newcommand{\rev}[1]{\textcolor{black}{#1}}

\newcommand{\mb}[1]{\mathbf{#1}}
\newcommand{\tb}[1]{\textbf{#1}}

\newcommand{\D}{\mathcal{D}}

\newcommand{\del}[1]{}
\newcommand{\norm}[1]{\left\lVert#1\right\rVert}

\newcommand{\HBS}[1]{{\textcolor{black}{ #1}}}
\newcommand{\QB}[1]{{\textcolor{black}{#1}}}

\def\BibTeX{{\rm B\kern-.05em{\sc i\kern-.025em b}\kern-.08em
    T\kern-.1667em\lower.7ex\hbox{E}\kern-.125emX}}
\begin{document}

\title{Federated Learning on Non-IID Data Silos: An Experimental Study
}

\author{
\IEEEauthorblockN{Qinbin Li$^*$}
\IEEEauthorblockA{
\textit{National University of Singapore}\\
Singapore\\
qinbin@comp.nus.edu.sg
}
\and
\IEEEauthorblockN{Yiqun Diao$^*$ \hspace{2em} Quan Chen}
\IEEEauthorblockA{
\textit{Shanghai Jiao Tong University}\\
Shanghai, China \\
\{diaoyiqun, chen-quan\}@sjtu.edu.cn
}
\and
\IEEEauthorblockN{Bingsheng He}
\IEEEauthorblockA{
\textit{National University of Singapore}\\
Singapore\\
hebs@comp.nus.edu.sg
}
\thanks{$^*$Equal contribution.}
}

\maketitle

\begin{abstract}
Due to the increasing privacy concerns and data regulations, training data have been increasingly fragmented, forming distributed databases of multiple ``data silos'' (e.g., within different organizations and countries). To develop effective machine learning services, there is a must to exploit data from such distributed databases without exchanging the raw data. Recently, federated learning (FL) has been a solution with growing interests, which enables multiple parties to collaboratively train a machine learning model without exchanging their local data. A key and common challenge on distributed databases is the heterogeneity of the data distribution among the parties. The data of different parties are usually non-independently and identically distributed (i.e., non-IID). There have been many FL algorithms to address the learning effectiveness under non-IID data settings. However, there lacks an experimental study on systematically understanding their advantages and disadvantages, as previous studies have very rigid data partitioning strategies among parties, which are hardly representative and thorough. In this paper, to help researchers better understand and study the non-IID data setting in federated learning, we propose comprehensive data partitioning strategies to cover the typical non-IID data cases. Moreover, we conduct extensive experiments to evaluate state-of-the-art FL algorithms. We find that non-IID does bring significant challenges in learning accuracy of FL algorithms, and none of the existing state-of-the-art FL algorithms outperforms others in all cases. Our experiments provide insights for future studies of addressing the challenges in ``data silos''. 
\end{abstract}

\section{Introduction}
\label{sec:intro}


\HBS{In recent years, we have witnessed some promising advancement with leveraging machine learning services, such as learned index structures~\cite{10.14778/3421424.3421425, 10.1145/3318464.3389711} and learned cost estimation~\cite{10.14778/3342263.3342644, 10.1145/3318464.3389741}.} As such, machine learning services have become emerging data-intensive workloads, such as Ease.ml \cite{10.1145/3187009.3177737}, Machine Learning Bazaar~\cite{10.1145/3318464.3386146} and Rafiki \cite{10.14778/3282495.3282499}. Despite the success of machine learning services, their effectiveness highly relies on large-volume high-quality training data. However, due to the increasing privacy concerns and data regulations such as GDPR~\cite{voigt2017eu}, training data have been increasingly fragmented, forming distributed databases of multiple ``data silos'' (e.g., within different organizations and countries). Due to the deployed data regulations, raw data are usually not allowed to transfer across organizations/countries. For example, a multinational corporation (MNC) provides services to users in multiple nations, whose personal data usually cannot be centralized to a single country due to the data regulations in many countries. 

To develop effective machine learning services, it is necessary to exploit data from such distributed databases without exchanging the raw data. While there are many studies working on privacy-preserving data management and data mining \cite{agrawal2000privacy,hynes2018demonstration,niu2017trading,rizvi2002maintaining,gdprdatabase} in a centralized setting, they cannot handle the cases of distributed databases. Thus, how to conduct data mining/machine learning from distributed databases without exchanging local data has become an emerging topic.

\QB{To address the above challenge, we borrow the federated learning (FL) \cite{kairouz2019advances,li2019flsurvey,litian2019survey,yang2019federated} approach from the machine learning community.} Originally proposed by Google, FL is a promising solution to enable many parties jointly train a machine learning model while keeping their local data decentralized. \rev{Here we focus on horizontal federated learning, where the parties share the same feature space but different sample space.} Instead of exchanging data and conducting centralized training, each party sends its model to the server, which updates and sends back the global model to the parties in each round. Since their raw data are not exposed, FL is an effective way to address privacy concerns. It has attracted many research interests \cite{li2020practical,dai2020federated,he2020group,lifedprox,karimireddy2019scaffold,liu2020secure,wu2020privacy} and been widely used in practice \cite{bonawitz2019towards,hard2018federated,kaissis2020secure}. Thus, we consider FL to develop machine learning services for distributed databases. 

One key and common data challenge in such distributed databases is that data distributions in different parties are usually non-independently and identically distributed (non-IID). For example, different areas can have very different disease distributions. Due to the ozone hole, the countries in the Southern Hemisphere may have more skin cancer patients than the Northern Hemisphere. Then, the label distributions differ across parties. Another example is that people have different writing styles even for the same world. In such a case, the feature distributions differ across parties. According to previous studies \cite{karimireddy2019scaffold,Li2020On,hsu2019measuring}, the non-IID data settings can degrade the effectiveness of machine learning services. 

There have been some studies trying to develop effective FL algorithms under non-IID data including FedProx \cite{lifedprox}, SCAFFOLD \cite{karimireddy2019scaffold}, and FedNova \cite{wang2020tackling}. However, there lacks an experimental study on systematically understanding their advantages and disadvantages, as the previous studies have very rigid data partitioning strategies among parties, which are hardly representative and thorough. In the experiments of these studies, they only try one or two partitioning strategies to simulate the non-IID data setting, which does not sufficiently cover different non-IID cases.
For example, in FedAvg~\cite{mcmahan2016communication}, each party only has samples of two classes. In FedNova \cite{wang2020tackling}, the number of samples of each class in each party follows Dirichlet distribution. The above partitioning strategies only cover the label skewed case. 
Thus, it is a necessity to evaluate those algorithms with a systematic exploration of different non-IID scenarios.  

In this paper, we break the barrier of experiments on non-IID data distribution challenges in FL by proposing NIID-Bench. Specifically, we introduce six non-IID data partitioning strategies which thoroughly consider different cases including label distribution skew, feature distribution skew, and quantity skew. Moreover, we conduct extensive experiments on nine datasets to evaluate the accuracy of four state-of-the-art FL algorithms including FedAvg \cite{mcmahan2016communication}, FedProx \cite{lifedprox}, SCAFFOLD \cite{karimireddy2019scaffold}, and FedNova \cite{wang2020tackling}. The experimental results provide insights for the future development of FL algorithms. Last, our code is publicly available \footnote{\url{https://github.com/Xtra-Computing/NIID-Bench}}. Researchers can easily use our code to try different partitioning strategies for the evaluation of existing algorithms or a new algorithm. We also maintain a leaderboard along with our code to rank state-of-the-art federated learning algorithms on different non-IID settings, which can benefit the federated learning community a lot. 

Through extensive studies, we have the following key findings. First, we find that non-IID does bring significant challenges in learning accuracy of FL algorithms, and none of the existing state-of-the-art FL algorithms outperforms others in all cases. \HBS{Second, the effectiveness of FL is highly related to the kind of data skews, e.g., the label distribution skew setting is more challenging than the quantity skew setting. This indicates the importance of having a more comprehensive benchmark on non-IID distributions. Last, in non-IID data setting, instability of the learning process widely exists due to techniques such as batch normalization and partial sampling. This can severely hurt the effectiveness of machine learning services on distributed data silos.}

\rev{Our main contributions are as follows:
\begin{itemize}
    \item We identity non-IID data distributions as a key and common challenge in designing effective federated learning algorithms for distributed data silos and develop a benchmark for researchers' study of federated learning on non-IID data.
    \item We summarize six different partitioning strategies to generate comprehensive non-IID data distribution cases. Among six partitioning strategies, four simple and effective partitioning strategies are designed by our study, while the other two strategies are adopted from existing studies due to their popularity. We also demonstrate the significance of those strategies. None of the previous studies~\cite{lifedprox,karimireddy2019scaffold,wang2020tackling,hsu2019measuring} are as comprehensive as ours. For example, paper \cite{hsu2019measuring} only covers a single partitioning strategy to generate the label distribution skew setting.
    \item Using the proposed partitioning strategies, we conduct an extensive experimental study on four state-of-the-art algorithms, including FedAvg \cite{mcmahan2016communication}, FedProx \cite{lifedprox}, SCAFFOLD \cite{karimireddy2019scaffold}, and FedNova \cite{wang2020tackling}. Moreover, we provide insightful findings and future directions for data management and learning for distributed data silos, which we believe are more and more common in the future.
\end{itemize}
}

The remainder of this paper is structured as follows. We introduce the preliminaries in Section \ref{sec:pre_rel}. \HBS{We review FL algorithms handling non-IID data in Section~\ref{sec:alg}, and present our non-IID data partition strategies in Section \ref{sec:noniid_sim}.} Section \ref{sec:exp} present the experimental results, followed by the future research directions in Section \ref{sec:future_dir}. We discuss the related work in Section \ref{sec:related_work}, and conclude in Section \ref{sec:con}.




\section{Preliminaries}
\label{sec:pre_rel}

\subsection{Notations}
Let $\D=\{(\mb{x}, y)\}$ denote the global dataset. Suppose there are $N$ parties, denoted as $P_1, ..., P_N$. The local dataset of $P_i$ is denoted as $\D^i=\{(\mb{x}_i, y_i)\}$. We use $w^t$ and $w_i^t$ to denote the global model and the local model of party $P_i$ in round $t$, respectively. \HBS{Thus, $w^t$ is the output model of the federated learning process.}

\subsection{FedAvg}
\begin{figure}[!]
\centering
\includegraphics[width=0.7\columnwidth]{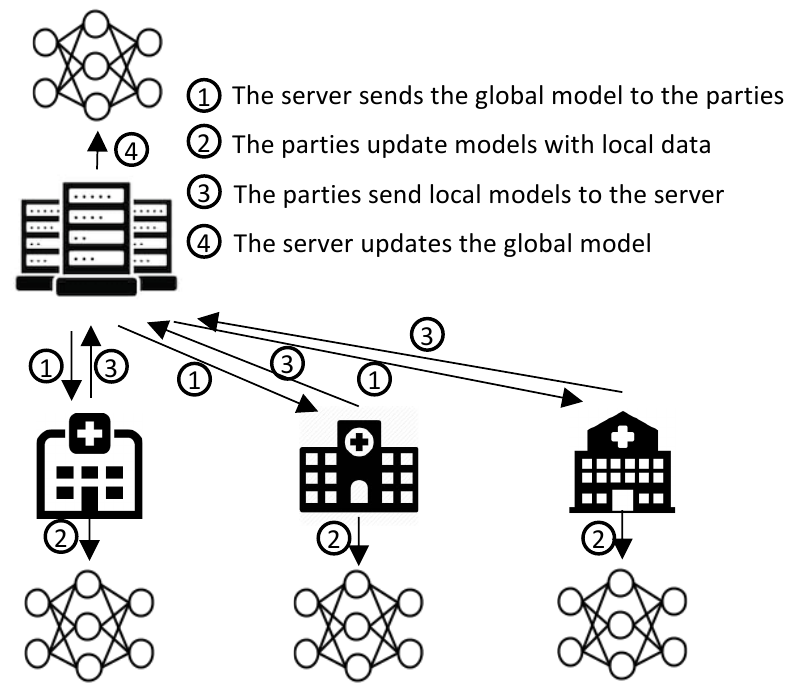}
\caption{The FedAvg framework.}
\label{fig:fedavg}
\end{figure}

FedAvg \cite{mcmahan2016communication} has been a de facto approach for FL. The framework of FedAvg is shown in Figure \ref{fig:fedavg}. In each round, first, the server sends the global model to the \rev{randomly selected parties}. Second, each party updates the model with its local dataset. Then, the updated models are sent back to the server. Last, the server averages the received local models as the updated global model. Unlike traditional distributed SGD, the parties update their local model with multiple epochs, which can decrease the number of communication rounds and is much more communication-efficient. However, the local updates may lead to a bad accuracy, as shown in previous studies \cite{karimireddy2019scaffold,Li2020On,hsu2019measuring}.

\subsection{Effect of Non-IID Data}
\begin{figure}[!]
\centering
\includegraphics[width=0.9\columnwidth]{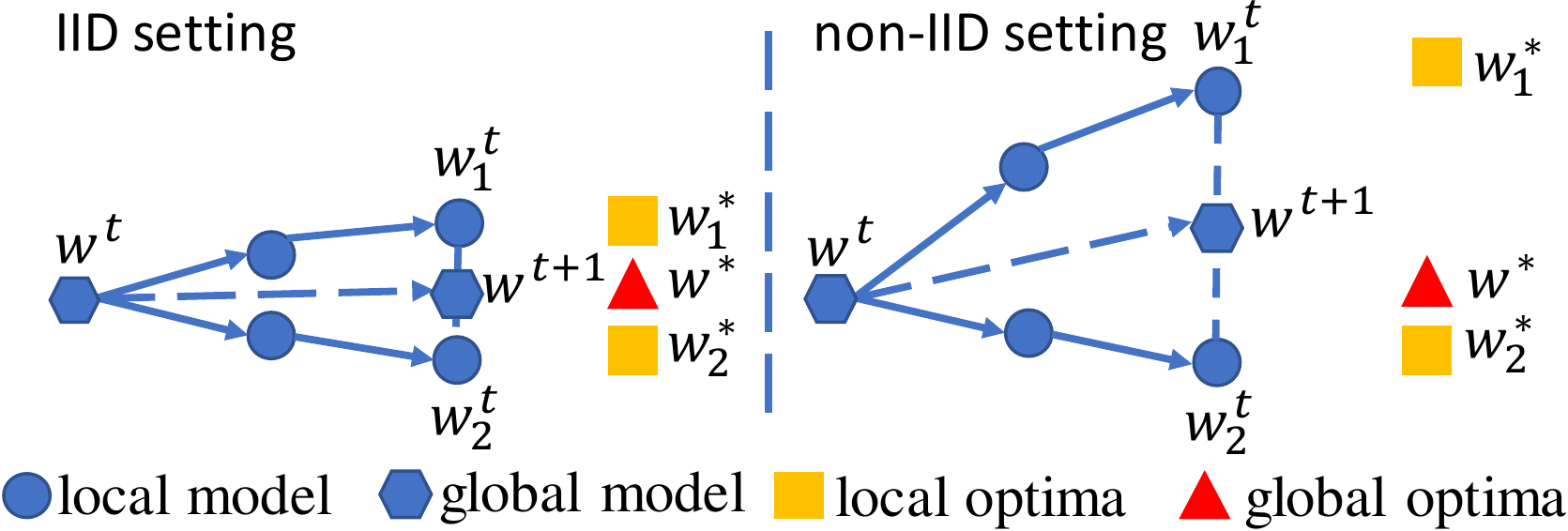}
\caption{Example of a drift under the non-IID setting.}
\label{skew-exa}
\end{figure}

A key challenge in FL is the non-IID data among the parties \cite{li2019flsurvey,kairouz2019advances}. Non-IID data can influence the accuracy of FedAvg a lot. Since the distribution of each local dataset is highly different from the global distribution, the local objective of each party is inconsistent with the global optima. Thus, there exists a \emph{drift} in the local updates \cite{karimireddy2019scaffold}. \rev{In other words, in the local training stage, the local models are updated towards the local optima, which can be far from the global optima.} The averaged model may also be far from the global optima especially when the local updates are large (e.g., a large number of local epochs) \cite{karimireddy2019scaffold,lifedprox,Wang2020Federated,wang2020tackling}. Eventually, the converged global model has much worse accuracy than IID setting. Figure \ref{skew-exa} demonstrates the issue of FedAvg under the non-IID data setting. Under the IID setting, the global optima $w^*$ is close to the local optima $w_1^*$ and $w_2^*$. Thus, the averaged model $w^{t+1}$ is also close to the global optima. However, under the non-IID setting, since $w^*$ is far from $w_1^*$, $w^{t+1}$ can be far from $w^*$. It is challenging to design an effective FL algorithm under the non-IID setting. \HBS{We will present the FL algorithms on handling non-IID data in the next section.}

\section{FL Algorithms on Non-IID Data}\label{sec:alg}

There have been some studies \cite{lifedprox,karimireddy2019scaffold,wang2020tackling} trying to address the drift issue in FL. Here we summarize several state-of-the-art and popular approaches as shown in Algorithm \ref{alg:fed} (FedAvg \cite{mcmahan2016communication}, FedProx \cite{lifedprox}, FedNova \cite{wang2020tackling}) and Algorithm \ref{alg:scaffold} (SCAFFOLD \cite{karimireddy2019scaffold}). These approaches are all based on FedAvg, and we use colors to mark the parts that specially designed in FedProx (red), SCAFFOLD (blue), and FedNova (orange). Note that the studied approaches have the same objective, i.e., learning an effective global model under the non-IID data setting. There are also other FL studies related to non-IID data setting, such as personalizing the local models for each party \cite{fallah2020personalized,dinh2020personalized,hanzely2020lower} and designing robust algorithms against different combinations of local distributions \cite{reisizadeh2020robust,deng2020distributionally,mohri2019agnostic}, which are out of the scope of this paper.

\begin{algorithm}[t]
\SetNoFillComment
\LinesNumbered
\SetArgSty{textnormal}
\KwIn{local datasets $\D^i$, number of parties $N$, number of communication rounds $T$, number of local epochs $E$, learning rate $\eta$}
\KwOut{The final model $w^T$}
\BlankLine
\tb{Server executes}:

initialize $x^0$\\
\For {$t=0, 1, ..., T-1$}{
    Sample a set of parties $S_t$\\
    $n\leftarrow \sum_{i\in S_t} |\D^i|$\\
    \For {$i\in S_t$ \tb{in parallel}}{
        send the global model $w^t$ to party $P_i$
        
         $\Delta w_i^t, \textcolor{orange}{\tau_i} \leftarrow$ \tb{LocalTraining}($i$, $w^t$)
    }
    For FedAvg/FedProx: $w^{t+1} \leftarrow w^t - \eta \sum_{i\in S_t} \frac{|\D^i|}{n} \Delta w_k^{t}$\\
    For FedNova: $w^{t+1} \leftarrow w^{t} - \eta \textcolor{orange}{\frac{ \sum_{i\in S_t} |\D^i|\tau_i}{n}}\sum_{i\in S_t} \frac{|D^i|\Delta w_i^t}{n\textcolor{orange}{\tau_i}} $

   
}

return $w^T$

\BlankLine
\tb{Party executes}:


For FedAvg/FedNova: $L(w; \tb{b}) = \sum_{(x,y)\in \tb{b}} \ell(w; x; y)$\\
For FedProx: $L(w; \tb{b}) = \sum_{(x,y)\in \tb{b}} \ell(w; x; y)\textcolor{red}{ + \frac{\mu}{2} \norm{w-w^t}^2}$ \\

\tb{LocalTraining}($i$, $w^t$):

$w_i^t \leftarrow w^t$ 

$\tau_i \leftarrow 0$

\For{epoch $k = 1, 2, ..., E$}{
    \For{each batch $\tb{b} = \{\tb{x}, y\}$ of $\D^i$}{
        $w_i^t \leftarrow w_i^t - \eta \nabla L(w_i^t; \tb{b})$
        
        $\tau_i \leftarrow \tau_i + 1$
    }
}
$\Delta w_i^t \leftarrow w^t - w_i^t$ 

return $\Delta w_i^t$, $\textcolor{orange}{\tau_i}$ to the server
\caption{A summary of FL algorithms including FedAvg/\textcolor{red}{FedProx}/\textcolor{orange}{FedNova}. We use red and orange colors to mark the part specially included in FedProx and FedNova, respectively.}
\label{alg:fed}
\end{algorithm}

\subsection{FedProx}
FedProx \cite{lifedprox} improves the local objective based on FedAvg. It directly limits the size of local updates. Specifically, as shown in Line 14 of Algorithm \ref{alg:fed}, it introduces an additional $L_2$ regularization term in the local objective function to limit the distance between the local model and the global model. This is a straightforward way to limit the local updates so that the averaged model is not so far from the global optima. A hyper-parameter $\mu$ is introduced to control the weight of the $L_2$ regularization. Overall, the modification to FedAvg is lightweight and easy to implement. FedProx introduces additional computation overhead and does not introduce additional communication overhead. However, one drawback is that users may need to carefully tune $\mu$ to achieve good accuracy. If $\mu$ is too small, then the regularization term has almost no effect. If $\mu$ is too big, then the local updates are very small and the convergence speed is slow.

\subsection{FedNova}
Another recent study, FedNova \cite{wang2020tackling}, improves FedAvg in the aggregation stage. It considers that different parties may conduct different numbers of local steps (i.e., the number of mini-batches in the local training) each round. This can happen when parties have different computation power given the same time constraint or parties have different local dataset size given the same number of local epochs and batch size. Intuitively, the parties with a larger number of local steps will have a larger local update, which will have a more significant influence on the global updates if simply averaged. Thus, to ensure that the global updates are not biased, FedNova normalizes and scales the local updates of each party according to their number of local steps before updating the global model (see Line 10 of Algorithm \ref{alg:fed}). FedNova also only introduces lightweight modifications to FedAvg, and negligible computation overhead when updating the global model.

\subsection{SCAFFOLD}
SCAFFOLD \cite{karimireddy2019scaffold} models non-IID as introducing variance among the parties and applies the variance reduction technique \cite{schmidt2017minimizing,johnson2013accelerating}. It introduces control variates for the server (i.e., $c$) and parties (i.e., $c_i$), which are used to estimate the update direction of the server model and the update direction of each client. Then, the drift of local training is approximated by the difference between these two update directions. Thus, SCAFFOLD corrects the local updates by adding the drift in the local training (Line 20 of Algorithm \ref{alg:scaffold}). SCAFFOLD proposes two approaches to update the local control variates (Line 23 of Algorithm \ref{alg:scaffold}), by computing the gradient of the local data at the global model or by reusing the previously computed gradients. The second approach has a lower computation cost while the first one may be more stable. Compared with FedAvg, intuitively, SCAFFOLD doubles the communication size per round due to the additional control variates.

\begin{algorithm}[t]
\SetNoFillComment
\LinesNumbered
\SetArgSty{textnormal}
\KwIn{same as Algorithm \ref{alg:fed}}
\KwOut{The final model $w^T$}

\BlankLine
\tb{Server executes}:

initialize $x^0$\\
\textcolor{blue}{$c^t \leftarrow \mb{0}$}\\
\For {$t=0, 1, ..., T-1$}{
    Randomly sample a set of parties $S_t$\\
    $n\leftarrow \sum_{i\in S_t} |\D^i|$\\
    \For {$i\in S_t$ \tb{in parallel}}{
        send the global model $w^t$ to party $P_i$
         $\Delta w_i^t, \textcolor{blue}{\Delta c} \leftarrow$ \tb{LocalTraining}($i$, $w^t$, \textcolor{blue}{$c^t$})

    }
    $w^{t+1} \leftarrow w^t - \eta \sum_{i\in S_t} \frac{|\D^i|}{n} \Delta w_k^{t}$
    
    
    $\textcolor{blue}{c^{t+1} \leftarrow c^{t} + \frac{1}{N} \Delta c}$
}

return $w^T$

\BlankLine
\tb{Party executes}:

$L(w; \tb{b}) = \sum_{(x,y)\in \tb{b}} \ell(w; x; y)$

$\textcolor{blue}{c_i \leftarrow \mb{0}}$\\
\tb{LocalTraining}($i$, $w^t$, $\textcolor{blue}{c^t}$):

$w_i^t \leftarrow w^t$ 

$\tau_i \leftarrow 0$

\For{epoch $k = 1, 2, ..., E$}{
    \For{each batch $\tb{b} = \{\tb{x}, y\}$ of $\D^i$}{
        $w_i^t \leftarrow w_i^t - \eta (\nabla L(w_i^t; \tb{b})\textcolor{blue}{ - c_i^t + c})$
        
        $\tau_i \leftarrow \tau_i + 1$
    }
}
$\Delta w_i^t \leftarrow w^t - w_i^t$ 

$\textcolor{blue}{c_i^*\leftarrow(i) \nabla L(w_i^t), or (ii) c_i - c + \frac{1}{\tau_i\eta} (w^t - w_i^t)}$

$\textcolor{blue}{\Delta c \leftarrow c_i^* - c_i}$

$\textcolor{blue}{c_i \leftarrow c_i^*}$

return $\Delta w_i^t$, $\textcolor{blue}{\Delta c}$ to the server
\caption{The SCAFFOLD algorithm. We use blue color to mark the part specially included in SCAFFOLD compared with FedAvg.}
\label{alg:scaffold}
\end{algorithm}

\subsection{Other Studies}

\QB{When preparing this paper, there are other contemporary works \cite{acar2021federated,li2021fedbn,wang2020addressing,li2021model} on federated learning under non-IID setting. \cite{acar2021federated} proposes FedDyn, which adds a regularization term in the local training based on the global model and the model from the previous round. \cite{li2021fedbn} proposes FedBN for feature shift non-IID setting, where the client batch-norm layers are updated locally without communicating to the server. \cite{wang2020addressing} applies a monitor to detect class imbalance in the training process, and proposes a new loss function to address it. \cite{li2021model} proposes model-contrastive learning. Their approach corrects the local training by comparing the representations learned by the current local model, the local model from the previous round, and the global model. We leave the comparison between these studies as future studies.}

\subsection{Motivation of this study}

Non-IID is a key and common data challenge for developing effective federated learning algorithms. Although previous studies~\cite{lifedprox, karimireddy2019scaffold, wang2020tackling} have demonstrated preliminary and promising results over FedAvg on non-IID data, as we will summarize in Table \ref{tbl:study_comp} in later section, all above studies have evaluated only one or two non-IID distributions, and tried rigid data partitioning strategies in the experiments. There is still no standard benchmark or a systematic study to evaluate the effectiveness of these FL algorithms. This motivates us to develop a benchmark with more comprehensive data distributions as well as data partitioning strategies, and then we can evaluate the pros and cons of existing algorithms and outline the challenges and opportunities for future federated learning on non-IID data.

\section{Simulating Non-IID Data Setting}
\label{sec:noniid_sim}

As existing studies only adopt limited partitioning strategies, they cannot represent a comprehensive view of non-IID cases. To bridge this gap, we develop a benchmark named NIID-Bench.

\subsection{Research Problems}
We need to address two key research problems. The first one is on data sets: whether to use real-world non-IID datasets or synthetic datasets. The second one is on how to design comprehensive non-IID scenarios.

\HBS{For the first problem, we choose to synthesize the distributed non-IID datasets by partitioning a real-world dataset into multiple smaller subsets.} Many existing studies \cite{mcmahan2016communication,karimireddy2019scaffold,wang2020tackling} use the partitioning approach to simulate the non-IID federated setting.  Compared with using real federated datasets \cite{caldas2018leaf,hu2020oarf}, adopting partitioning strategies has the following advantages. \rev{First, while it is challenging to evaluate the imbalance properties (e.g., imbalanced level and imbalanced case) in real federated datasets, partitioning strategies can easily quantify and control the imbalance properties of the local data. Thus, researchers can easily investigate the behavior of algorithms by trying different imbalanced settings, which is essential to the development of FL algorithms. Second, when using synthetic datasets, one can easily set different factors (e.g., number of parties, size of data) that are important in the FL experiments. However, a real federated dataset usually corresponds to a fixed federated setting. Last, due to data regulation and privacy concerns, meaningful real federated datasets are difficult to obtain~\cite{hu2020oarf}. Even if we can obtain such real datasets, they do not have the previous two advantages of synthetic data sets. It is more flexible to develop partitioning strategies on existing widely used public datasets, which already have lots of centralized training knowledge as reference, as well as to simulate different non-IID scenarios. There are also limitations of using generated datasets compared with using real federated datasets. The generated datasets may not fully capture the real data distributions, which can be complicated and challenging to quantity. Note that the usage of generated federated datasets and real federated datasets are orthogonal. It is an interesting future study to find and study meaningful real-world data sets and application scenarios.}

\rev{For the second problem, an existing study \cite{kairouz2019advances} gives a very good and comprehensive summary on non-IID data cases from a distribution perspective. Specifically, considering the local data distribution $P(x_i, y_i)=P(x_i|y_i)P(y_i)$ or $P(x_i, y_i)=P(y_i|x_i)P(x_i)$, the previous study~\cite{kairouz2019advances} summaries five different non-IID cases: (1) label distribution skew (i.e., $P(y_i)$ is different among parties); (2) feature distribution skew (i.e., $P(x_i)$ is different among parties); (3) same label but different features (i.e., $P(x_i|y_i)$ is different among parties); (4) same features but different labels (i.e., $P(y_i|x_i)$ is different among parties); (5) quantity skew (i.e., $P(x_i, y_i)$ is same but the amount of data is different among parties). Here the third case is mainly related to vertical FL (the parties share the same sample IDs but different features). As mentioned in the third paragraph of Section \ref{sec:intro}, we focus on horizontal FL in this paper, where each party shares the same feature space but owns different samples. The fourth case is not applicable in most FL studies, which assume there is a common knowledge $P(y|x)$ among the parties to learn. Otherwise, techniques such as domain adaption \cite{Peng2020Federated} or personalized federated learning (i.e., each party learns a personalized local model) \cite{dinh2020personalized,fallah2020personalized} can be applied in federated learning, which is out of the scope of our paper. Thus, we consider label distribution skew, feature distribution skew, and quantity skew as possible non-IID data distribution cases in this paper. While the five non-IID data cases cover all possible single type of skew, there may be mixed types of skew, which we will discuss in Section \ref{sec:mixed_type}.}

\begin{figure}[!]
\centering
\subfloat[The label distribution for Criteo. The value in cell $(a,b)$ is the amount of data samples of class $b$ belonging to Party $a$.\label{fig:criteo}]{\includegraphics[width=0.9\linewidth]{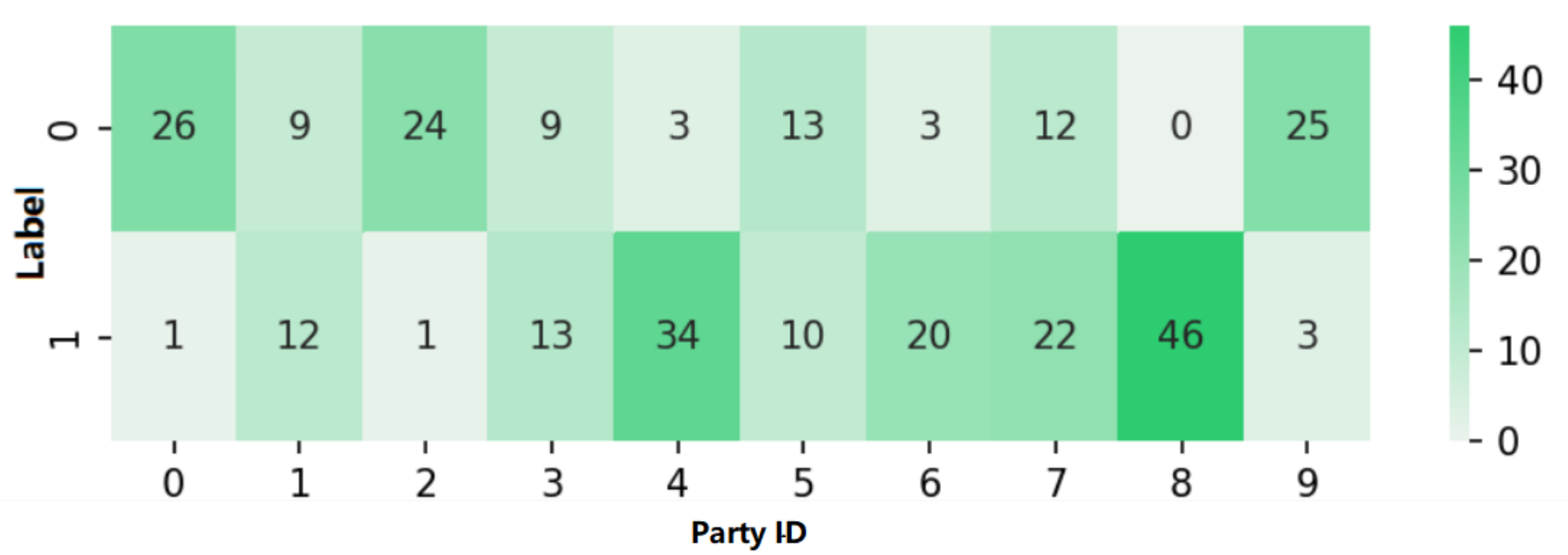}%
}
\hfill
\subfloat[The feature distribution for Digits. The triangles are the visualized features of SVHN and the circles are the visualized features of MNIST.\label{fig:digits}]{\includegraphics[width=0.8\linewidth]{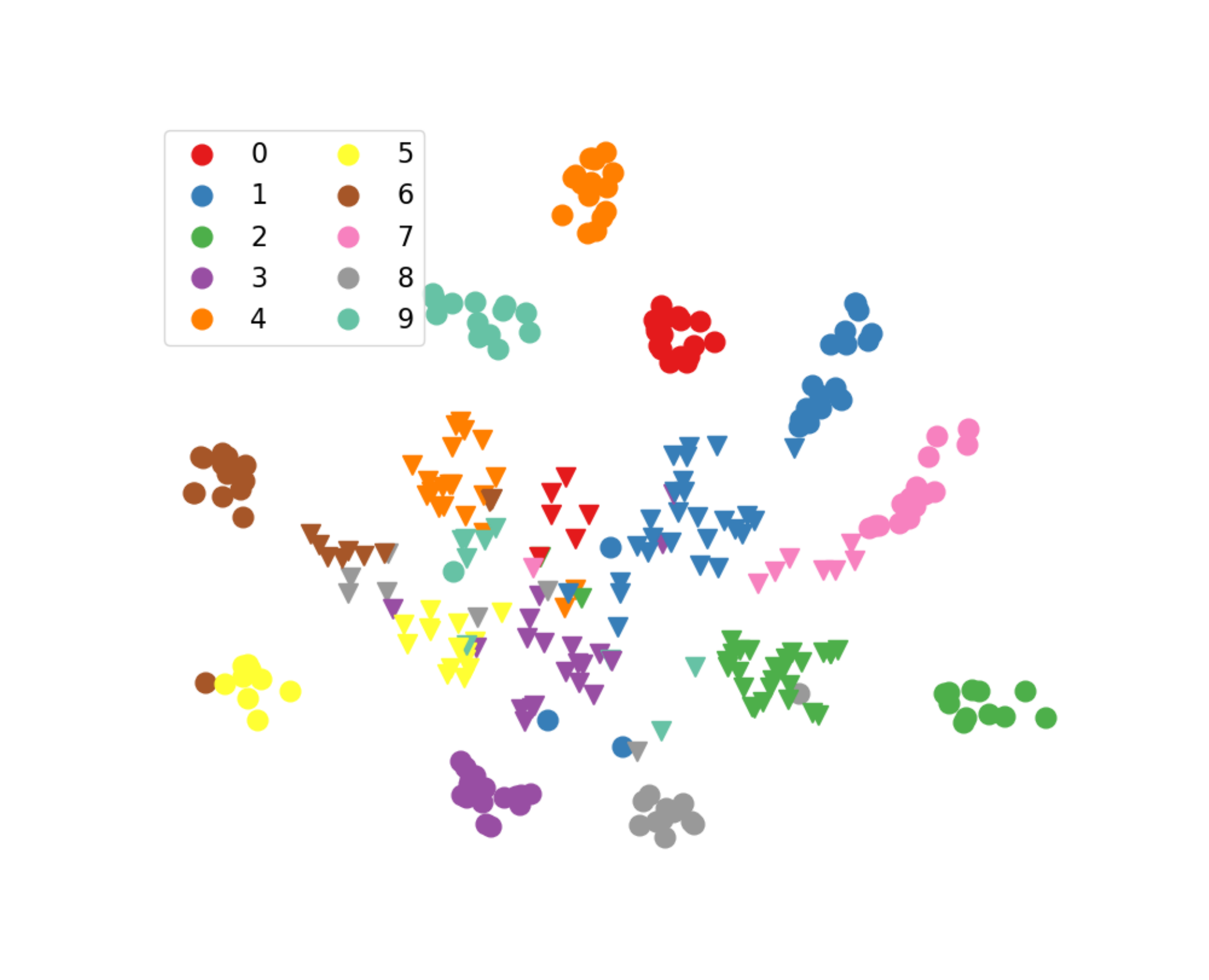}%
}
\caption{\rev{The non-IID properties of \emph{criteo} and \emph{digits}.}}
\label{fig:real-non-iid}
\end{figure}

\rev{We use two real-world datasets, \emph{Criteo} \cite{DiemertMeynet2017} and \emph{Digits} \cite{Peng2020Federated}, to demonstrate the non-IID properties. Criteo contains feature values and click feedback for millions of display ads, which can be used for clickthrough rate prediction. Digits contains multiple subsets for digit classification. In Criteo, taking each user as a party, we select ten parties and draw the label distribution as shown in Figure \ref{fig:criteo}. We can observe that there exists both label distribution skew (e.g., Party 0 and Party 4) and quantity skew (e.g., Party 0 and Party 8) among the parties. In Digits, taking each subset (e.g., \emph{MNIST} and \emph{SVHN}) as a party, we train a model using these subsets and draw the feature distribution using t-SNE \cite{maaten2008visualizing} as shown in Figure \ref{fig:digits}. For each class, althougth MNIST and SVHN have the same label, the feature distributions of MNIST and SVHN are significantly different from each other. Feature skew exists in the Digits dataset. These two examples show that the considered non-IID data cases are reasonable and practical. 
}

\subsection{Label Distribution Skew}
In label distribution skew, the label distributions $P(y_i)$ vary across parties. Such a case is common in practice. For example, some hospitals are more specialized in several specific kinds of diseases and have more patient records on them. To simulate label distribution skew, we introduce two different label imbalance settings: quantity-based label imbalance and distribution-based label imbalance.

\paragraph{Quantity-based label imbalance}
Here each party owns data samples of a fixed number of labels. This is first introduced in the experiments of FedAvg \cite{mcmahan2016communication}, where the data samples with the same label are divided into subsets and each party is only assigned 2 subsets with different labels. Following FedAvg, such a setting is also used in many other studies \cite{geyer2017differentially,lifedprox}. \cite{felix2020federated} considers a highly extreme case, where each party only has data samples with a single label. We introduce a general partitioning strategy to set the number of labels that each party has. Suppose each party only has data samples of $k$ different labels. We first randomly assign $k$ different label IDs to each party. Then, for the samples of each label, we randomly and equally divide them into the parties which own the label. In this way, the number of labels in each party is fixed, and there is no overlap between the samples of different parties. For ease of presentation, we use $\#C=k$ to denote such a partitioning strategy.



\begin{figure}[!]
\centering
\includegraphics[width=0.8\columnwidth]{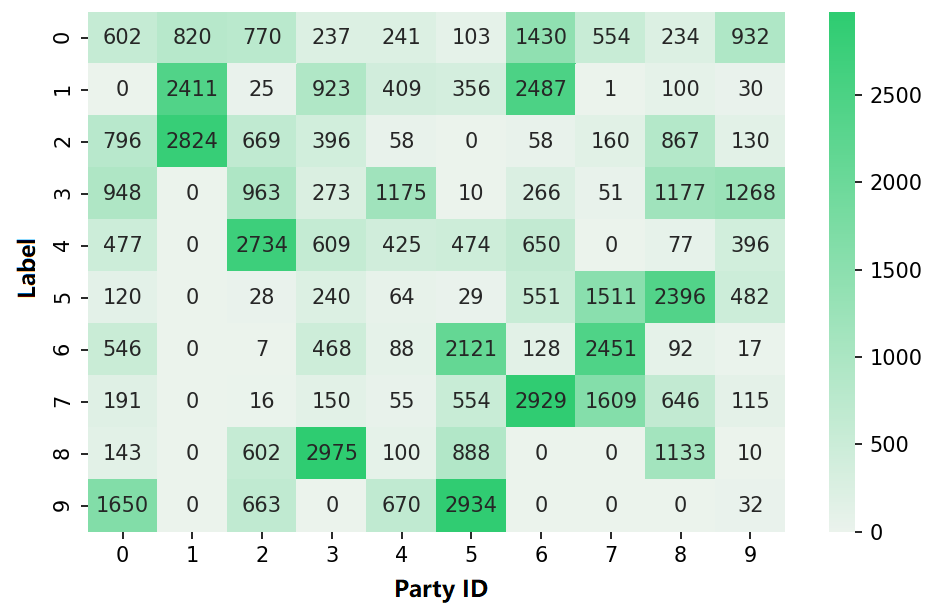}
\caption{An example of distribution-based label imbalance partition on MNIST \cite{lecun1998gradient} dataset with $\beta=0.5$. The value in each rectangle is the number of data samples of a class belonging to a certain party.}
\label{fig:diri}
\vspace{-15pt}
\end{figure}

\paragraph{Distribution-based label imbalance}
\label{sec:dis_label_imb}
Another way to simulate label imbalance is that each party is allocated a proportion of the samples of each label according to Dirichlet distribution. \QB{Dirichlet distribution is commonly used as prior distribution in Bayesian statistics \cite{huang2005maximum} and is an appropriate choice to simulate real-world data distribution.} Specifically, we sample $p_k \sim Dir_N(\beta)$ and allocate a $p_{k,j}$ proportion of the instances of class $k$ to party $j$. Here $Dir(\cdot)$ denotes the Dirichlet distribution and $\beta$ is a concentration parameter ($\beta > 0$). This partitioning strategy was first used in \cite{pmlr-v97-yurochkin19a} and has been used in many recent studies \cite{Wang2020Federated,lin2020ensemble,wang2020tackling,li2020model}. An advantage of this approach is that we can flexibly change the imbalance level by varying the concentration parameter $\beta$. If $\beta$ is set to a smaller value, then the partition is more unbalanced. An example of such a partitioning strategy is shown in Figure \ref{fig:diri}. For ease of presentation, we use $p_k \sim Dir(\beta)$ to denote such a partitioning strategy.

\begin{figure}[!]
\centering
\subfloat[add noises from $Gau(0.001)$]{\includegraphics[width=0.45\linewidth]{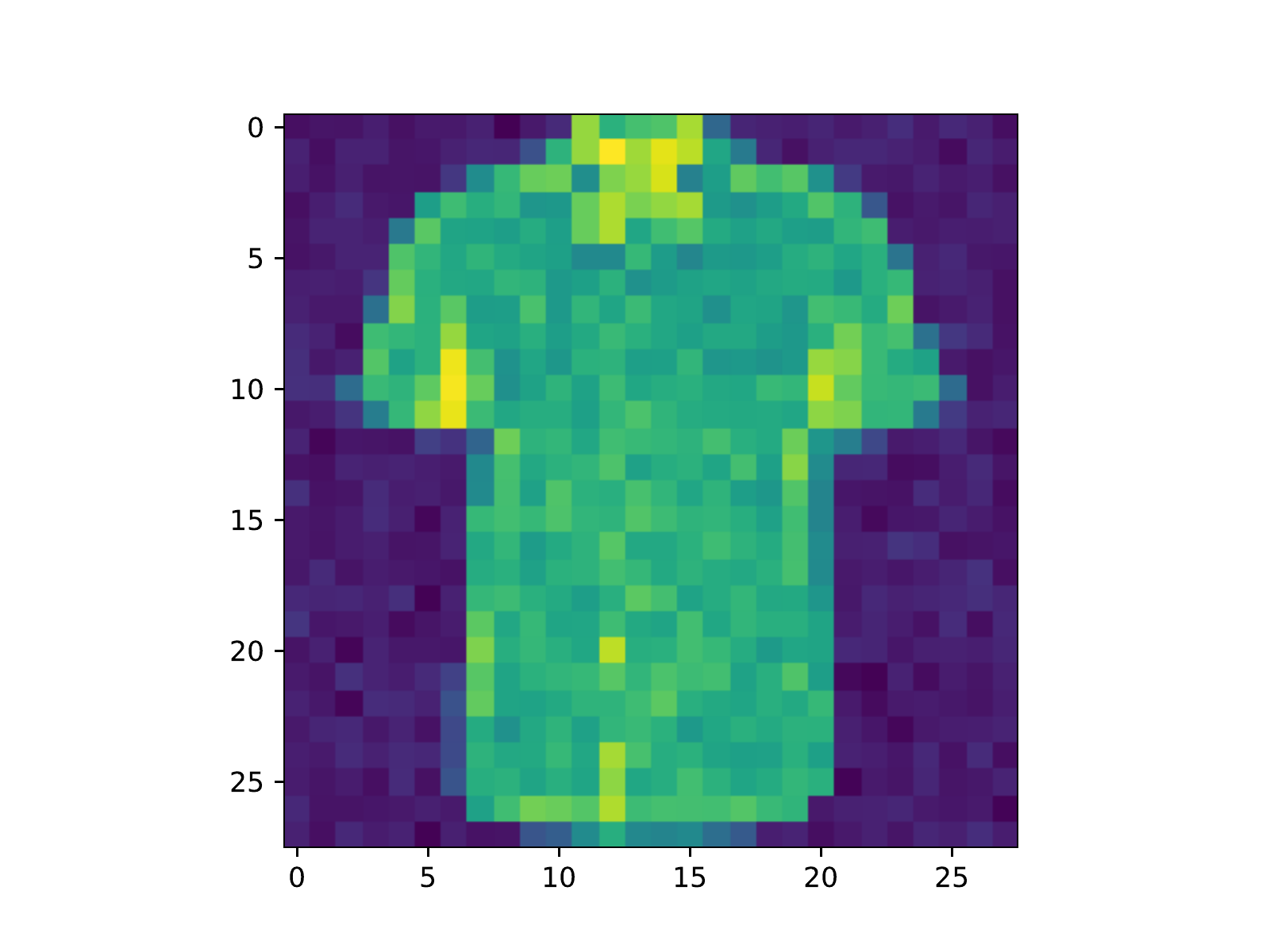}%
}
\subfloat[add noises from $Gau(0.01)$]{\includegraphics[width=0.45\linewidth]{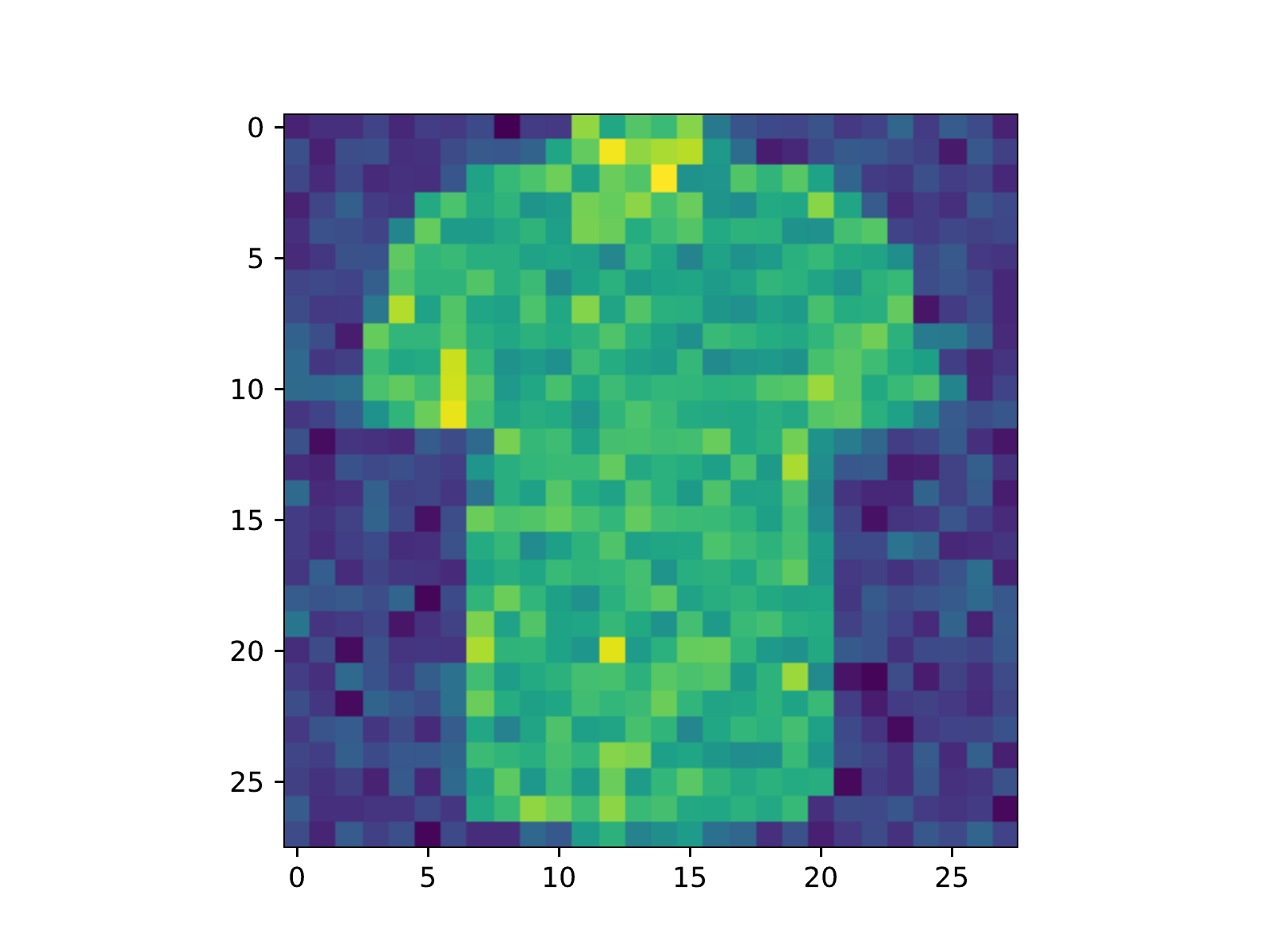}%
}
\caption{An example of adding noises on FMNIST \cite{xiao2017fashion} dataset. On party $P_1$, noises sampled from $Gau(0.001)$ are added into its images. On party $P_2$, noises sampled from $Gau(0.01)$ are added into its images.}
\label{fig:addnoise}
\end{figure}







\subsection{Feature Distribution Skew}
In feature distribution skew, the feature distributions $P(x_i)$ vary across parties although the knowledge $P(y_i|\mb{x}_i)$ is same. For example, cats may vary in coat colors and patterns in different areas. Here we introduce three different settings to simulate feature distribution skew: noise-based feature imbalance, synthetic feature imbalance, and real-world feature imbalance.

\paragraph{Noise-based feature imbalance} We first divide the whole dataset into multiple parties randomly and equally. For each party, we add different levels of Gaussian noise to its local dataset to achieve different feature distributions. \QB{We choose Gaussian noise due to its popularity especially in images \cite{zhang2017beyond}.} Specifically, given user-defined noise level $\sigma$, we add noises $\hat{\mb{x}} \sim Gau(\sigma\cdot i/N)$ for Party $P_i$, where $Gau(\sigma\cdot i/N)$ is a Gaussian distribution with mean 0 and variance $\sigma\cdot i/N$. Users can change $\sigma$ to increase the feature dissimilarity among the parties. Figure \ref{fig:addnoise} is an example of noise-based feature imbalance on FMNIST dataset \cite{xiao2017fashion}. For ease of presentation, we use $\hat{\mb{x}} \sim Gau(\sigma)$ to present such a partitioning strategy.



\begin{figure}[!]
\centering
\includegraphics[width=0.4\linewidth]{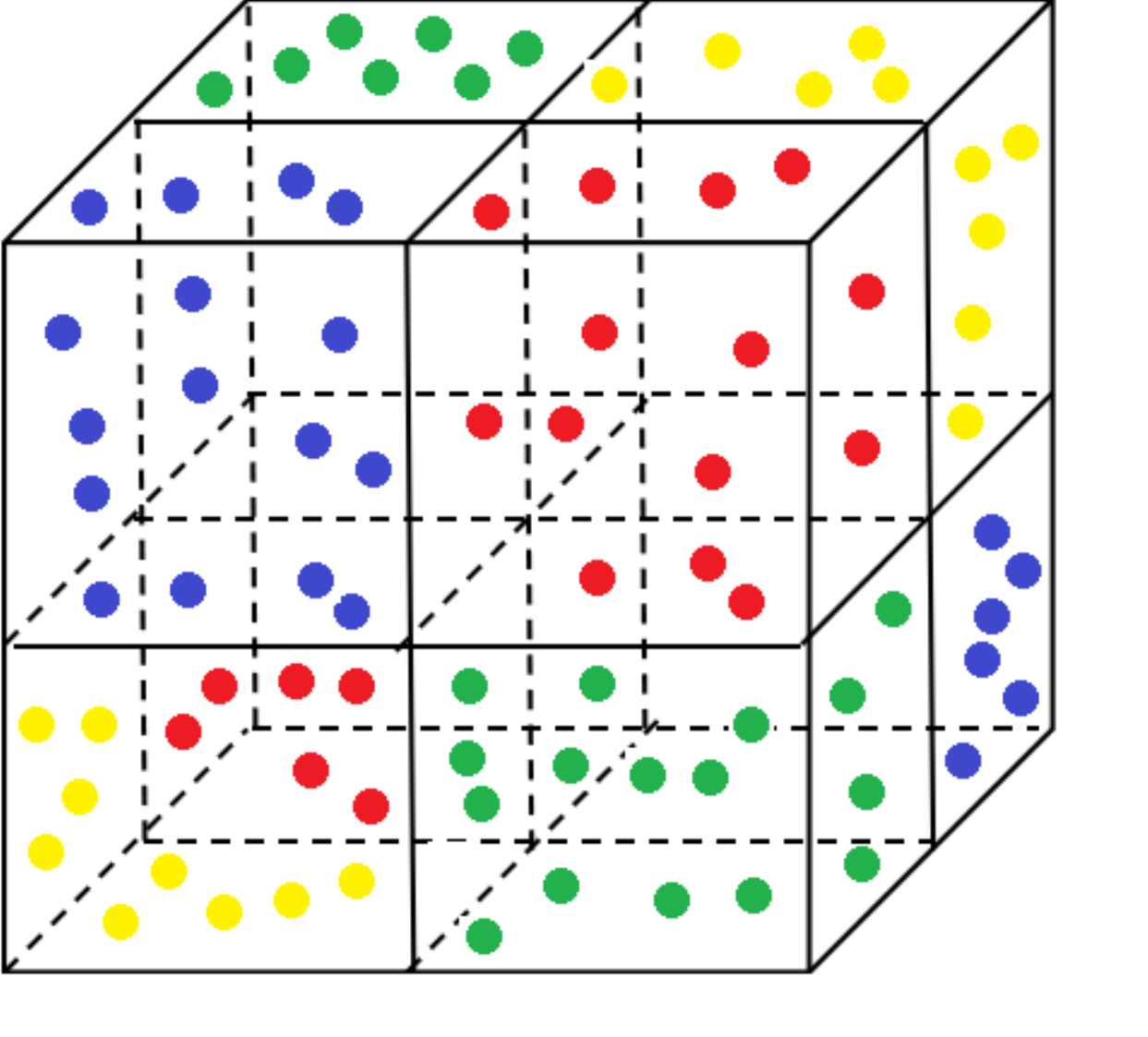}
\caption{The visualization of our FCUBE dataset. The data points within the upper four cubes have label 0 and within the lower four cubes have label 1. There are a total of eight cubes with four colors
. The data points with the same color are assigned to a party.}
\label{fig:synthe}
\end{figure}

\begin{table*}[t]
\caption{The experimental settings in existing studies and our benchmark. Note that the quantity-based, noised-based, and quantity skew partitioning strategies in the existing studies are different form the strategies proposed in our study.}
\vspace{-5pt}
\label{tbl:study_comp}
\newcommand{\y}{\ding{51}}
\newcommand{\n}{\ding{55}}
\centering
\resizebox{0.8\linewidth}{!}{%
\begin{tabular}{|c|c|c|c|c|c|c|}
\hline
\multicolumn{2}{|c|}{Partitioning strategies} & FedAvg & FedProx & SCAFFOLD & FedNova &NIID-Bench\\ \hline
\multirow{2}{*}{Label distribution skew} & quantity-based & \y & \y  & \n &\n &\y \\ \cline{2-7} 
 & distribution-based &\n  &\n  &\y &\y&\y \\ \hline
\multirow{3}{*}{Feature distribution skew} & noise-based &\n  &\n  &\n&\n &\y \\ \cline{2-7} 
 & synthetic &\n  &\y  &\n &\n&\y \\ \cline{2-7} 
 & real-world &\n  &\y  &\n &\n &\y \\ \hline
\multicolumn{2}{|c|}{Quantity skew} &\n  &\n  &\n &\y &\y  \\ \hline
\end{tabular}
}
\end{table*}

\paragraph{Synthetic feature imbalance} We generate a synthetic feature imbalance federated dataset named FCUBE. Suppose the distribution of data points is a cube in three dimensions (i.e, $(x_1, x_2, x_3)$) which have two different labels classified by plane $x_1=0$. As shown in Figure \ref{fig:synthe}, we divide the cube into 8 parts by planes $x_1=0$, $x_2=0$, and $x_3=0$. Then, we allocate two parts which are symmetric of (0,0,0) to a subset for each party. In this way, feature distribution varies among parties while labels are still balanced. 

\paragraph{Real-world feature imbalance} The EMNIST dataset \cite{cohen2017emnist} collects handwritten characters/digits from different writers. Then, like \cite{caldas2018leaf}, it is natural to partition the dataset into different parties according to the writers. Since the character features usually differ among writers (e.g, stroke width, slant), there is a natural feature distribution skew among different parties. Specifically, for the digit images of EMNIST, we divide and assign the writers (and their digits) into each party randomly and equally. Since each party has different writers, the feature distributions are different among the parties. Like \cite{caldas2018leaf}, we call this federated dataset as FEMNIST.




\subsection{Quantity Skew}

In quantity skew, the size of the local dataset $|\D^i|$ varies across parties. Although data distribution may still be consistent among the parties, it is interesting to see the effect of the quantity imbalance in FL. Like distribution-based label imbalance setting, we use Dirichlet distribution to allocate different amounts of data samples into each party. We sample $q \sim Dir_N(\beta)$ and allocate a $q_j$ proportion of the total data samples to $P_j$. The parameter $\beta$ can be used to control the imbalance level of the quantity skew. For ease of presentation, we use $q \sim Dir(\beta)$ to denote such a partitioning strategy.

\subsection{Experiments in Existing Studies}
\HBS{Table \ref{tbl:study_comp} compares the partitioning strategies in NIID-bench with the experimental settings in existing studies. We can observe that each study only covers partial non-IID cases. It is impossible to directly compare the results presented in different papers. In contrast, NIID-bench consists of six partitioning strategies, which are more comprehensive and representative for representing different non-IID data cases.}




\section{Experiments}
\label{sec:exp}

To investigate the effectiveness of existing FL algorithms on non-IID data setting, we conduct extensive experiments on nine public datasets, including six image datasets (i.e., MNIST \cite{lecun1998gradient}, CIFAR-10 \cite{krizhevsky2009learning}, FMNIST \cite{xiao2017fashion}, SVHN \cite{netzer2011reading}, FCUBE, FEMNIST \cite{caldas2018leaf}) and three tabular datasets (i.e., adult, rcv1, and covtype)\footnote{\url{https://www.csie.ntu.edu.tw/~cjlin/libsvmtools/datasets/}}. The statistics of the datasets are summarized in Table \ref{tbl:datasets}. For the image datasets, we use a CNN, which has two 5x5 convolution layers followed by 2x2 max pooling (the first with 6 channels and the second with 16 channels) and two fully connected layers with ReLU activation (the first with 120 units and the second with 84 units). For the tabular datasets, we use a MLP with three hidden layers. The numbers of hidden units of three layers are 32, 16, and 8. The number of parties is set to 10 by default, except for FCUBE where the number of parties is set to 4. All parties participate in every round to eliminate the effect of randomness brought by party sampling by default~\cite{mcmahan2016communication}. We use the SGD optimizer with learning rate 0.1 for rcv1 and learning rate 0.01 for the other datasets (tuned from $\{0.1, 0.01, 0.001\}$) and momentum 0.9. The batch size is set to 64 and the number of local epochs is set to 10 by default.

\textbf{Benchmark metrics.} We use the top-1 accuracy on the test dataset as a metric to compare the studied algorithms. We run all the studied algorithms for the same number of rounds for fair comparison. The number of rounds is set to 50 by default unless specified.

Due to the page limit, for the experiments on the effect of batch size and model architecture, please refer to Appendix D and E of the technical report \cite{li2021federated}, respectively.

\begin{table}[]
\centering
\caption{The statistics of datasets in the experiments.}
\vspace{-5pt}
\label{tbl:datasets}
\resizebox{\columnwidth}{!}{
\begin{tabular}{|c|c|c|c|c|}
\hline
Datasets & \#training instances & \#test instances & \#features & \#classes \\ \hline
MNIST & 60,000 & 10,000 & 784 & 10 \\ \hline
FMNIST & 60,000 & 10,000 & 784 & 10 \\ \hline
CIFAR-10 & 50,000 & 10,000 & 1,024 & 10 \\ \hline
SVHN & 73,257 & 26,032 & 1,024 & 10 \\ \hline
adult & 32,561 & 16,281 & 123 & 2 \\ \hline
rcv1 & 15,182 & 5,060 & 47,236 & 2 \\ \hline
covtype & 435,759 & 145,253 & 54 & 2 \\ \hline
FCUBE & 4,000 & 1,000 & 3 & 2 \\ \hline
FEMNIST &341,873  &40,832  & 784 & 10 \\ \hline
\end{tabular}
}
\end{table}

\begin{table*}[!]
\centering
\caption{The top-1 accuracy of different approaches. We run three trials and report the mean accuracy and standard derivation. For FedProx, we tune $\mu$ from $\{0.001, 0.01, 0.1, 1\}$ and report the best accuracy.}
\label{tbl:perf}
\resizebox{0.85\textwidth}{!}{
\begin{tabular}{|c|c|c|c|c|c|c|}
\hline
category & dataset & partitioning & FedAvg & FedProx & SCAFFOLD & FedNova  \\ \hline\hline
\multirow{22}{*}{\begin{tabular}[c]{@{}c@{}}Label\\distribution\\skew\end{tabular}} & \multirow{4}{*}{MNIST} & $p_k \sim Dir(0.5)$ & 98.9\%$\pm$0.1\%  & 98.9\%$\pm$0.1\%  & \tb{99.0\%$\pm$0.1\%} & 98.9\%$\pm$0.1\% \\ \cline{3-7} 
 &  & $\#C=1$ & 29.8\%$\pm$7.9\%  & \tb{40.9\%$\pm$23.1\%}  & 9.9\%$\pm$0.2\% & 39.2\%$\pm$22.1\%\\ \cline{3-7} 
 &  & $\#C=2$ & \tb{97.0\%$\pm$0.4\%}  & 96.4\%$\pm$0.3\%  & 95.9\%$\pm$0.3\% & 94.5\%$\pm$1.5\% \\ \cline{3-7} 
 &  & $\#C=3$ & \tb{98.0\%$\pm$0.2\%}  & 97.9\%$\pm$0.4\%  & 96.6\%$\pm$1.5\% & \tb{98.0\%$\pm$0.3\%}\\ \cline{2-7} 
 & \multirow{4}{*}{FMNIST} & $p_k \sim Dir(0.5)$ & 88.1\%$\pm$0.6\%  & 88.1\%$\pm$0.9\%  & 88.4\%$\pm$0.5\% & \tb{88.5\%$\pm$0.5\%}  \\ \cline{3-7} 
 &  & $\#C=1$ & 11.2\%$\pm$2.0\%  & \tb{28.9\%$\pm$3.9\%}  & 12.8\%$\pm$4.8\% & 14.8\%$\pm$5.9\% \\ \cline{3-7} 
 &  & $\#C=2$ & \tb{77.3\%$\pm$4.9\%}  & 74.9\%$\pm$2.6\%  & 42.8\%$\pm$28.7\% & 70.4\%$\pm$5.1\%\\ \cline{3-7} 
 &  & $\#C=3$ & 80.7\%$\pm$1.9\%  & \tb{82.5\%$\pm$1.9\%}  & 77.7\%$\pm$3.8\% & 78.9\%$\pm$3.0\%\\ \cline{2-7} 
 & \multirow{4}{*}{CIFAR-10} & $p_k \sim Dir(0.5)$ & 68.2\%$\pm$0.7\% & 67.9\%$\pm$0.7\% & \tb{69.8\%$\pm$0.7\%}& 66.8\%$\pm$1.5\% \\ \cline{3-7} 
 &  & $\#C=1$ & 10.0\%$\pm$0.0\% & \tb{12.3\%$\pm$2.0\%} & 10.0\%$\pm$0.0\% &10.0\%$\pm$0.0\%\\ \cline{3-7} 
 &  & $\#C=2$ & 49.8\%$\pm$3.3\% & \tb{50.7\%$\pm$1.7\%} & 49.1\%$\pm$1.7\% &46.5\%$\pm$3.5\% \\ \cline{3-7} 
 &  & $\#C=3$ & \tb{58.3\%$\pm$1.2\%} & 57.1\%$\pm$1.2\% & 57.8\%$\pm$1.4\% &54.4\%$\pm$1.1\%\\ \cline{2-7} 
 & \multirow{4}{*}{SVHN} & $p_k \sim Dir(0.5)$ & 86.1\%$\pm$0.7\% & 86.6\%$\pm$0.9\% & \tb{86.8\%$\pm$0.3\%}&86.4\%$\pm$0.6\% \\ \cline{3-7} 
 &  & $\#C=1$ & 11.1\%$\pm$0.0\% & \tb{19.6\%$\pm$0.0\%} & 6.7\%$\pm$0.0\%&10.6\%$\pm$0.8\% \\ \cline{3-7} 
 &  & $\#C=2$ & \tb{80.2\%$\pm$0.8\%} & 79.3\%$\pm$0.9\% & 62.7\%$\pm$11.6\% &75.4\%$\pm$4.8\% \\ \cline{3-7} 
 &  & $\#C=3$ & 82.0\%$\pm$0.7\% & \tb{82.1\%$\pm$1.0\%}  & 77.2\%$\pm$2.0\% &80.5\%$\pm$1.2\%\\ \cline{2-7} 
 & \multirow{2}{*}{adult} & $p_k \sim Dir(0.5)$ & 78.4\%$\pm$0.9\% & \tb{80.5\%$\pm$0.7\%} & 76.4\%$\pm$0.0\%&52.3\%$\pm$26.7\% \\ \cline{3-7} 
 &  & $\#C=1$ & \tb{82.5\%$\pm$2.2\%} &76.4\%$\pm$0.0\% & 23.6\%$\pm$0.0\%&50.8\%$\pm$0.9\% \\ \cline{2-7} 
  & \multirow{2}{*}{rcv1} & $p_k \sim Dir(0.5)$ & 48.2\%$\pm$0.7\% & \tb{70.3\%$\pm$13.3\%} & 64.4\%$\pm$24.3\%&49.3\%$\pm$2.1\% \\ \cline{3-7} 
 &  & $\#C=1$ & \tb{51.8\%$\pm$0.7\%} &\tb{51.8\%$\pm$0.7\%} & \tb{51.8\%$\pm$0.7\%}&\tb{51.8\%$\pm$0.7\%} \\ \cline{2-7} 
  & \multirow{2}{*}{covtype} & $p_k \sim Dir(0.5)$ & \tb{77.2\%$\pm$7.4\%} & 70.9\%$\pm$0.7\% & 67.7\%$\pm$14.9\%&74.8\%$\pm$12.9\% \\ \cline{3-7} 
 &  & $\#C=1$ & 48.8\%$\pm$0.1\% & \tb{59.1\%$\pm$2.1\%} & 49.6\%$\pm$1.4\%&50.4\%$\pm$1.4\% \\ \cline{3-7} 
 \hline
 \multicolumn{3}{|c|}{number of times that performs the best} &8 &11 &4&3 \\ \hline\hline
\multirow{6}{*}{\begin{tabular}[c]{@{}c@{}}Feature\\distribution\\skew\end{tabular}} 
& MNIST & \multirow{4}{*}{$\hat{\mb{x}} \sim Gau(0.1)$} & \tb{99.1\%$\pm$0.1\%} & \tb{99.1\%$\pm$0.1\%} & \tb{99.1\%$\pm$0.1\%}&\tb{99.1\%$\pm$0.1\%} \\ \cline{2-2} \cline{4-7} 
 & FMNIST &  & 89.1\%$\pm$0.3\% & 89.0\%$\pm$0.2\% & \tb{89.3\%$\pm$0.0\%} &89.0\%$\pm$0.1\% \\ \cline{2-2} \cline{4-7} 
 & CIFAR-10 &  & 68.9\%$\pm$0.3\% & 69.3\%$\pm$0.2\% & \tb{70.1\%$\pm$0.2\%}&68.5\%$\pm$1.3\% \\ \cline{2-2} \cline{4-7} 
 & SVHN &  & \tb{88.1\%$\pm$0.5\%} & \tb{88.1\%$\pm$0.2\%} & \tb{88.1\%$\pm$0.4\%}& \tb{88.1\%$\pm$0.4\%} \\ \cline{2-7} 
 & FCUBE & synthetic & \tb{99.8\%$\pm$0.2\%} & \tb{99.8\%$\pm$0.0\%} & 99.7\%$\pm$0.3\%&99.7\%$\pm$0.1\%  \\ \cline{2-7} 
 & FEMNIST & real-world & \tb{99.4\%$\pm$0.0\%} & 99.3\%$\pm$0.1\% & \tb{99.4\%$\pm$0.1\%}& 99.3\%$\pm$0.1\% \\ \hline
 \multicolumn{3}{|c|}{number of times that performs the best} &4 &3 &5&2 \\ \hline\hline
\multirow{7}{*}{\begin{tabular}[c]{@{}c@{}}Quantity\\skew\end{tabular}} & MNIST & \multirow{7}{*}{$q \sim Dir(0.5)$} & \tb{99.2\%$\pm$0.1\%} & \tb{99.2\%$\pm$0.1\%} & 99.1\%$\pm$0.1\%&99.1\%$\pm$0.1\% \\ \cline{2-2} \cline{4-7} 
 & FMNIST &  & 89.4\%$\pm$0.1\% & \tb{89.7\%$\pm$0.3\%} & 88.8\%$\pm$0.4\%&86.1\%$\pm$2.9\%\\ \cline{2-2} \cline{4-7} 
 & CIFAR-10 &  &\tb{72.0\%$\pm$0.3\%} & 71.2\%$\pm$0.6\% & 62.4\%$\pm$4.1\%&10.0\%$\pm$0.0\% \\ \cline{2-2} \cline{4-7} 
 & SVHN &  & 88.3\%$\pm$1.0\% & \tb{88.4\%$\pm$0.4\%} & 11.0\%$\pm$7.4\%&41.3\%$\pm$21.1\% \\ \cline{2-2} \cline{4-7} 
  & adult &  & 82.2\%$\pm$0.1\% & \tb{84.8\%$\pm$0.2\%} & 81.6\%$\pm$4.5\% & 43.2\%$\pm$33.9\% \\ \cline{2-2} \cline{4-7} 
& rcv1 &  & 96.7\%$\pm$0.3\% & \tb{96.8\%$\pm$0.4\%} & 49.0\%$\pm$1.9\%&51.8\%$\pm$0.7\% \\ \cline{2-2} \cline{4-7} 
& covtype &  & \tb{88.1\%$\pm$0.2\%} & 84.6\%$\pm$0.2\% & 63.2\%$\pm$20.8\%&51.2\%$\pm$3.2\% \\ \hline
 \multicolumn{3}{|c|}{number of times that performs the best} &3 &5 &0&0 \\ \hline\hline
 \multirow{9}{*}{\begin{tabular}[c]{@{}c@{}}Homogeneous\\partition\end{tabular}} & MNIST & \multirow{9}{*}{IID} & 99.1\%$\pm$0.1\% & 99.1\%$\pm$0.1\% & \tb{99.2\%$\pm$0.0\%} &99.1\%$\pm$0.1\%\\ \cline{2-2} \cline{4-7} 
 & FMNIST &  & 89.6\%$\pm$0.3\% & 89.5\%$\pm$0.2\% & \tb{89.7\%$\pm$0.2\%}&89.4\%$\pm$0.2\%  \\ \cline{2-2} \cline{4-7} 
 & CIFAR-10 &  & 70.4\%$\pm$0.2\% & 70.2\%$\pm$0.1\% & \tb{71.5\%$\pm$0.3\%}&69.5\%$\pm$1.0\%  \\ \cline{2-2} \cline{4-7} 
 & SVHN &  & \tb{88.5\%$\pm$0.5\%} & \tb{88.5\%$\pm$0.8\%} & 88.0\%$\pm$0.8\% &88.4\%$\pm$0.5\% \\\cline{2-2} \cline{4-7} 
  & FCUBE &  & 99.7\%$\pm$0.1\% & 99.6\%$\pm$0.2\% & 99.8\%$\pm$0.1\%&\tb{99.9\%$\pm$0.1\%}  \\ \cline{2-2} \cline{4-7}
  & FEMNIST & & 99.3\%$\pm$0.1\% & \tb{99.4\%$\pm$0.1\%} & \tb{99.4\%$\pm$0.0\%} &99.3\%$\pm$0.0\% \\ \cline{2-2} \cline{4-7}
  & adult & & 82.6\%$\pm$0.4\% & \tb{84.8\%$\pm$0.2\%} & 83.8\%$\pm$2.5\% &82.6\%$\pm$0.0\% \\ \cline{2-2} \cline{4-7}
  & rcv1 & & \tb{96.8\%$\pm$0.4\%} & 96.6\%$\pm$0.6\% & 80.9\%$\pm$27.8\% &96.6\%$\pm$0.4\% \\ \cline{2-2} \cline{4-7}
  & covtype & & 87.9\%$\pm$0.1\% & 85.2\%$\pm$0.0\% & \tb{88.0\%$\pm$2.3\%} &87.9\%$\pm$0.2\% \\ \hline
 \multicolumn{3}{|c|}{number of times that performs the best} &2 &3 &5&1 \\ \hline
\end{tabular}
}
\end{table*}

\subsection{Overall Accuracy Comparison}\label{subsec:overall}
The accuracy of existing approaches including FedAvg, FedProx, SCAFFOLD, and FedNova under different non-IID data settings is shown in Table \ref{tbl:perf}. \HBS{For comparison, we also present the results for IID scenarios (i.e., homogeneous partitions). } Next we show the insights from different perspectives.

\subsubsection{Comparison among different non-IID settings}\hfill\\
\noindent \fbox{\parbox{0.98\linewidth}{
\tb{Finding (1):} The label distribution skew case where each party only has samples of a single class is the most challenging setting, while the feature distribution skew and quantity skew setting have little influence on the accuracy of FedAvg.
}}

From Table \ref{tbl:perf}, we can observe that there is a gap between the accuracy of existing algorithms on several non-IID data settings and on the homogeneous setting. First, among different non-IID data settings, \HBS{all studied FL algorithms} perform worse on the label distribution skew case. Second, in label distribution skew setting, \HBS{the algorithms} have the worst accuracy when each party only has data from a single label. As expected, the accuracy increases as the number of classes in each party increases.
Third, for feature distribution skew setting, except for CIFAR-10, existing algorithms have a very close accuracy compared with the IID setting. Last, in quantity skew setting, FedAvg has almost no accuracy loss. Since the weighted averaging is adopted in FedAvg, it can already handle the quantity imbalance well. Overall, the label distribution skew influences the accuracy of FL algorithms most among all non-IID settings. There is room for existing algorithms to be improved to handle scenarios such as quantity-based label imbalance. 

\begin{figure}[!]
\centering
\includegraphics[width=0.8\columnwidth]{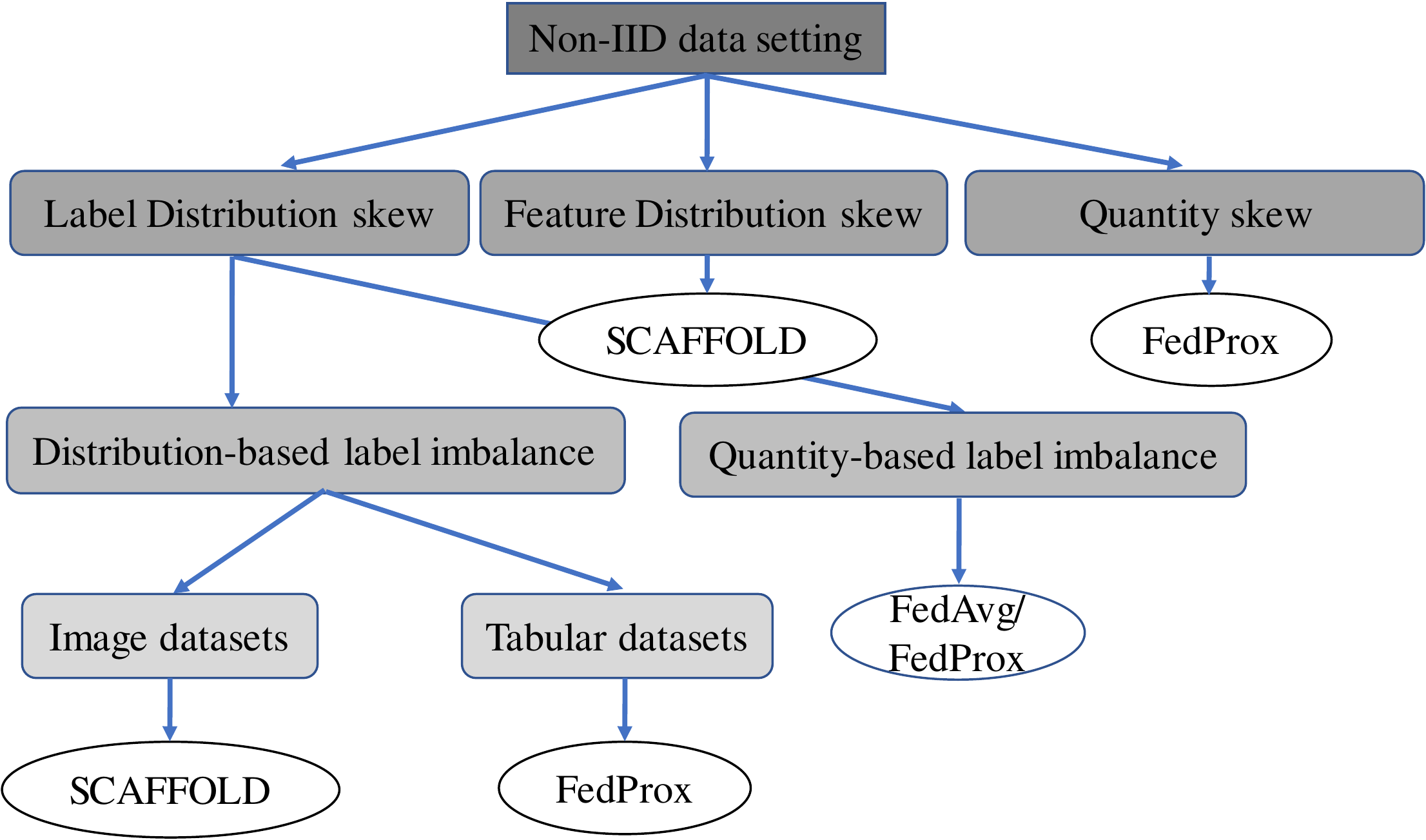}
\caption{\QB{The decision tree to determine the (almost) best FL algorithm given the non-IID setting.}}
\label{fig:tree_alg}
\end{figure}

\HBS{We draw a decision tree to summarize the suitable FL algorithm for each non-IID setting as shown in Figure \ref{fig:tree_alg} according to our observations. This decision tree is helpful for users to choose the algorithm for their learning according to the non-IID distribution and the datasets. For example, if the local datasets are likely to have feature distribution skew (e.g., the digits from different writers), then SCAFFOLD may be the best algorithm for FL. If the local datasets have almost the same data distribution but different sizes (e.g., databases with different capacities), then FedProx is likely the appropriate algorithm. If there is no prior knowledge on the local datasets, how to determine the distribution is a challenging problem and more research efforts are needed (see Section \ref{sec:oppo_data})}.

\subsubsection{Comparison among different algorithms}\hfill\\
\noindent \fbox{\parbox{0.98\linewidth}{
\tb{Finding (2):} No algorithm consistently outperforms the other algorithms in all settings. The state-of-the-art algorithms significantly outperform FedAvg only in several cases. 
}}

We have the following observations in aspect of different algorithms. First, in label distribution skew and quantity skew cases, FedProx usually achieves the best accuracy. In feature distribution skew case, SCAFFOLD usually achieves the best accuracy. Second, in some cases (e.g., $p_k \sim Dir(0.5)$, feature distribution skew and quantity skew), the improvement \HBS{of the three non-IID FL algorithms} is insignificant compared with FedAvg, which is smaller than 1\%. Third, when $\#C=1$, FedProx can significantly outperform FedAvg, SCAFFOLD and FedNova. Fourth, for SCAFFOLD, its accuracy is quite unstable. It can significantly outperform the other two approaches in some cases (e.g., $Dir(0.5)$ and $K=1$ on CIFAR-10). However, it may also have much worse accuracy than the other two approaches (e.g., $K=1$ and $K=2$ on SVHN). Last, for FedNova, it does not show much superiority compared with other FL algorithms. Compared with the accuracy of FedAvg on the homogeneous partition, there is still a lot of room for improvement in the non-IID setting.

\subsubsection{Comparison among different tasks}\hfill\\
\noindent \fbox{\parbox{0.98\linewidth}{
\tb{Finding (3):} CIFAR-10 and tabular datasets are challenging tasks under non-IID settings. MNIST is a simple task under most non-IID settings where the studied algorithms perform similarly well.
}}

Among nine different datasets, while heterogeneity significantly degrades the accuracy of FL algorithms on CIFAR-10 and tabular datasets, such influence is smaller in other datasets. Among image datasets, the classification task on CIFAR-10 is more complex than the other datasets in a centralized setting. Thus, when each party only has a skewed subset, the task will be more challenging and the accuracy is worse. Also, it is interesting that all the four algorithms cannot handle tabular datasets well in the non-IID setting. The accuracy loss is quite large especially for the label distribution skew case.  \HBS{We suggest that the challenging tasks like CIFAR-10 and rcv1 should be included in the benchmark for distributed data silos.}

\subsection{Communication Efficiency}
\noindent \fbox{\parbox{0.98\linewidth}{
\tb{Finding (4):} FedProx has almost the same convergence speed compared with FedAvg, while SCAFFOLD and FedNova are more unstable in training.
}}


\begin{figure}[t]
    \centering
    \subfloat[$p_k \sim Dir(0.5)$]{\includegraphics[width=0.48\columnwidth]{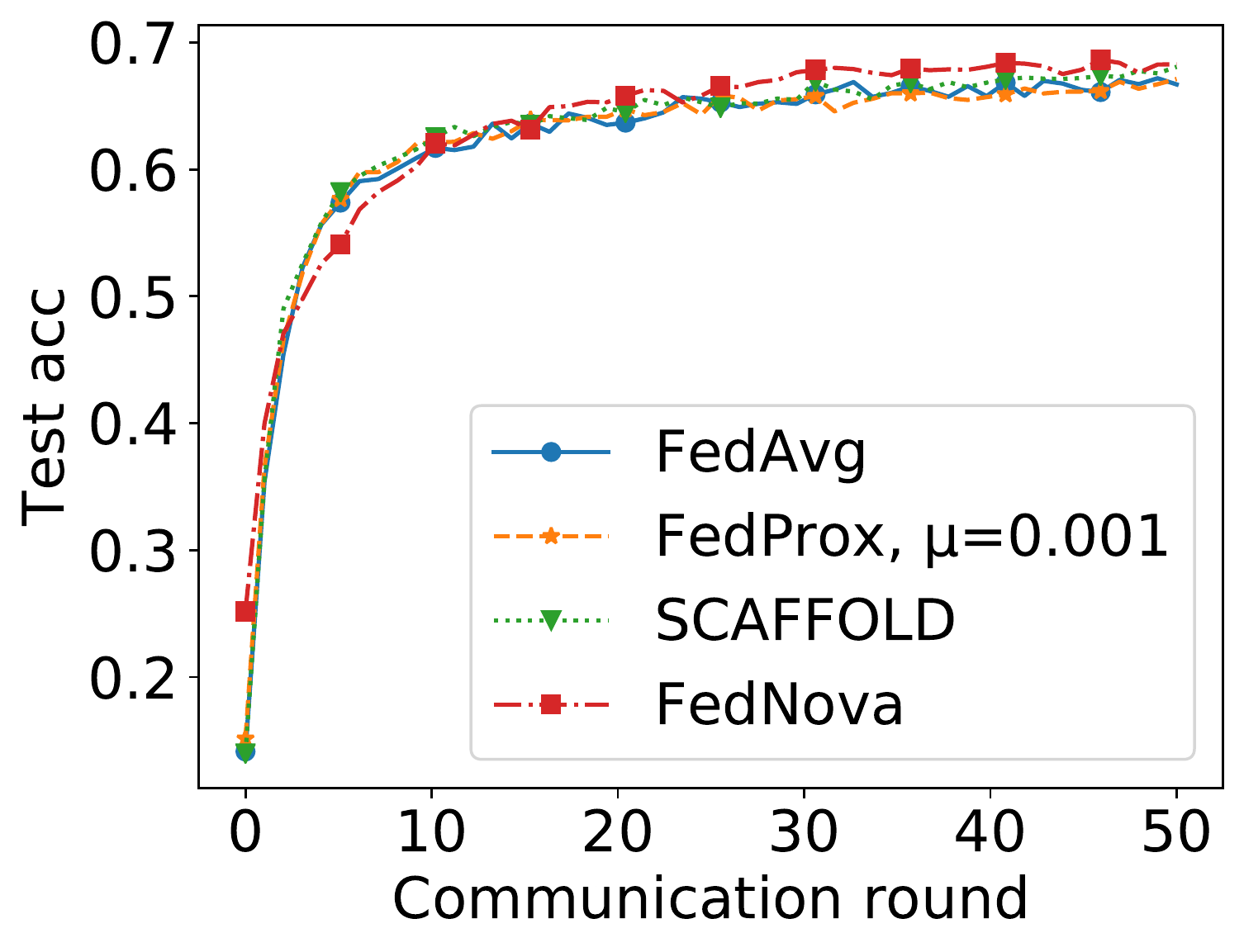}}
    \subfloat[$\hat{\mb{x}} \sim Gau(0.1)$]{\includegraphics[width=0.48\columnwidth]{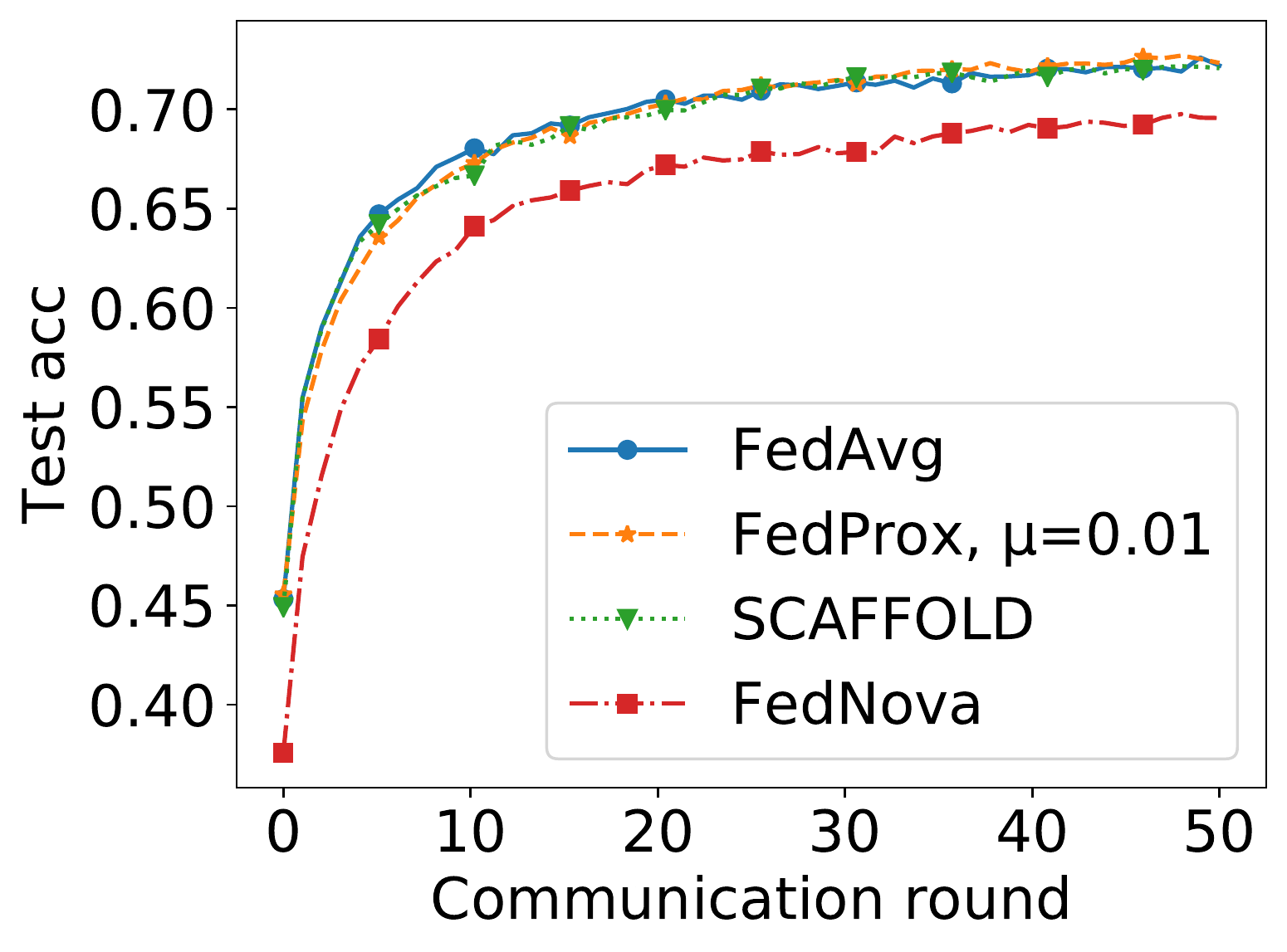}}
    \caption{The training curves of different approaches on CIFAR-10.}
    \label{fig:cifar10_curve}
\end{figure}


Figure \ref{fig:cifar10_curve} shows the training curves of the studied algorithms on CIFAR-10. Here we try two different partitioning strategies that cover label skew and feature skew. For the results on other partitioning strategies and other datasets, please refer to Appendix A of our technical report~\cite{li2021federated}. For FedProx, we show the curve with the best $\mu$. First, for the $\#C=1$ setting, FedAvg and FedProx are very unstable, while SCAFFOLD and FedNova even cannot improve as the number of rounds increases. Second, for the $q \sim Dir(0.5)$ setting, FedNova is quite unstable and the accuracy changes rapidly as the number of communication rounds increases. Moreover, FedProx is very close to FedAvg during the whole training process in many cases. Since the best $\mu$ is always small, the regularization term in FedProx has little influence on the training. Thus, FedProx and FedAvg usually have similar convergence speed and final accuracy. How to achieve stable learning and fast convergence is still an open problem on non-IID data.

\subsection{Robustness to Local Updates}
\noindent \fbox{\parbox{0.98\linewidth}{
\tb{Finding (5):} The number of local epochs can have a large effect on the accuracy of existing algorithms. The optimal value of the number of local epochs is \HBS{very sensitive to non-IID distributions.}
}}

\begin{figure}[!]
    \centering
    \subfloat[$p_k \sim Dir(0.5)$]{\includegraphics[width=0.48\columnwidth]{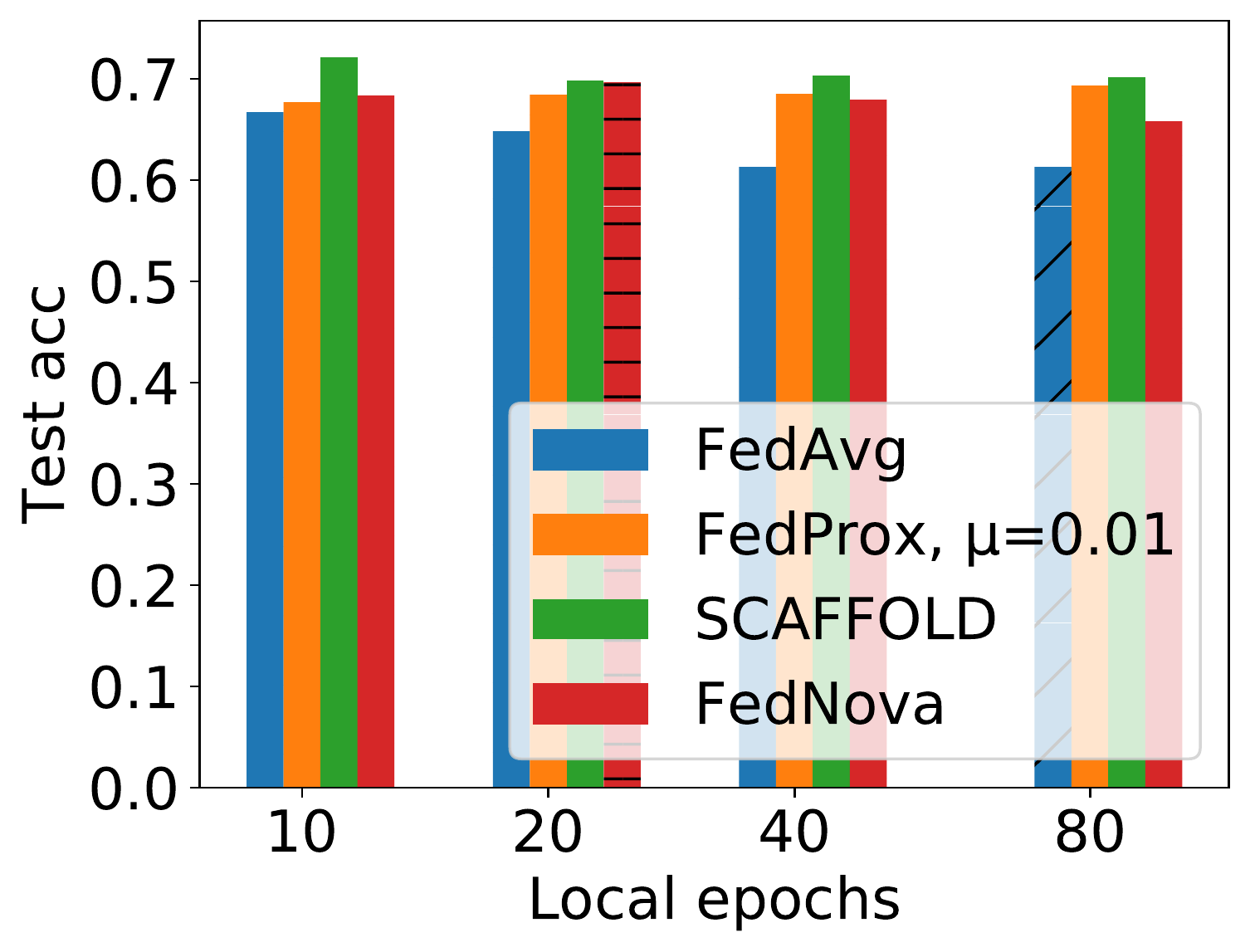}}
    \subfloat[$\hat{\mb{x}} \sim Gau(0.1)$]{\includegraphics[width=0.48\columnwidth]{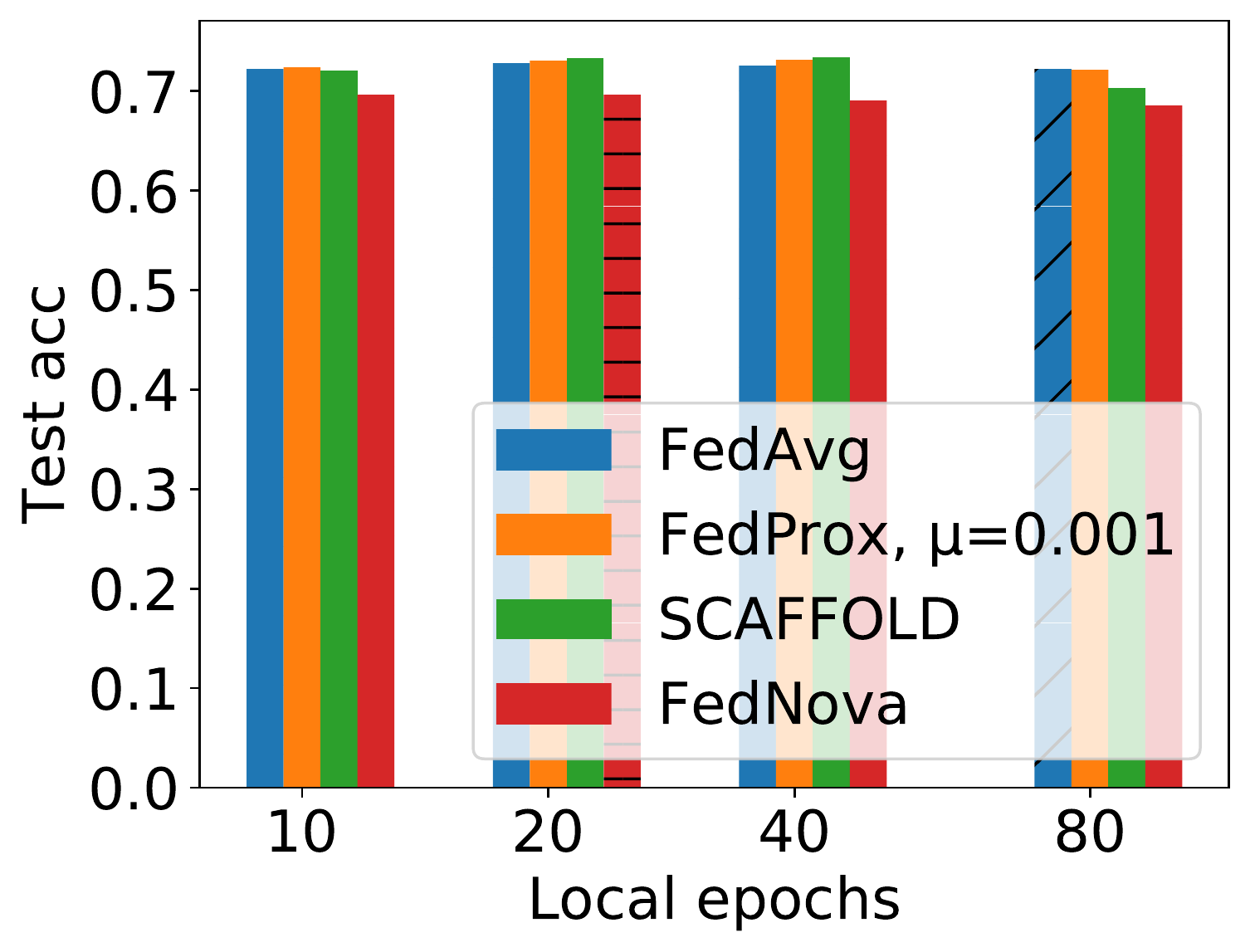}}
    \caption{The test accuracy with different numbers of local epochs on CIFAR-10.}
    \label{fig:cifar10_epoch}
\end{figure}

We vary the number of local epochs from $\{10, 20, 40, 80\}$ and report the final accuracy on CIFAR-10 in Figure \ref{fig:cifar10_epoch}. Please refer to Appendix B of our technical report~\cite{li2021federated} for the results of other settings and datasets. On the one hand, we can find that the number of local epochs has a large effect on the accuracy of FL algorithms. For example, when $\#C=2$, the accuracy of all algorithms generally degrades significantly when the number of local epochs is set to 80. On the other hand, the optimal number of local epochs differ in different settings. For example, when $\#C=1$ and $\#C=2$, the optimal number of local epochs is 20 for FedAvg, and is 10 on the settings $p_k \sim Dir(0.5)$ and $\#C=3$. In summary, existing algorithms are not robust enough against large local updates. \HBS{Non-IID distributions have to be considered to determine the best number of local epochs.}

\begin{figure}[t]
    \centering
    \subfloat[$p_k \sim Dir(0.5)$]{\includegraphics[width=0.48\columnwidth]{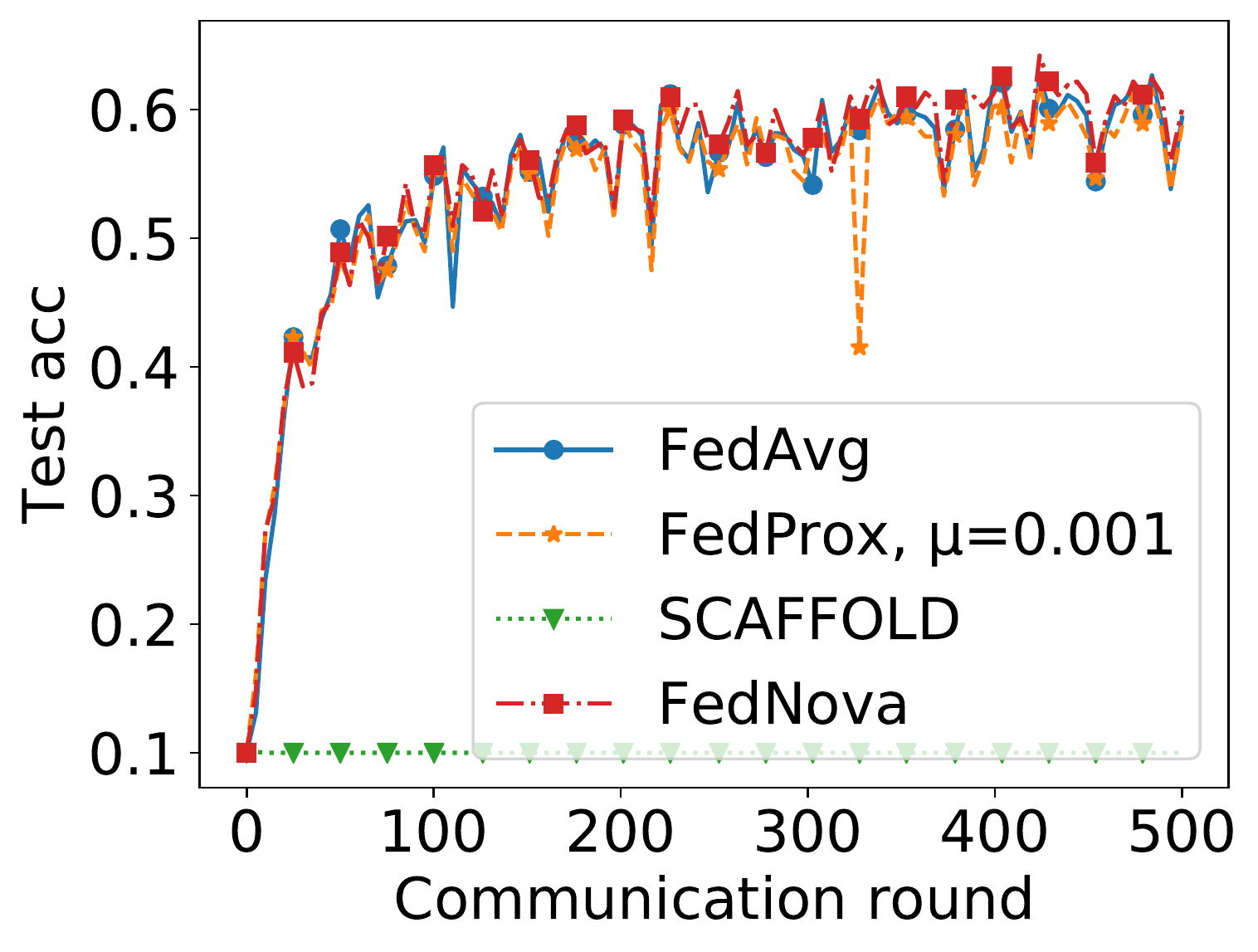}}
    \subfloat[$q \sim Dir(0.5)$]{\includegraphics[width=0.48\columnwidth]{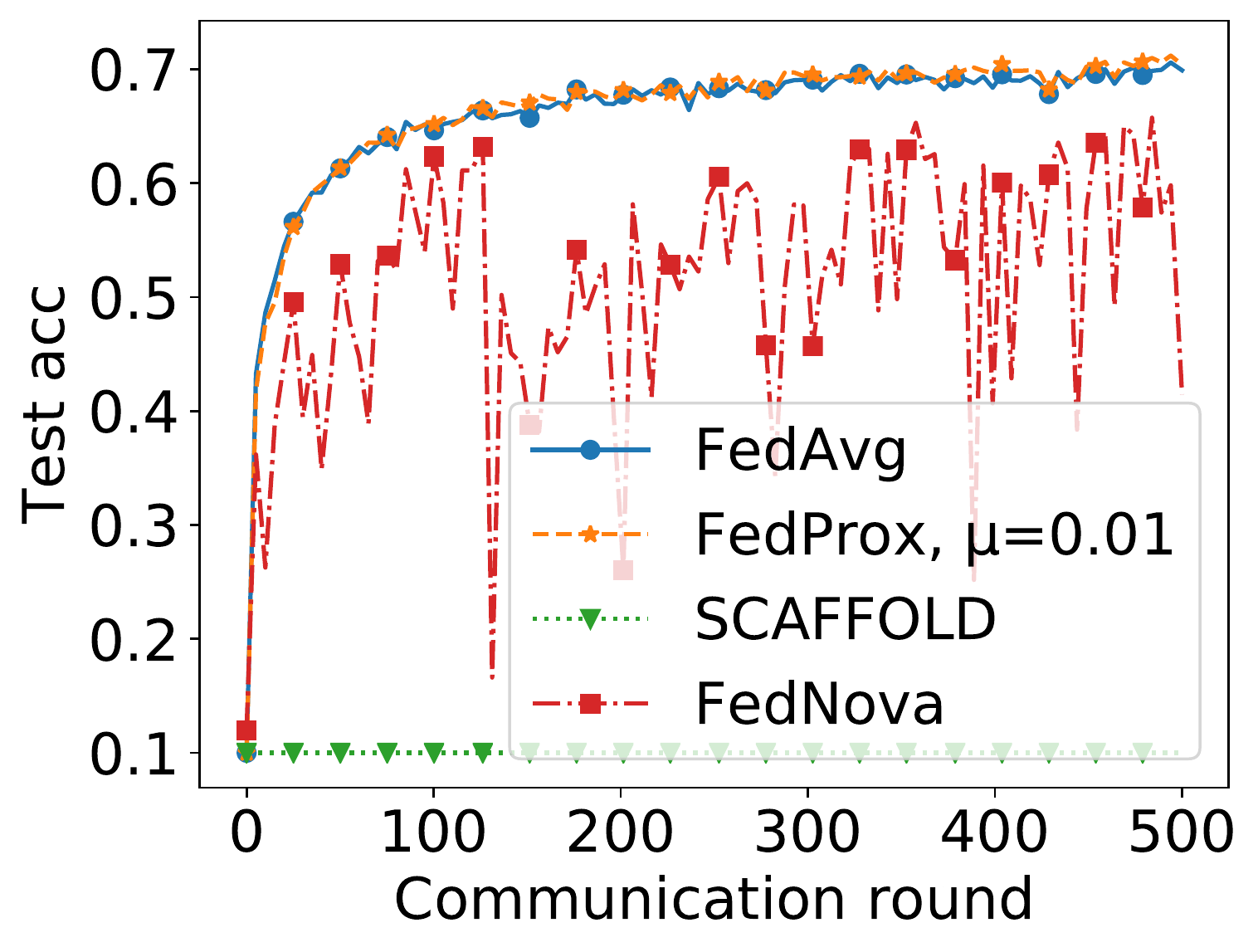}}
    \caption{The training curves of different approaches on CIFAR-10 with 100 parties and sample fraction 0.1.}
    \label{fig:100_parties}
\end{figure}

\subsection{Party Sampling}
\label{sec:scala}
\noindent \fbox{\parbox{0.98\linewidth}{
\tb{Finding (6):} In the partial participation setting, SCAFFOLD cannot work effectively, while the other FL algorithms have a very unstable accuracy during training.
}}

\HBS{In some scenarios, not all the data silos will participate the entire training process. In such a setting, the sampling technique is usually applied (Line 6 of Algorithm \ref{alg:fed}). To simulate this scenario, we set the number of parties to 100 and the sample fraction to 0.1.} We run experiments on CIFAR-10 and the results are shown in Figure \ref{fig:100_parties}. Please refer to Appendix C of the technical report \cite{li2021federated} for the results with other partitioning strategies. We can find that the training curves are quite unstable in most non-IID settings. Due to the sampling technique, the local distributions among different rounds can vary, and thus the averaged gradients may have very different directions among rounds. Moreover, we can find that SCAFFOLD has a bad accuracy on all settings. Since the frequency of updating local control variates (Lines 23-25 of Algorithm \ref{alg:scaffold}) is low, the estimation of the update direction may be very inaccurate using the control variates.

\subsection{\rev{Scalability}}
\noindent \fbox{\parbox{0.98\linewidth}{
\rev{\tb{Finding (7):} The accuracy of all approaches decrease when increasing the number of parties.}
}}

\begin{figure}[!]
    \centering
    \subfloat[$p_k \sim Dir(0.5)$]{\includegraphics[width=0.48\columnwidth]{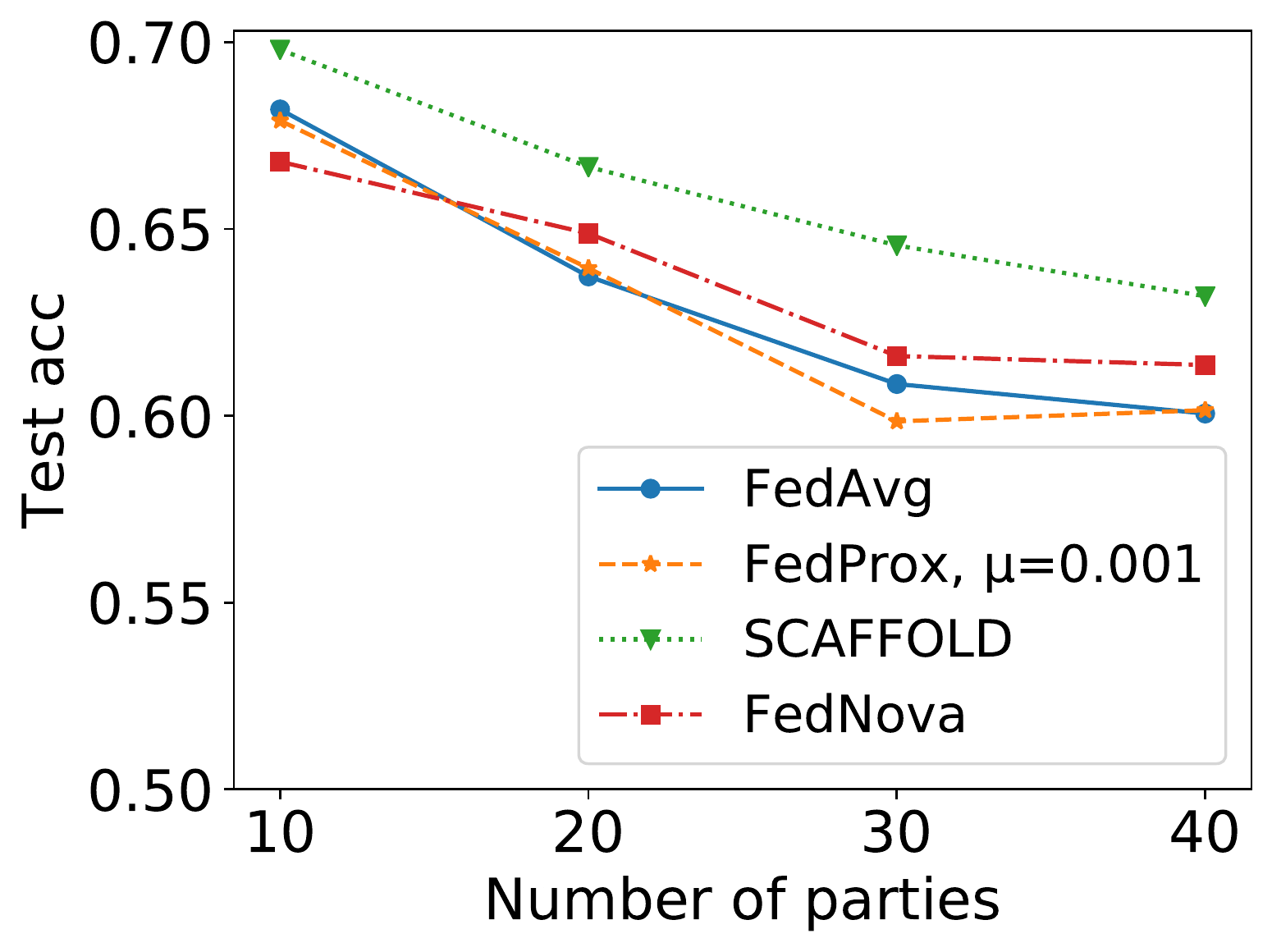}}
    \subfloat[$\hat{\mb{x}} \sim Gau(0.1)$]{\includegraphics[width=0.48\columnwidth]{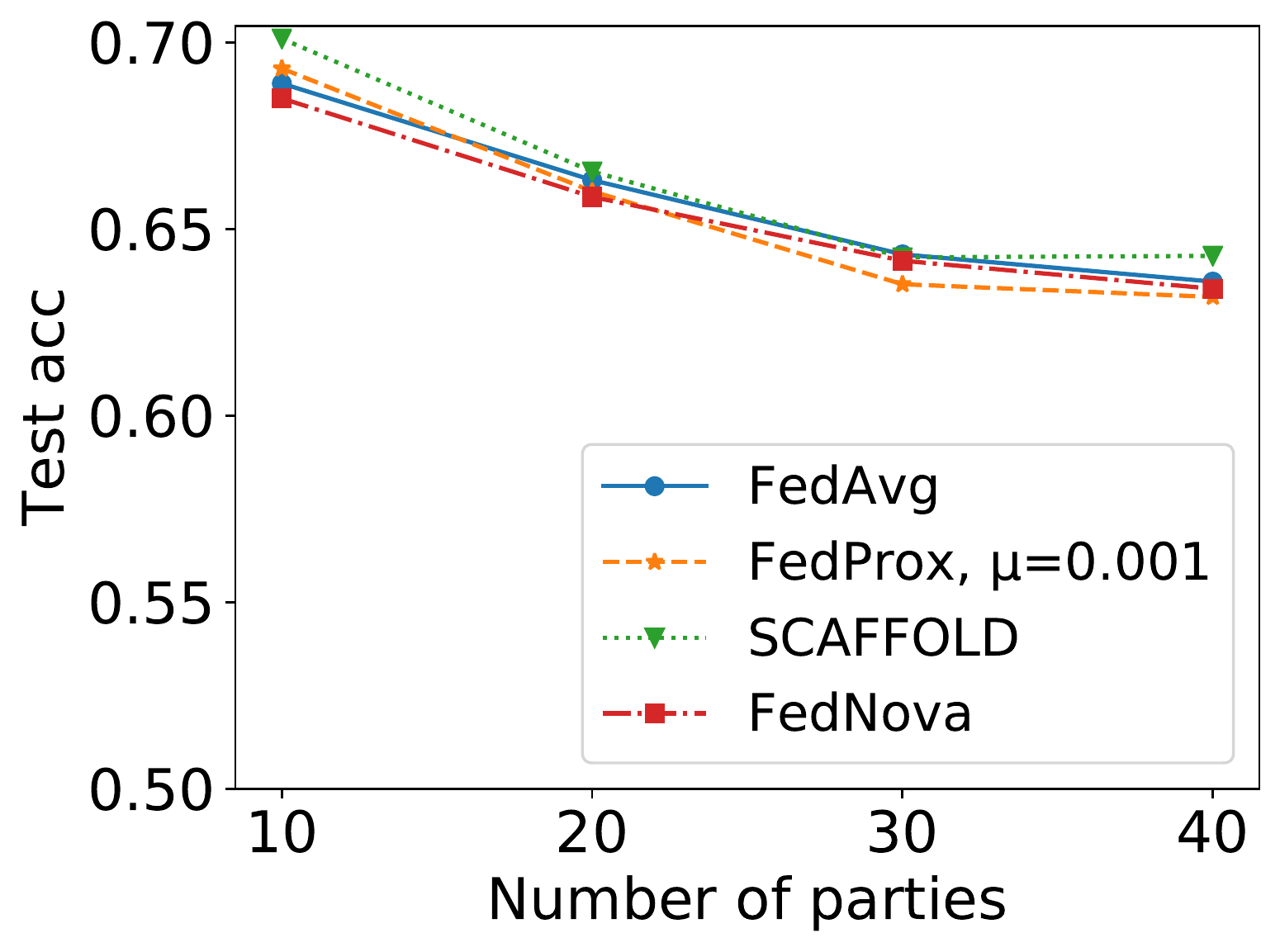}}
    \caption{\rev{The test accuracy with different number of parties on CIFAR-10.}}
    \label{fig:cifar10_client_number}
\end{figure}

\rev{We study the effect of number of clients on studied approaches as shown in Figure \ref{fig:cifar10_client_number}. Here we run all approaches for 50 rounds. We can observe that the accuracy decreases significantly when increasing the number of clients. When the number of parties is large, the amount of local data is small and it is easy to overfit in the local training stage. How to design effective and communication-efficient algorithms on a large-scale setting with small data in the client is still an open problem.
}

\subsection{\rev{Efficiency}}
\label{sec:efficiency}
\noindent \fbox{\parbox{0.98\linewidth}{
\rev{\tb{Finding (8):} The computation overhead of FedProx is large compared with FedAvg. Moreover, the communication cost of SCAFFOLD is twice of that of FedAvg.}
}}

\rev{To compare the efficiency of different FL algorithms, we show the overall computation time and communication costs of each approach in Table \ref{tab:efficiency}. We can observe that the computation costs of FedAvg, SCAFFOLD, and FedNova are close. FedProx has a much higher computation cost than the other algorithms. From Algorithm \ref{alg:fed}, FedProx directly modifies the objective, which causes additional computation overhead in the gradient descent of each batch. FedNova and SCAFFOLD only introduce very small number of addition and multiplication operations each round, which is negligible. For the communication costs, since SCAFFOLD needs to communicate control variates in each round as shown in Algorithm \ref{alg:scaffold}, its communication cost is twice of that of the other algorithms.}

\begin{table}[]
\centering
\caption{\rev{The computation time (second) and communication size (MB) per round of different approaches.}}
\label{tab:efficiency}
\resizebox{\columnwidth}{!}{%
\begin{tabular}{|c|c|c|c|c|}
\hline
         & MNIST  & CIFAR-10 & adult & rcv1    \\ \hline
FedAvg   & 73s     & 193s     & 15s    & 66s      \\ \hline
FedProx  & 133s    & 233s     & 44s    & 76s      \\ \hline
SCAFFOLD & 77s     & 197s     & 14s    & 66s      \\ \hline
FedNova  & 73s     & 189s     & 17s    & 65s      \\ \hline\hline
FedAvg   & 1.95MB & 2.73MB   & 0.20MB & 66.54MB  \\ \hline
FedProx  & 1.95MB  & 2.73MB   & 0.20MB & 66.54MB  \\ \hline
SCAFFOLD & 3.91MB  & 5.46MB   & 0.41MB & 133.08MB \\ \hline
FedNova  & 1.95MB  & 2.73MB   & 0.20MB & 66.54MB  \\ \hline
\end{tabular}%
}
\vspace{-10pt}
\end{table}

\subsection{\rev{Mixed Types of Skew}}
\label{sec:mixed_type}
\noindent \fbox{\parbox{0.98\linewidth}{
\rev{\tb{Finding (9):} FL is more challenging when there exists mixed types of skew among the local data.
}}}

\rev{In practice, there may exist mixed types of skew among parties. Here we combine multiple partitioning strategies to generate such cases. We try two different settings: 1) we first divide the whole dataset into each party by the distribution-based label imbalanced partitioning strategy. Then, we add noises to the data of each party according to the noise-based feature imbalance strategy. Therefore, there exists both label distribution skew and feature distribution skew among the local data of different parties. 2) we first divide the whole dataset into each party by the quantity imbalanced partitioning strategy. Then, we add noises to the data of each party according to the noise-based feature imbalance strategy. Therefore, there exists both feature distribution skew and quantity skew among the local data of different parties. The results are shown in Table \ref{tbl:mixed_skew}.} 

\rev{For the first case, we can observe that the accuracies of all approaches degrade when there exists mixed types of skew compared with a single type of skew, which is reasonable since both label imbalance and feature imbalance bring challenges in the training process.} 

\rev{For the second case, while quantity skew does not affect the accuracy of FedAvg and FedProx, the accuracy of both feature and quantity skew setting is close to the accuracy of the feature skew setting. However, for SCAFFOLD and FedNova, the accuracy of both feature and quantity skew setting is poor since quantity skew degrades the accuracy significantly.}

\rev{Overall, as we observe more significant model quality degradation in mixed non-IID settings, it is an important direction to design algorithms for settings with mixed types of skew, which are common in reality. For example, the images taken in different areas have different label distributions, while the feature distributions also differ due to the cameras (e.g., contrast).}



\begin{table}[]
\centering
\caption{\rev{The performance of different approaches with different imbalance cases on CIFAR-10.}}
\label{tbl:mixed_skew}
\resizebox{0.48\textwidth}{!}{%
\begin{tabular}{|c|c|c|c|c|}
\hline
          Case 1                  & FedAvg & FedProx & SCAFFOLD & FedNova \\ \hline
label skew             & 68.2\% & 67.9\%  & 69.8\%   & 66.8\%  \\ \hline
feature skew           & 68.9\% & 69.3\%  & 70.1\%   & 68.5\%  \\ \hline
label and feature skew & 66.1\% & 64.8\%  & 67.8\%   & 65.9\%  \\ \hline \hline
          Case 2                  & FedAvg & FedProx & SCAFFOLD & FedNova \\ \hline
feature skew           & 68.9\% & 69.3\%  & 70.1\%   & 68.5\%  \\ \hline
quantity skew           & 72.0\% &	71.2\% &	62.4\%	& 10.0\%  \\ \hline
feature and quantity skew & 69.1\% & 69.2\% &	62.2\% & 10.0\% \\ \hline
\end{tabular}%
}
\vspace{-10pt}
\end{table}
\subsection{\rev{Insights on the Experimental Results}}

\rev{We summarized the insights from the experimental studies as follows.}
\begin{itemize}
    \item \rev{The design and evaluation of future FL algorithms should consider more comprehensive settings, including different non-IID data partitioning strategies and tasks. There is not a single studied algorithm that consistently outperforms the other algorithms or has a good performance in all settings. Thus, it is still a promising research direction to address issues in distributed data silos with FL.}
    \item \rev{Accuracy and communication efficiency are two important metrics in the evaluation of FL algorithms under non-IID data settings. Our study demonstrates the trade-off between them, and also the stability of those two metrics in the training process.}
    \item \rev{FL introduces new training factors (e.g., number of local epochs, batch normalization, party sampling, number of parties) compared with centralized training due to non-IID data setting, while some training factors share the similar behavior as the centralized training (e.g., batch size). These challenging factors deserve more attention in the evaluation of future FL studies.}
    \item \rev{Mixed types of skew brings more challenges than a single type of skew. As we observe more significant model quality degradation in mixed non-IID settings, it is important to investigate effective algorithms working on multiple types of skew, which is more practical in reality.}
\end{itemize}





\section{Future Directions}
\label{sec:future_dir}

\HBS{We present some following promising future directions for data management and federated learning on non-IID distributed databases. }

\subsection{Opportunities for data management}
\label{sec:oppo_data}
\noindent\textbf{Integration with learned database systems:} Existing learned systems are mostly based on centralized databases, such as learned index structures~\cite{10.14778/3421424.3421425, 10.1145/3318464.3389711} and learned cost estimation~\cite{10.14778/3342263.3342644, 10.1145/3318464.3389741}. We believe that, as the concerns on data privacy and data regulation grow, we will see more distributed databases and existing learned systems and algorithms need to be revisited. For example, it could be very interesting to enable federated search and develop learned index structures for multiple ``data silos'' without exchanging the local data.

\noindent\textbf{Light-weight data techniques for profiling non-IID data:} From our experimental study, different non-IID distributions have a large effect on the accuracy and stability of FL algorithms. Thus, it would be helpful if we can know the non-IID distribution in prior before conducting FL. This made a decade of database research relevant, such as data sampling \cite{chaudhuri1998random} and sketching \cite{gilbert2002fast}. Another potential approach is to use mete data to represent the non-IID distributions. However, it is still an open problem on how to extend current statistics estimation (such as cardinality estimation) to non-IID distribution.

\noindent\textbf{Non-IID resistant sampling for partial participation:} As in Finding (8), the sampling approach can bring instability in FL. Instead of random sampling, selective sampling according to the data distribution features of the parties may significantly increase the learning stability. One inspiration is from the skew resistant data techniques~\cite{10.1145/233269.233340, 10.1145/1807128.1807140}, which can be potentially extended to the partial participation in FL training. \rev{Moreover, stratified sampling \cite{neyman1992two} can be a good solution. By classifying the parties to subgroups, representative parties can be selected in each round in a more balanced way \cite{abdulrahman2020fedmccs}.}

\noindent\textbf{Privacy-preserving data mining:} Although there is no raw data transfer in FL, the model may still leak sensitive information about the training data due to possible inference attacks \cite{shokri2017membership,fredrikson2015model}. Thus, techniques such as differential privacy \cite{dwork2011differential} are useful to protect the local databases. How to decrease the accuracy loss while ensuring the differential privacy guarantee is a challenge research direction.

\noindent\textbf{Query on Federated Databases:} As we focus on distributed databases due to privacy concerns, federated databases \cite{sheth1990federated} also need to be revisited. On the one hand, how to combine the SQL query with machine learning on federated databases is an important problem. On the other hand, how to preserve the data privacy while supporting both query and learning on federated databases also needs to be investigated.

\subsection{Opportunities for better FL design}
\noindent\textbf{A Party with a Single Label:} From Table \ref{tbl:perf}, the accuracy of FL algorithms is very bad if each party only has data of a single label. This setting is seemingly unrealistic. However, it has many real-world applications in practice. For example, we can use FL to train a speaker recognition model, while each mobile device only has the voices of its single user. 



\noindent\textbf{Fast Training:} From Figure \ref{fig:cifar10_curve}, the training speed of existing FL algorithms are usually close to each other. FedProx, SCAFFOLD, and FedNova do not show much superiority on the communication efficiency. To improve the training speed, researchers can work on the following two directions. One possible solution is to develop communication-efficient FL algorithms with only a few rounds. There are some studies \cite{li2020model,guha2019one} that propose FL algorithms using a single communication round. In their studies, a public dataset is needed, which may potentially limit the applications. Another possible solution is to develop fast initialization approach to reduce the number of rounds while achieving the same accuracy for FL. In the experiments of a previous study \cite{li2020model}, they show that their approach is also promising if applied as an initialization step.

\noindent\textbf{Automated Parameter Tuning for FL:} FL algorithms suffer from large local updates. The number of local epochs is an important parameter in FL. While one traditional way is to develop approaches robustness to the local updates, another way is to design efficient parameter tuning approaches for FL. A previous paper \cite{dai2020federated} studies Bayesian optimization in the federated setting, which can be used to search hyper-parameters. Approaches for the setting of number of local epochs need to be investigated.


\noindent\textbf{Towards Robust Algorithms against Different Non-IID Settings:} As in Finding (2), no algorithm consistently performs the best in all settings. It is a natural question whether and how we can develop a robust algorithm for different non-IID settings. We may have to first investigate the common characteristics of FL processes under different non-IID settings. The intuitions of existing algorithms are same: the local model updates towards the local optima, and the averaged model is far from the global optima. We believe the design of FL algorithms under non-IID settings can be improved if we can observe more detailed and common behaviours in the training.

\noindent\textbf{Aggregation of Heterogeneous Batch Normalization:} From our Finding (7), simple averaging is not a good choice for batch normalization. Since the batch normalization in each party records the statistics of local data distribution, there is also heterogeneity among the batch normalization layers of different parties. The averaged batch normalization layer may not catch the local distribution after sending back to the parties. A possible solution is to only average the learned parameters but leave the statistics (i.e., mean and variance) alone \cite{andreux2020siloed}. More specialized designs for particular layers in deep learning need to be investigated.

\section{Related Work}
\label{sec:related_work}

\rev{Although the existing study \cite{kairouz2019advances} provides non-IID data cases, it does not provide the partitioning strategies to generate the corresponding non-IID data distributions. We go beyond the previous study and summarize six different partitioning strategies to generate three non-IID data distribution cases. Among these six partitioning strategies, the two partitioning strategies in Section IV-A-b and Section IV-B-c are adopted from existing FL studies due to their popularity, while the other four effective partitioning strategies are designed by our study. Next, we introduce these partitioning strategies in detail.}

There are some existing benchmarks for federated learning \cite{caldas2018leaf,hu2020oarf,he2020fedml,liang2020isolated}. LEAF \cite{caldas2018leaf} provides some realistic federated datasets including images and texts. Specifically, LEAF partitions the existing datasets according to its data recourses, e.g., partitioning the data in Extended MNIST \cite{cohen2017emnist} based on the writer of the digit or character. OARF \cite{hu2020oarf} proposes federated datasets by combining multiple related real-world public datasets. Moreover, it provides various metrics including utility, communication overhead, privacy loss, and mimics the federated systems in the real world. However, both LEAF and OARF do not provide an algorithm-level comparison. FedML \cite{he2020fedml} provides reference implementations of federated learning algorithms such as FedAvg, FedNOVA \cite{wang2020tackling} and FedOpt \cite{reddi2020adaptive}. There are no new datasets, metrics, and settings in FedML. FLBench \cite{liang2020isolated} is proposed for isolated data island scenario. Its framework covers domains including medical, finance, and AIoT. However, currently, FLBench is not open-sourced and it does not provide any experiments.

The above benchmarks do not provide analysis of existing federated learning algorithms on different non-IID settings, which is our focus in this paper. To the best of our knowledge, there is one existing benchmark \cite{liu2020evaluation} for federated learning on the non-IID data setting. However, it only provides two partitioning approaches: random split and split by labels. In this paper, we provide comprehensive partitioning strategies and datasets to cover different non-IID settings. \rev{Moreover, we conduct extensive experiments to compare and analyze existing federated learning algorithms.}

\section{Conclusion}
\label{sec:con}

\HBS{There has been a growing interest in exploiting distributed databases (e.g., in different organizations and countries) to improve the effectiveness of machine learning services. In this paper, we study non-IID data as one key challenge in such distributed databases, and develop a benchmark named NIID-bench. Specifically, we introduce six data partitioning strategies which are much more comprehensive than the previous studies. Furthermore, we conduct comprehensive experiments to compare existing algorithms and demonstrate their strength and weakness. This study sheds light on some future directions to build effective machine learning services on distributed databases. }

\bibliographystyle{abbrv}
\bibliography{reference}

\appendix

\subsection{Training Curves}
\label{app:curve}
\begin{figure*}[t]
    \centering
    \subfloat[$\#C=1$]{\includegraphics[width=0.24\textwidth]{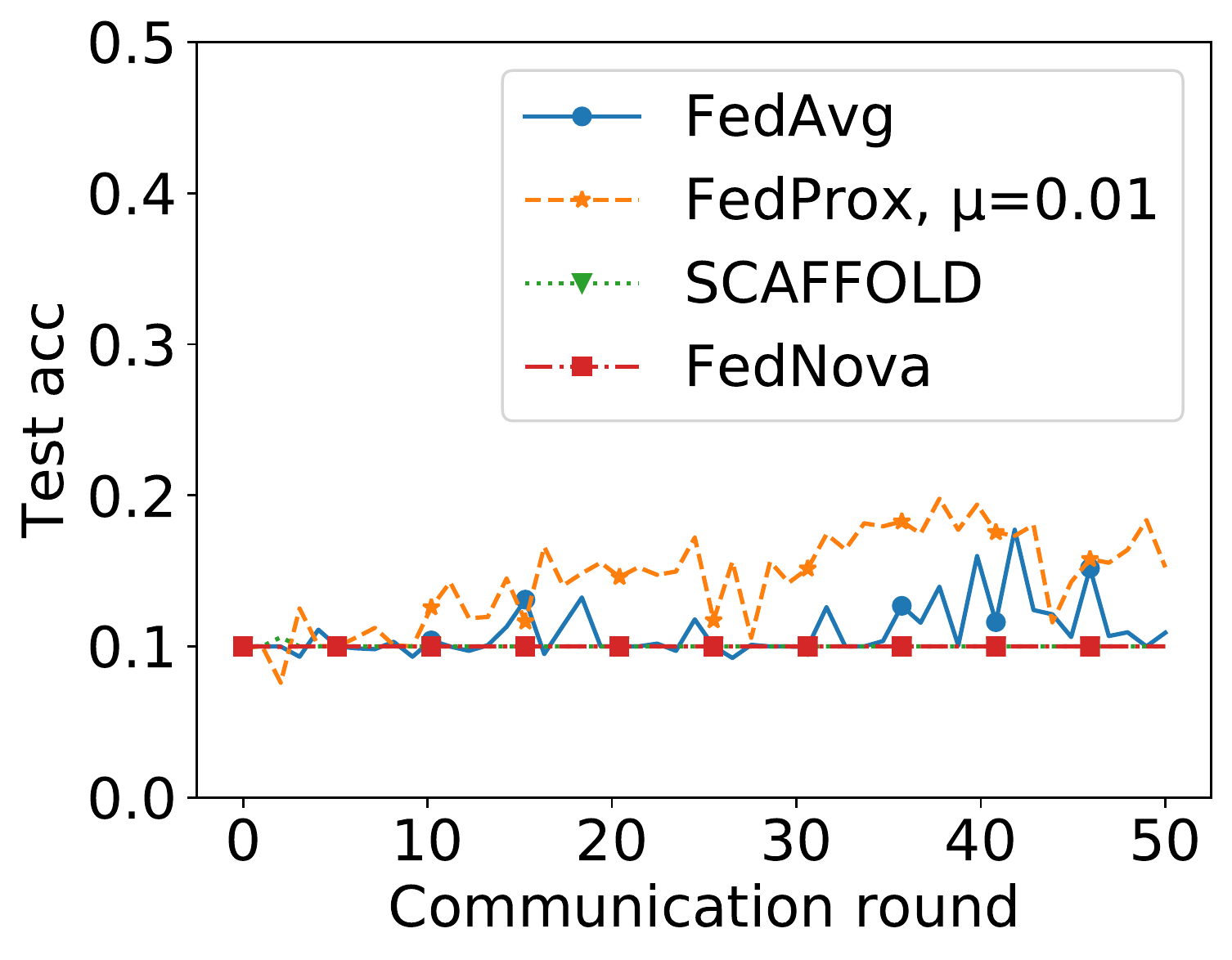}}
    \subfloat[$\#C=2$]{\includegraphics[width=0.24\textwidth]{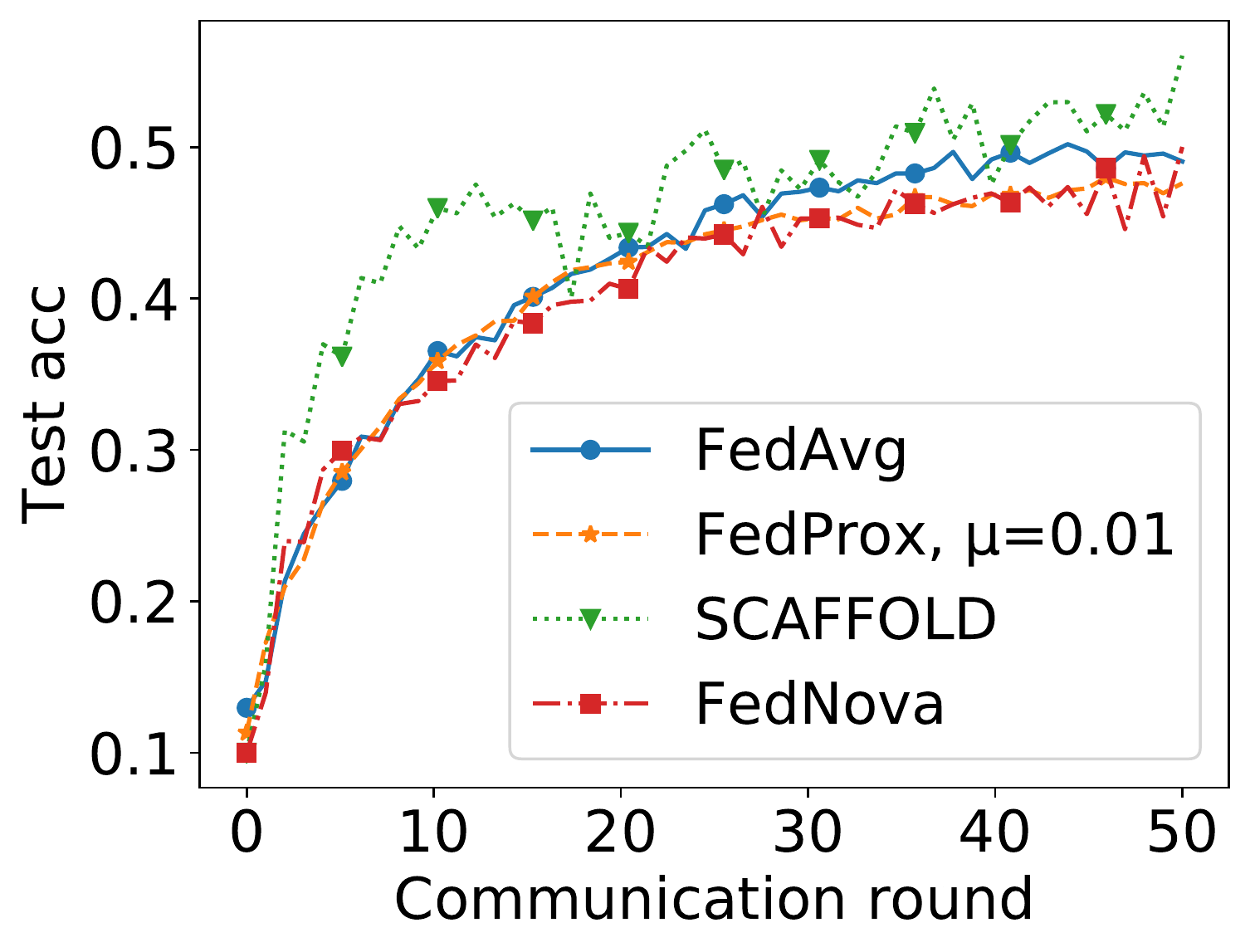}}
    \subfloat[$\#C=3$]{\includegraphics[width=0.24\textwidth]{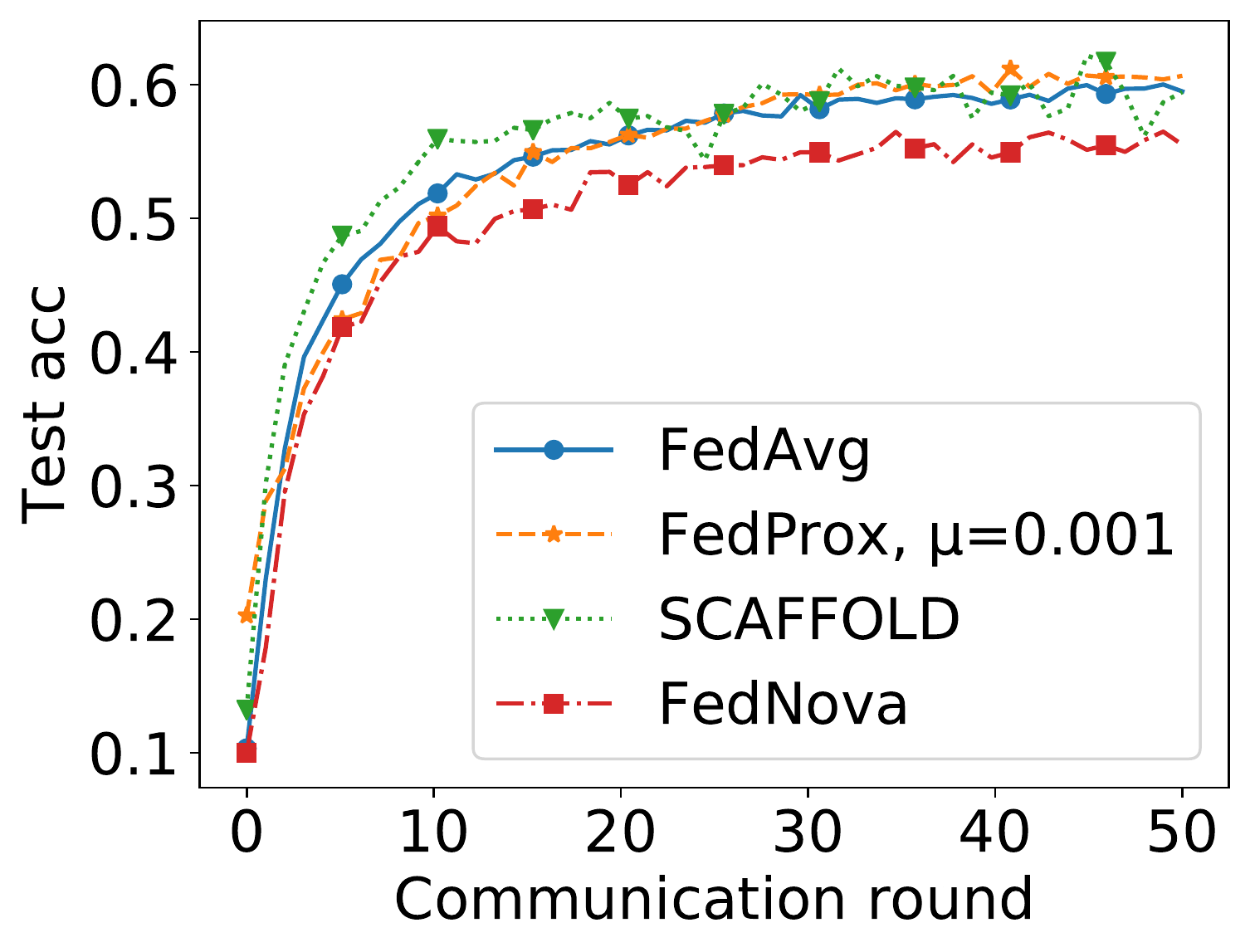}}
    \subfloat[$q \sim Dir(0.5)$]{\includegraphics[width=0.24\textwidth]{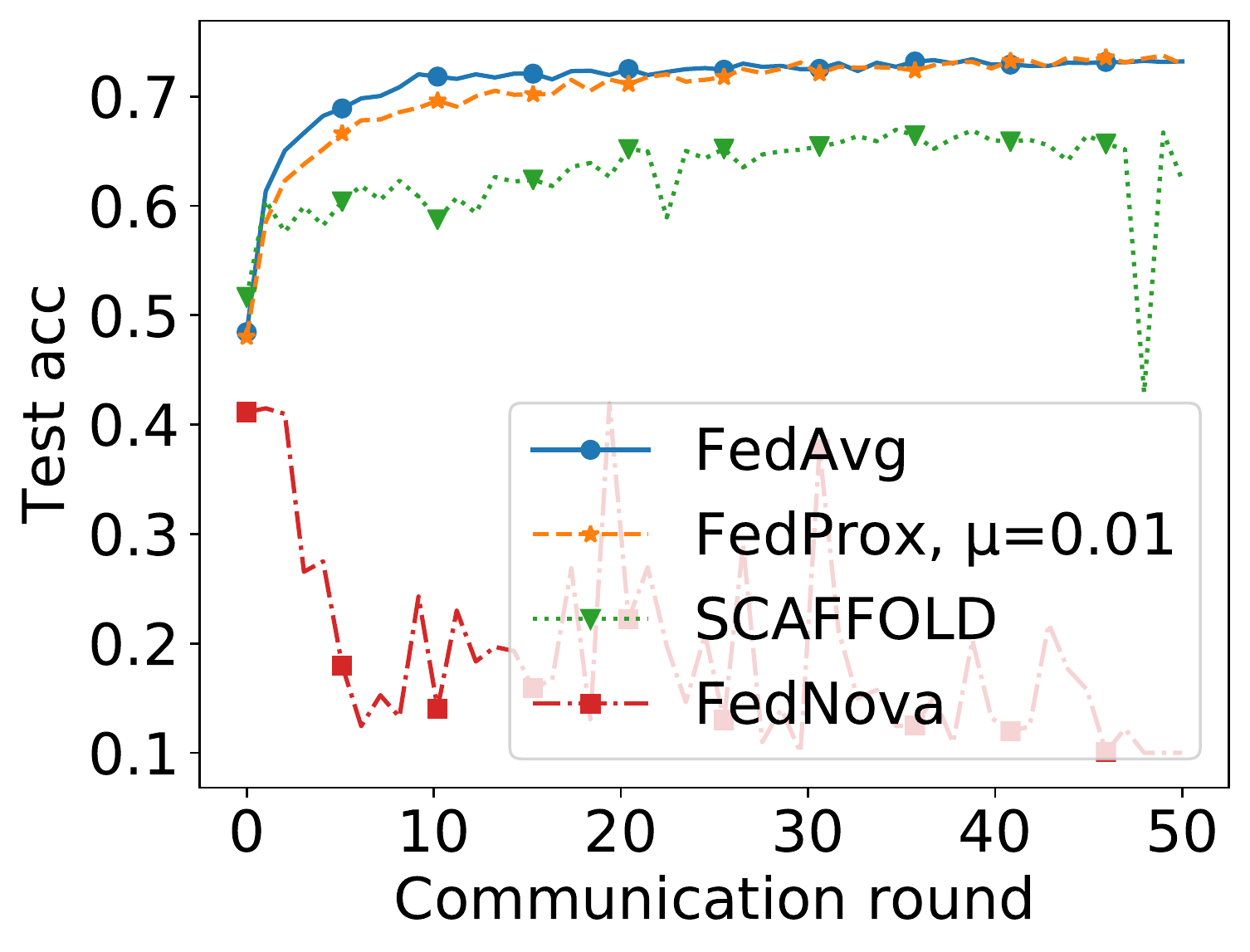}}
    \caption{The training curves of different approaches on CIFAR-10.}
    \label{fig:cifar10_curve_otherfour}
\end{figure*}

\begin{figure*}[h]
    \centering
    \subfloat[$p_k \sim Dir(0.5)$]{\includegraphics[width=0.33\textwidth]{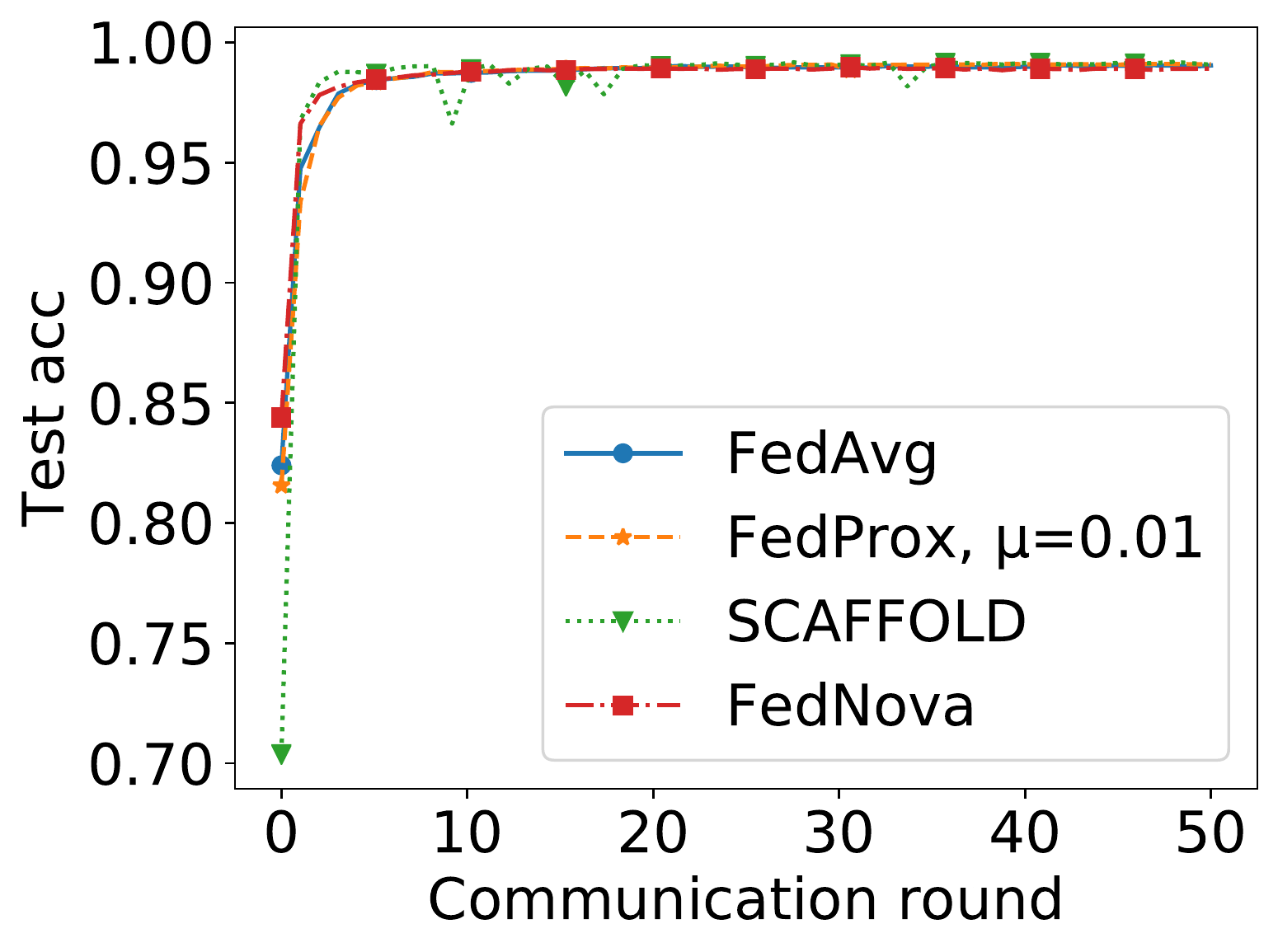}}
    \subfloat[$\#C=1$]{\includegraphics[width=0.33\textwidth]{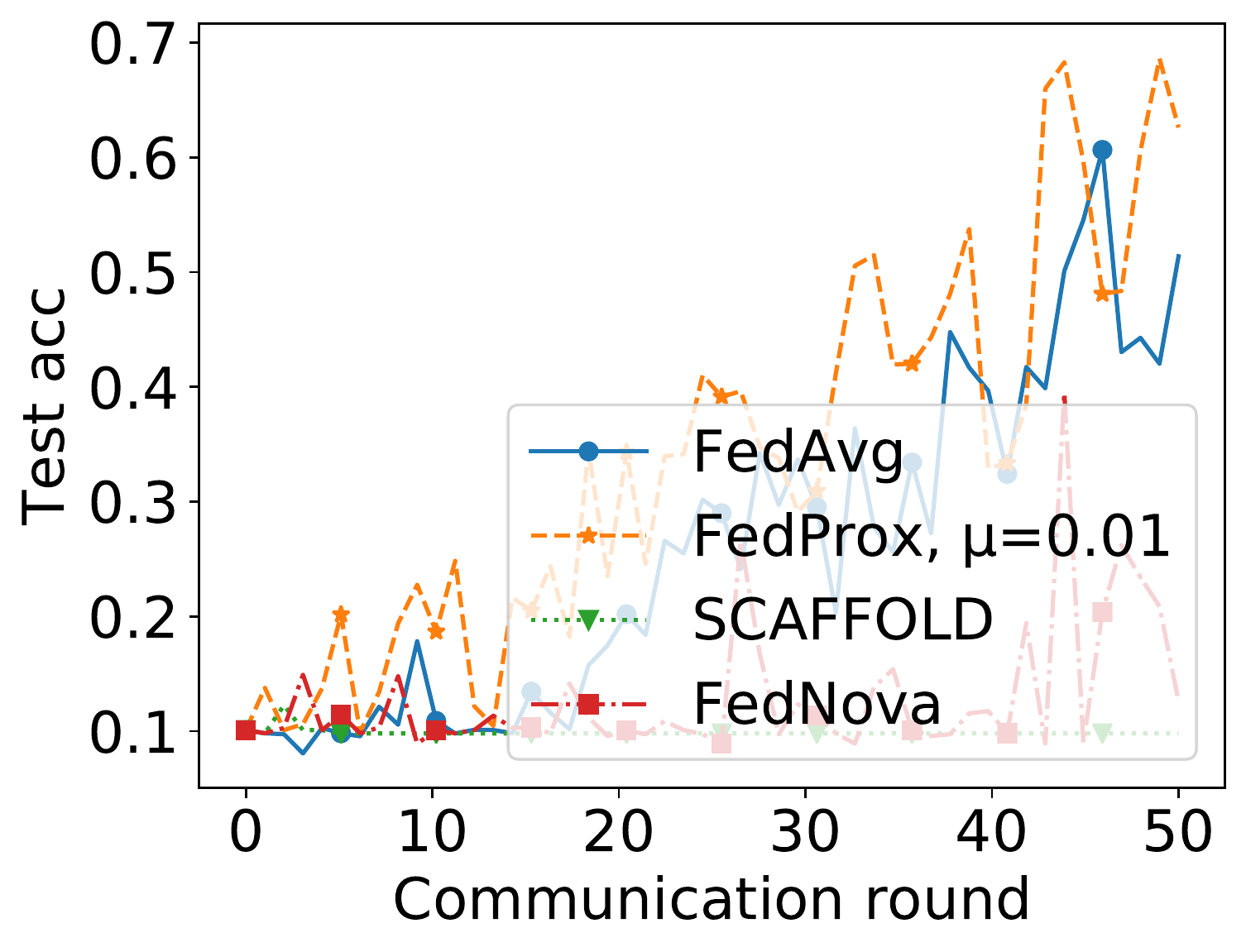}}
    \subfloat[$\#C=2$]{\includegraphics[width=0.33\textwidth]{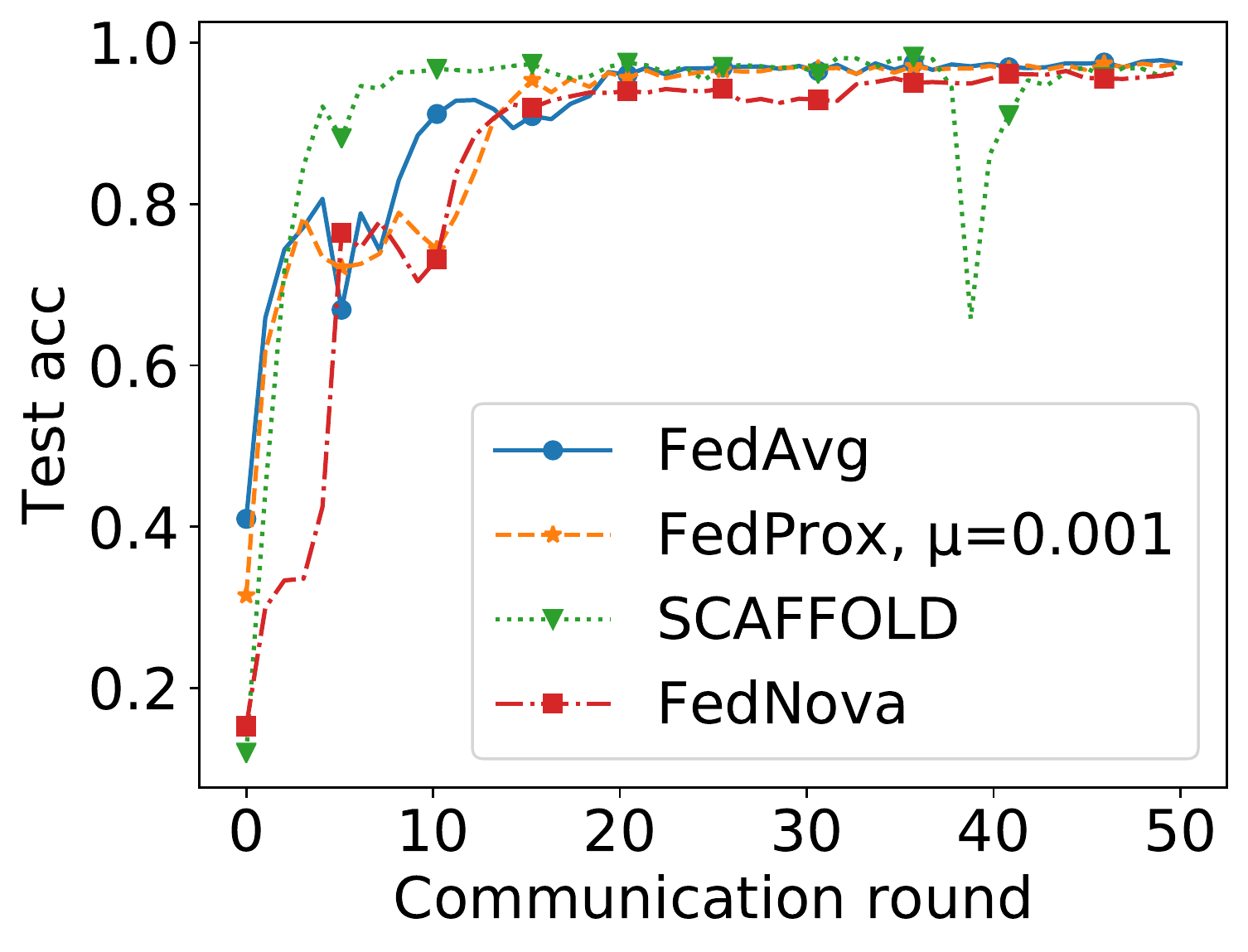}}
    \hfill
    \subfloat[$\#C=3$]{\includegraphics[width=0.33\textwidth]{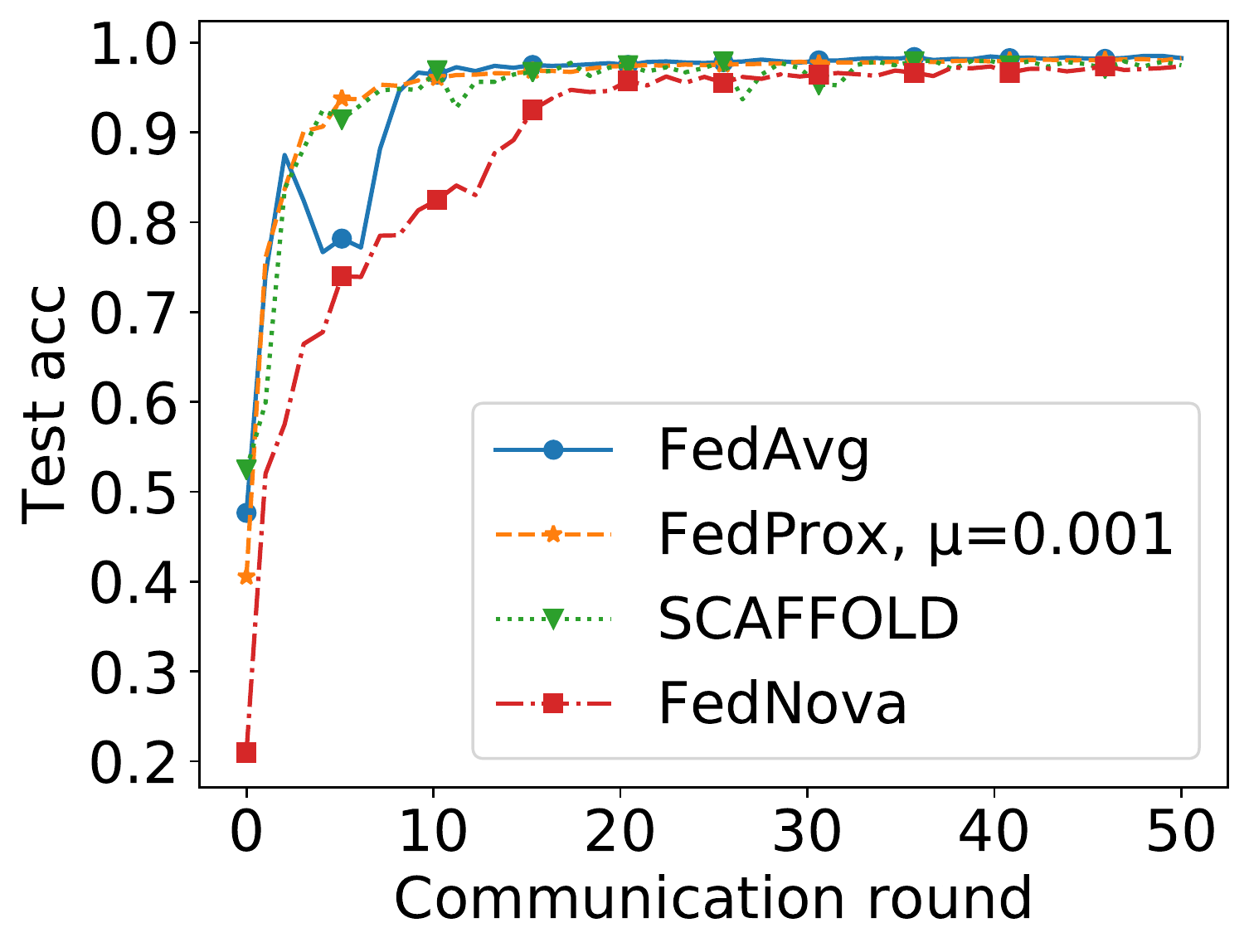}}
    \subfloat[$\hat{\mb{x}} \sim Gau(0.1)$]{\includegraphics[width=0.33\textwidth]{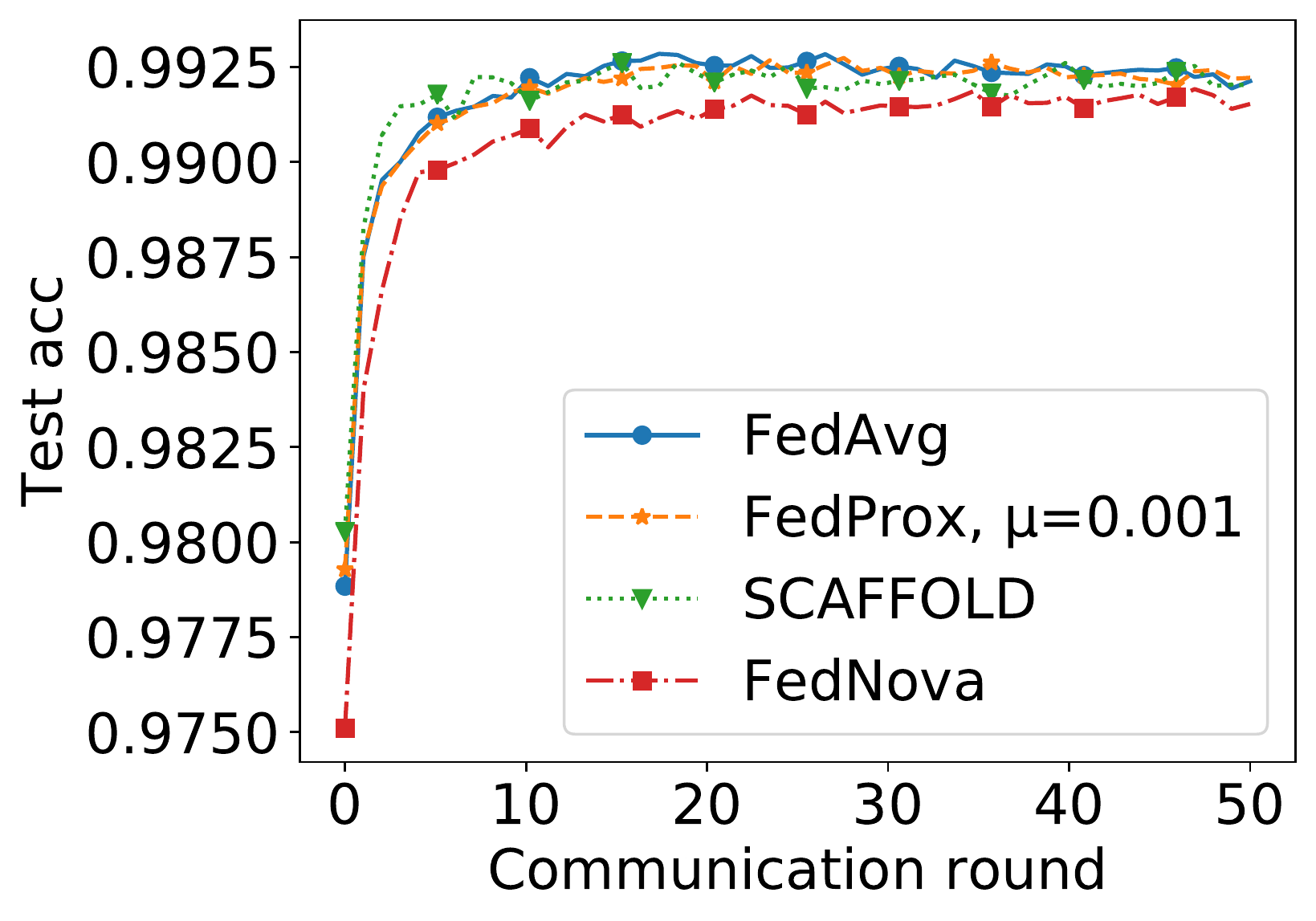}}
    \subfloat[$p \sim Dir(0.5)$]{\includegraphics[width=0.33\textwidth]{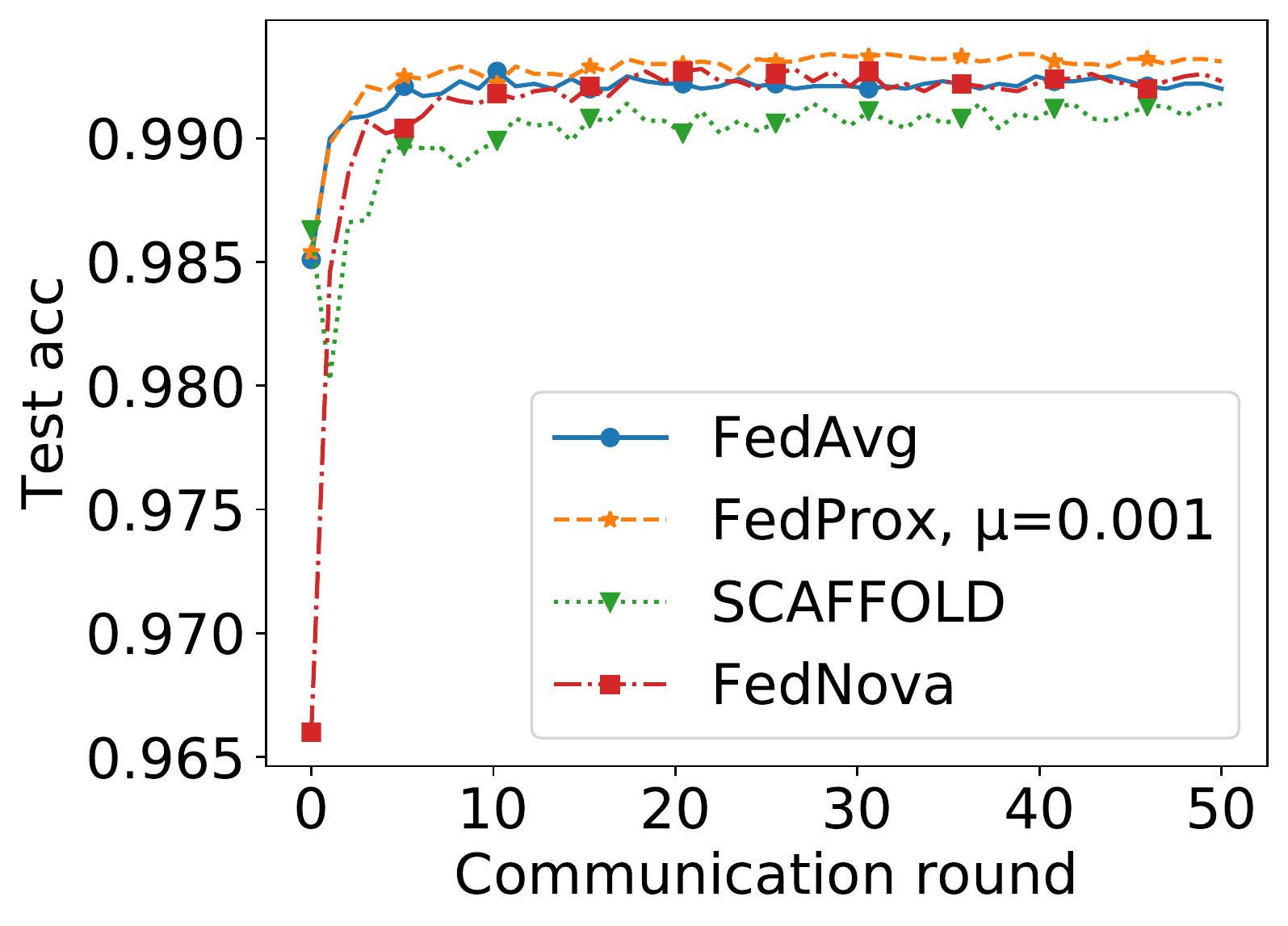}}
    \caption{The training curves of different approaches on MNIST.}
    \label{fig:mnist_curve}
\end{figure*}

\begin{figure*}
    \centering
    \subfloat[$p_k \sim Dir(0.5)$]{\includegraphics[width=0.33\textwidth]{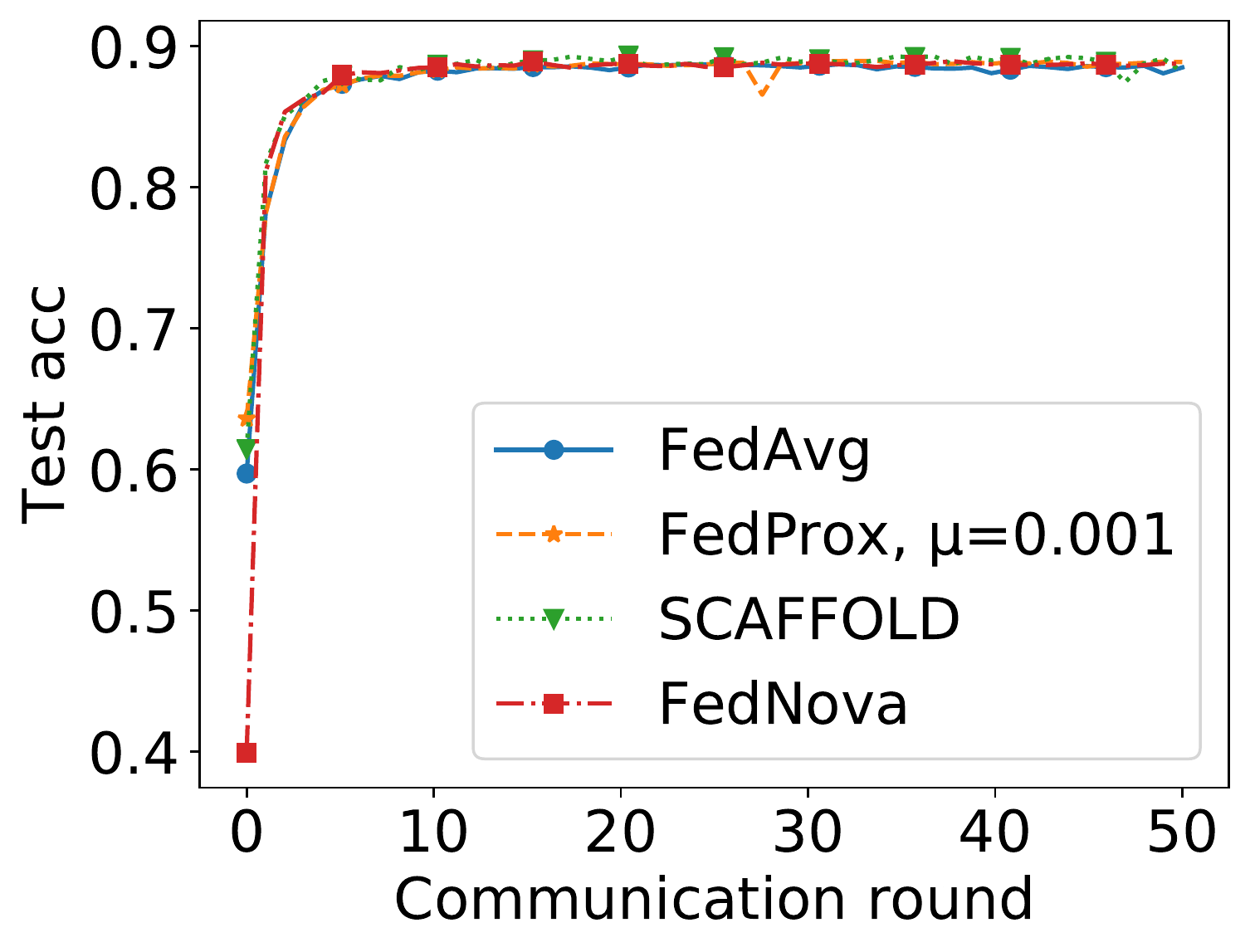}}
    \subfloat[$\#C=1$]{\includegraphics[width=0.33\textwidth]{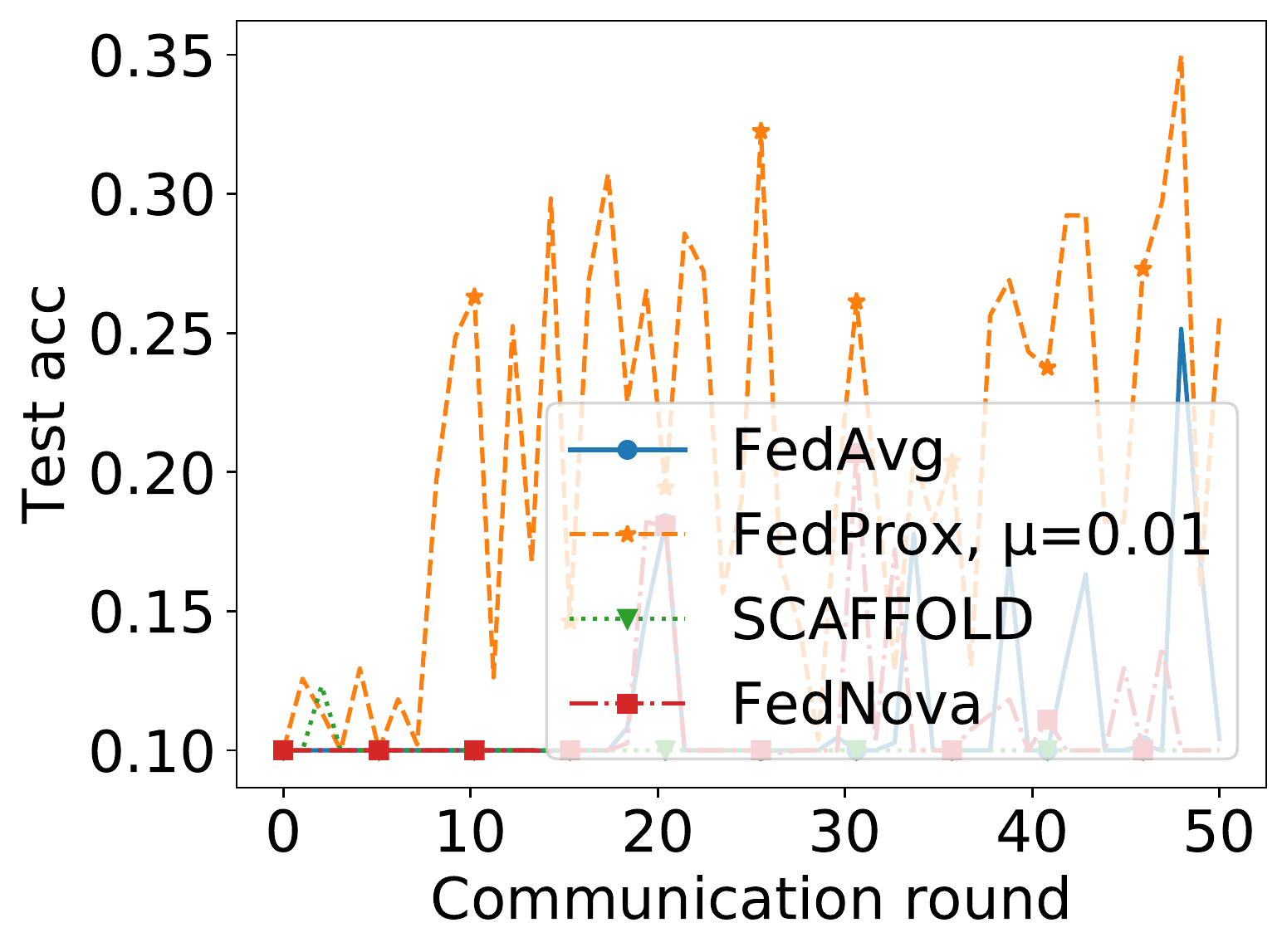}}
    \subfloat[$\#C=2$]{\includegraphics[width=0.33\textwidth]{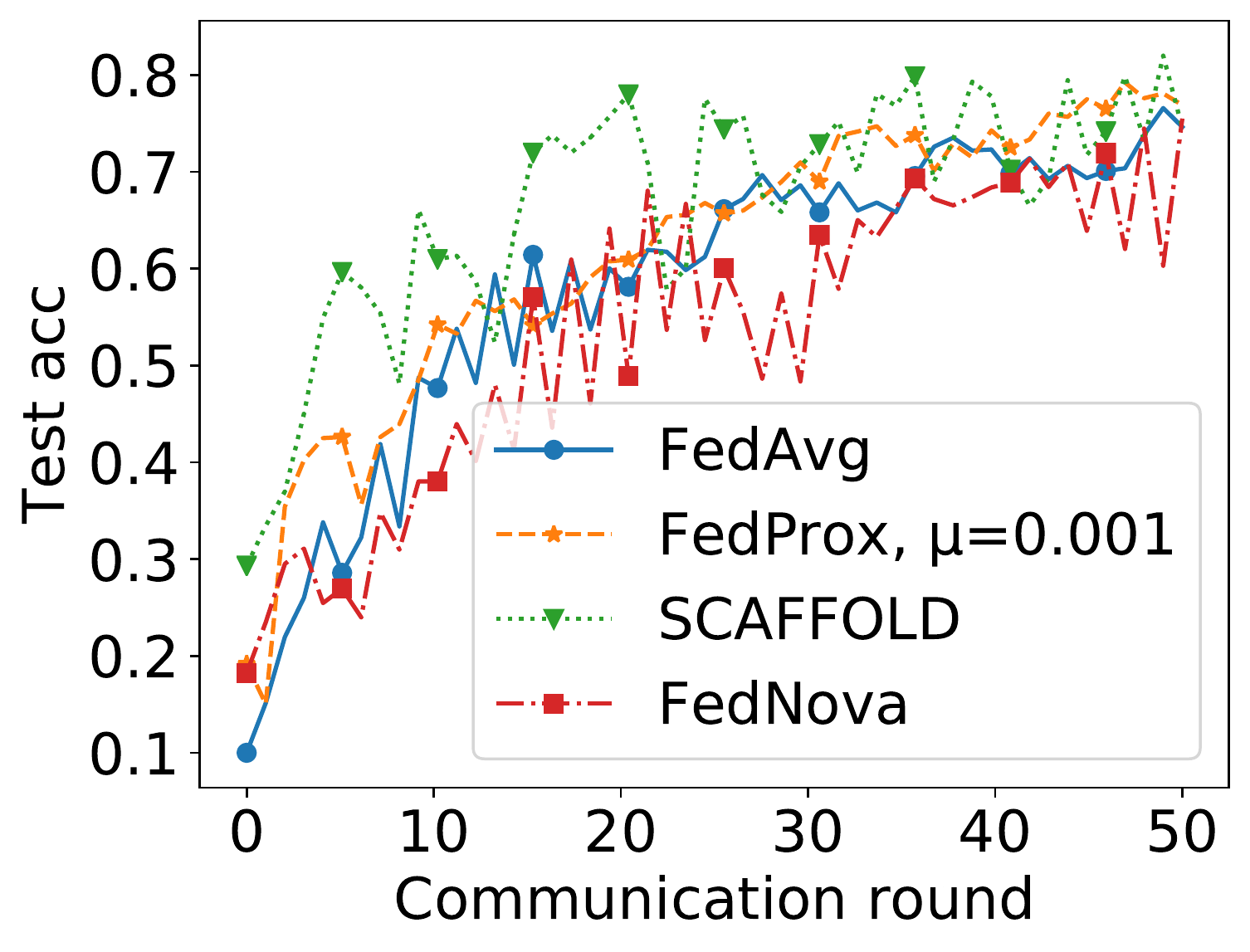}}
    \hfill
    \subfloat[$\#C=3$]{\includegraphics[width=0.33\textwidth]{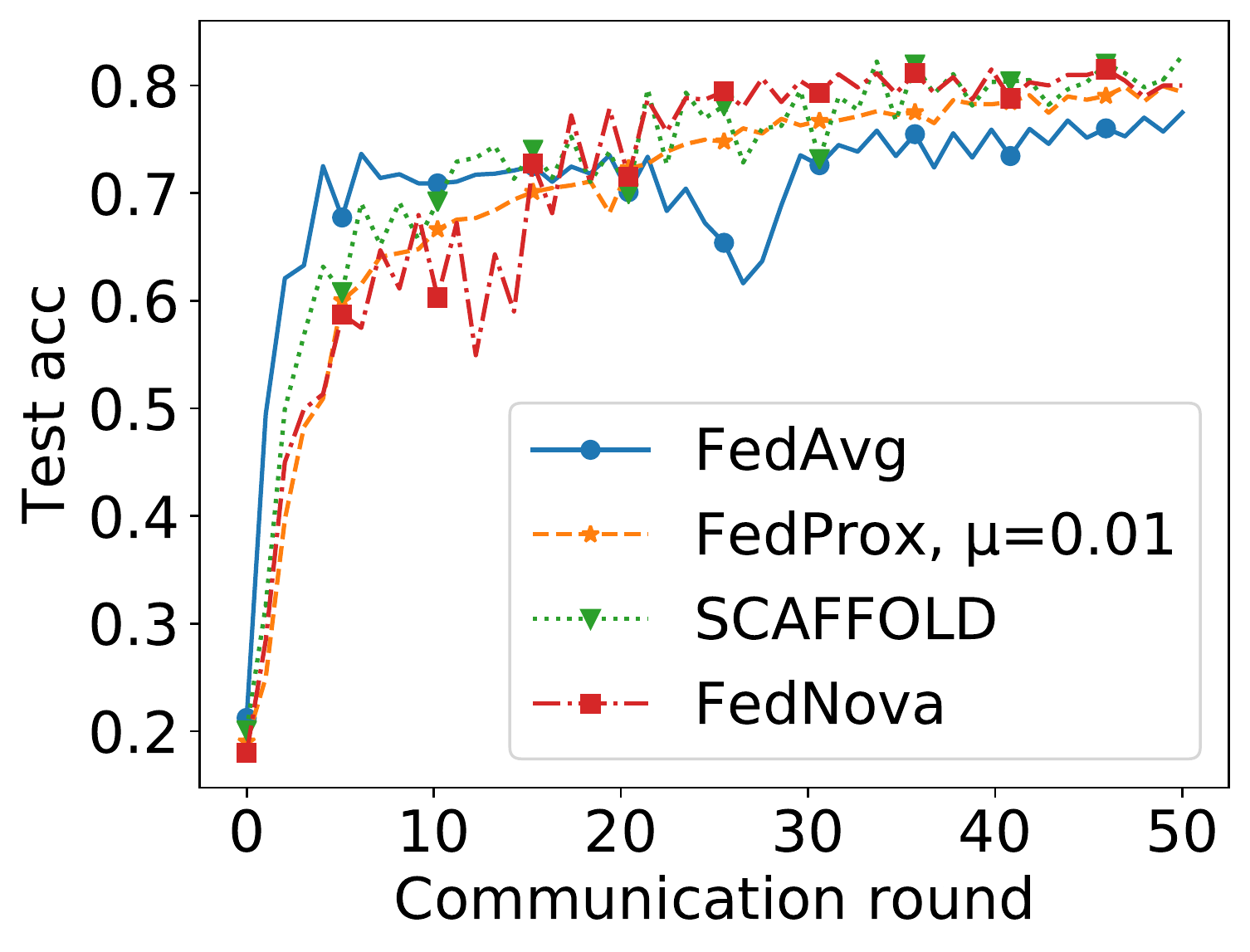}}
    \subfloat[$\hat{\mb{x}} \sim Gau(0.1)$]{\includegraphics[width=0.33\textwidth]{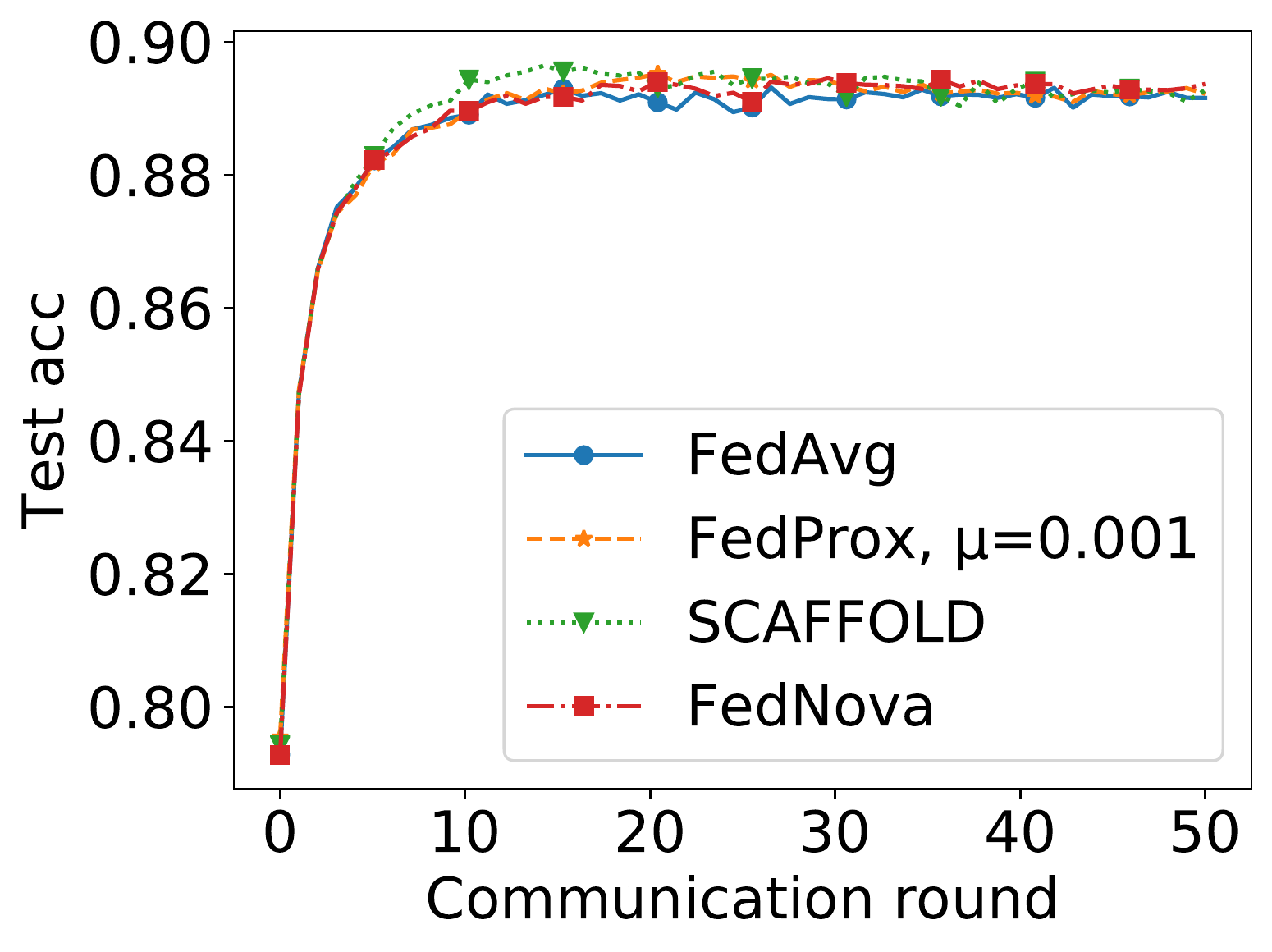}}
    \subfloat[$p \sim Dir(0.5)$]{\includegraphics[width=0.33\textwidth]{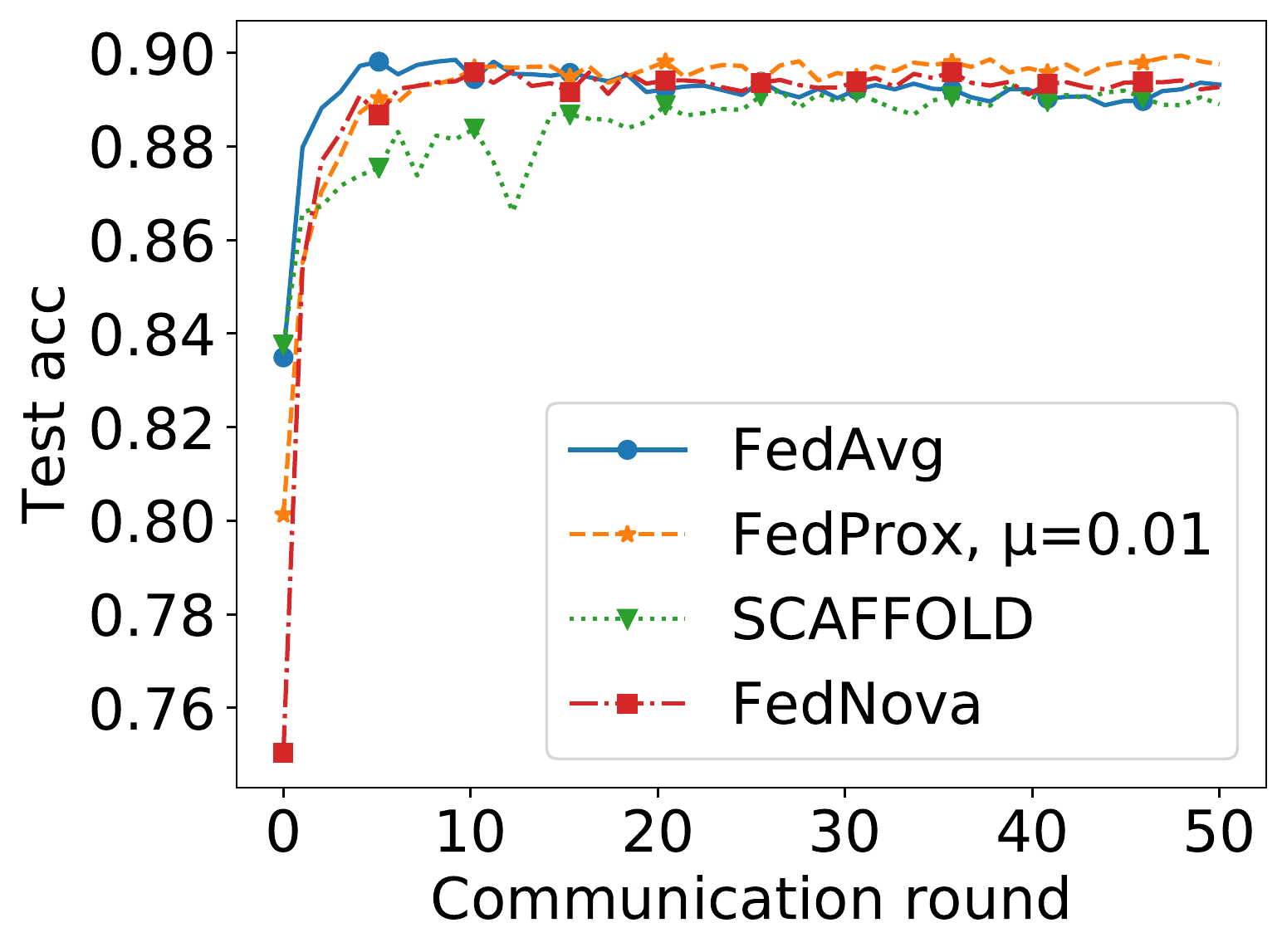}}
    \caption{The training curves of different approaches on FMNIST.}
    \label{fig:fmnist_curve}
\end{figure*}

\begin{figure*}
    \centering
    \subfloat[$p_k \sim Dir(0.5)$]{\includegraphics[width=0.33\textwidth]{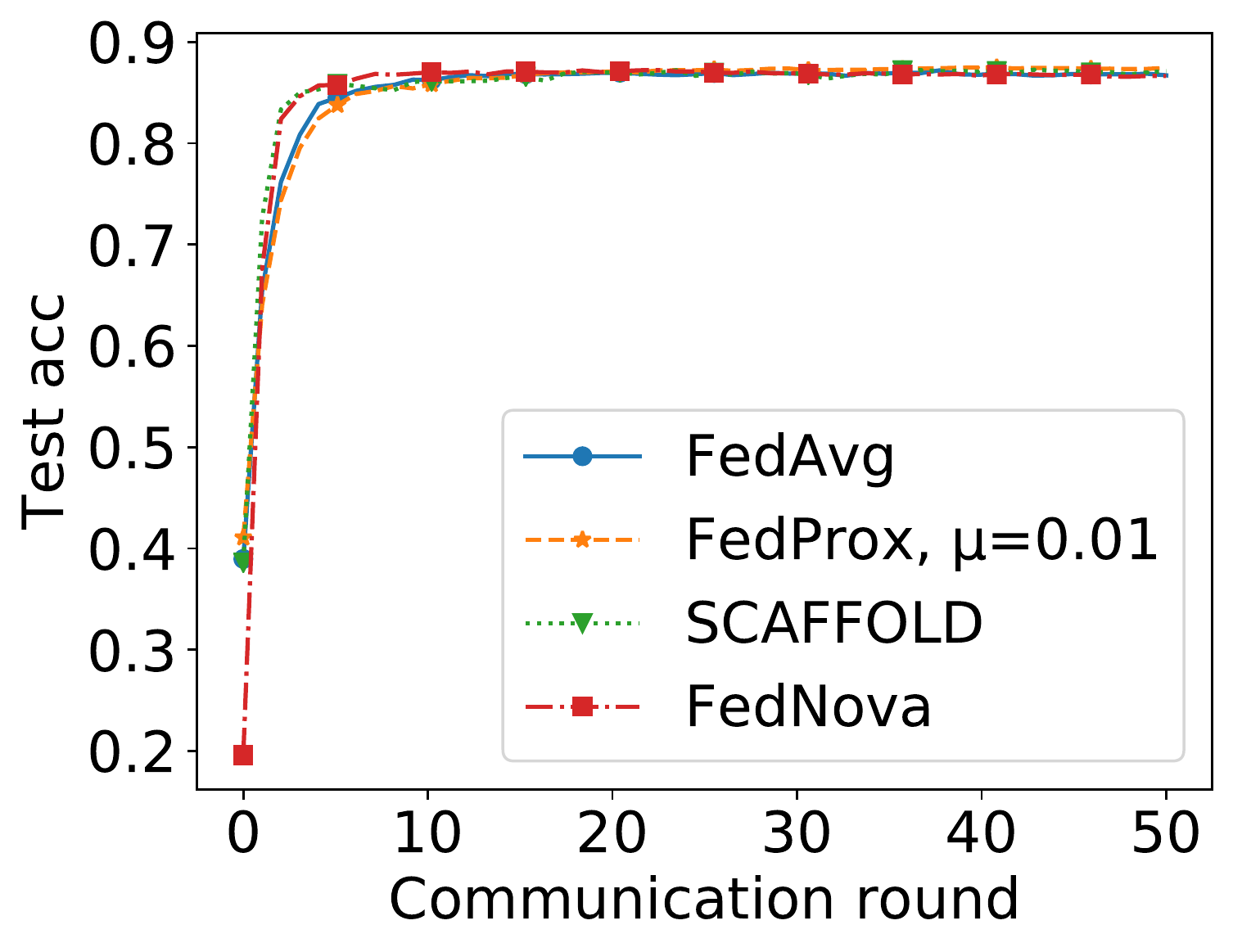}}
    \subfloat[$\#C=1$]{\includegraphics[width=0.33\textwidth]{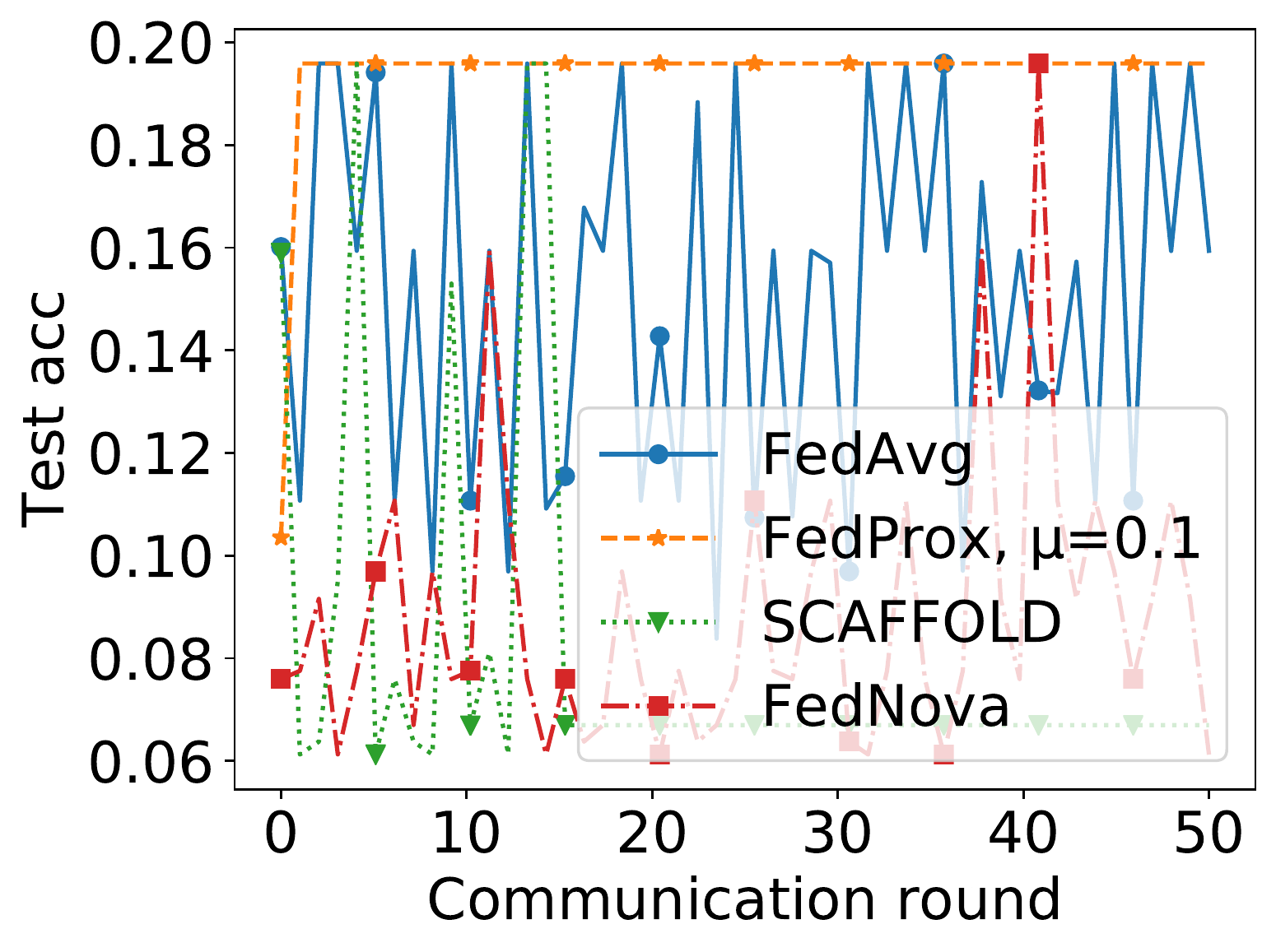}}
    \subfloat[$\#C=2$]{\includegraphics[width=0.33\textwidth]{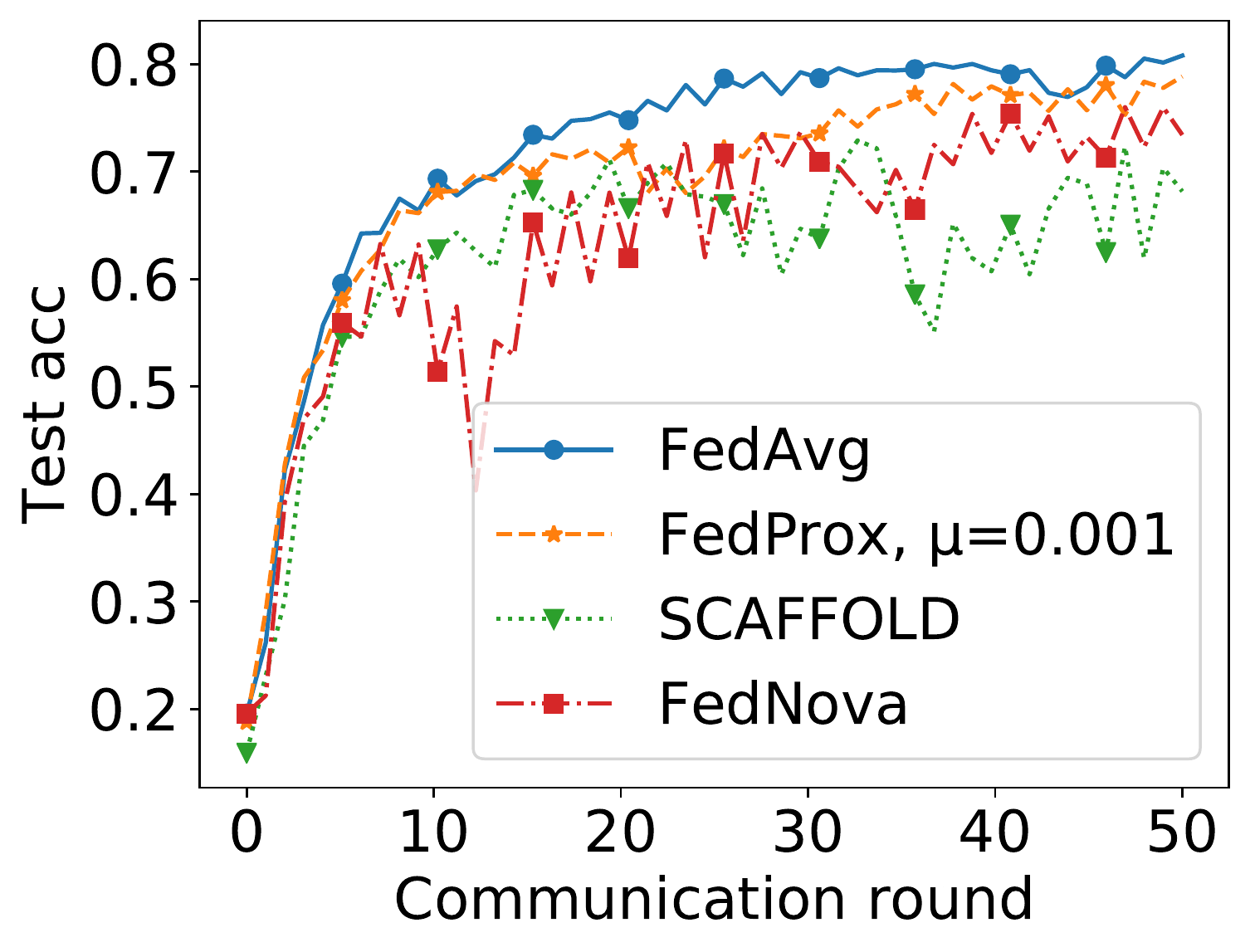}}
    \hfill
    \subfloat[$\#C=3$]{\includegraphics[width=0.33\textwidth]{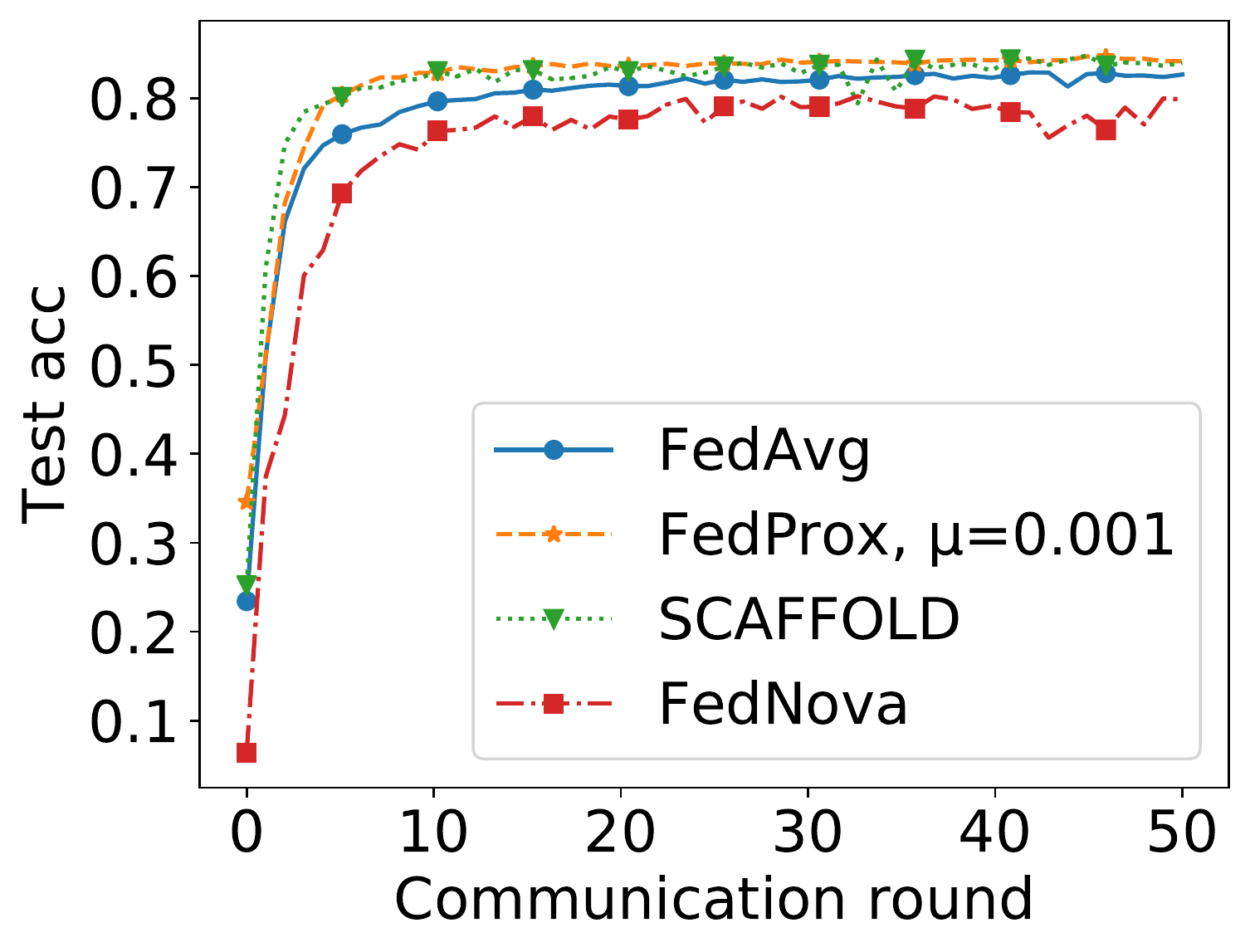}}
    \subfloat[$\hat{\mb{x}} \sim Gau(0.1)$]{\includegraphics[width=0.33\textwidth]{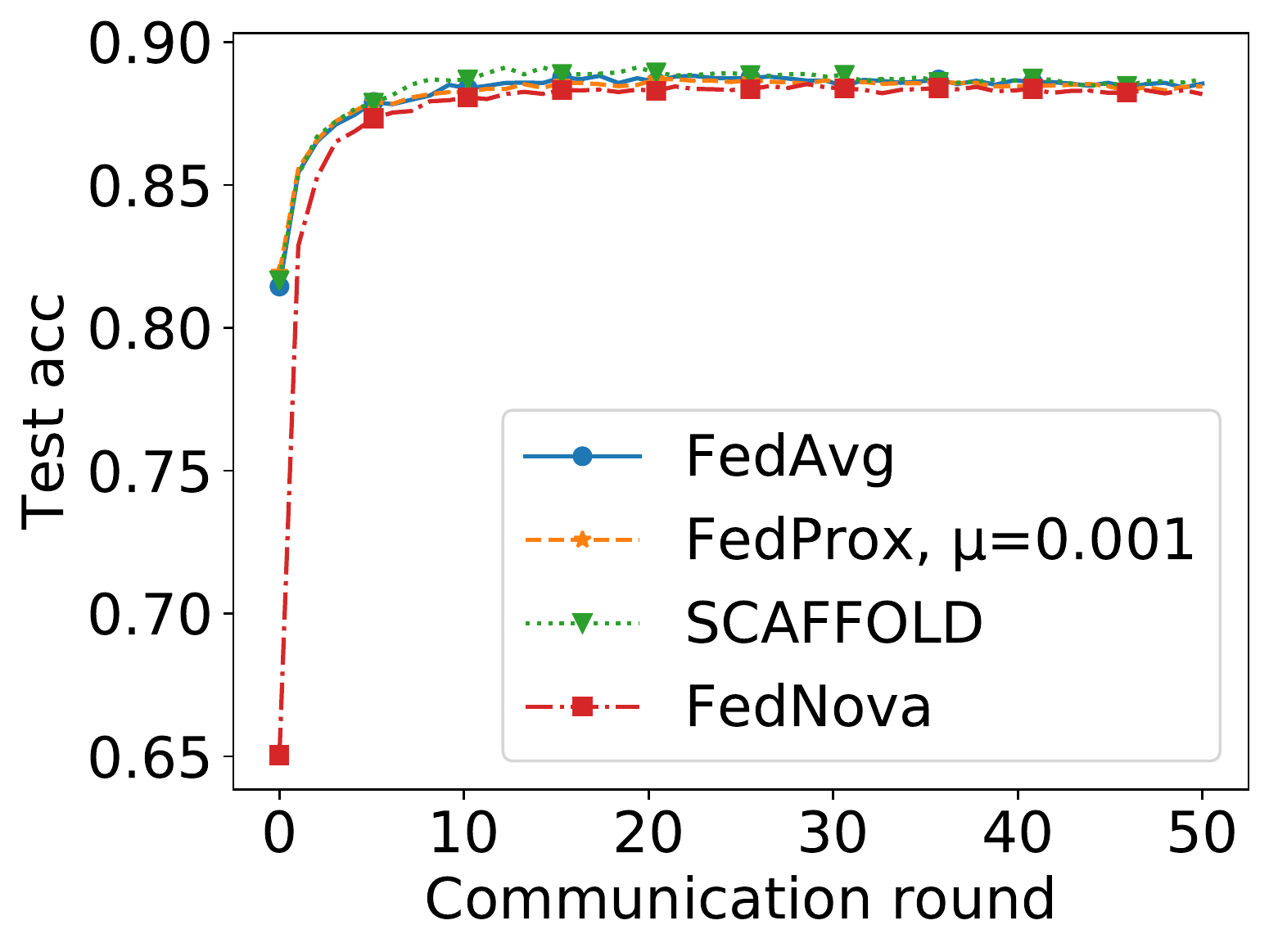}}
    \subfloat[$p \sim Dir(0.5)$]{\includegraphics[width=0.33\textwidth]{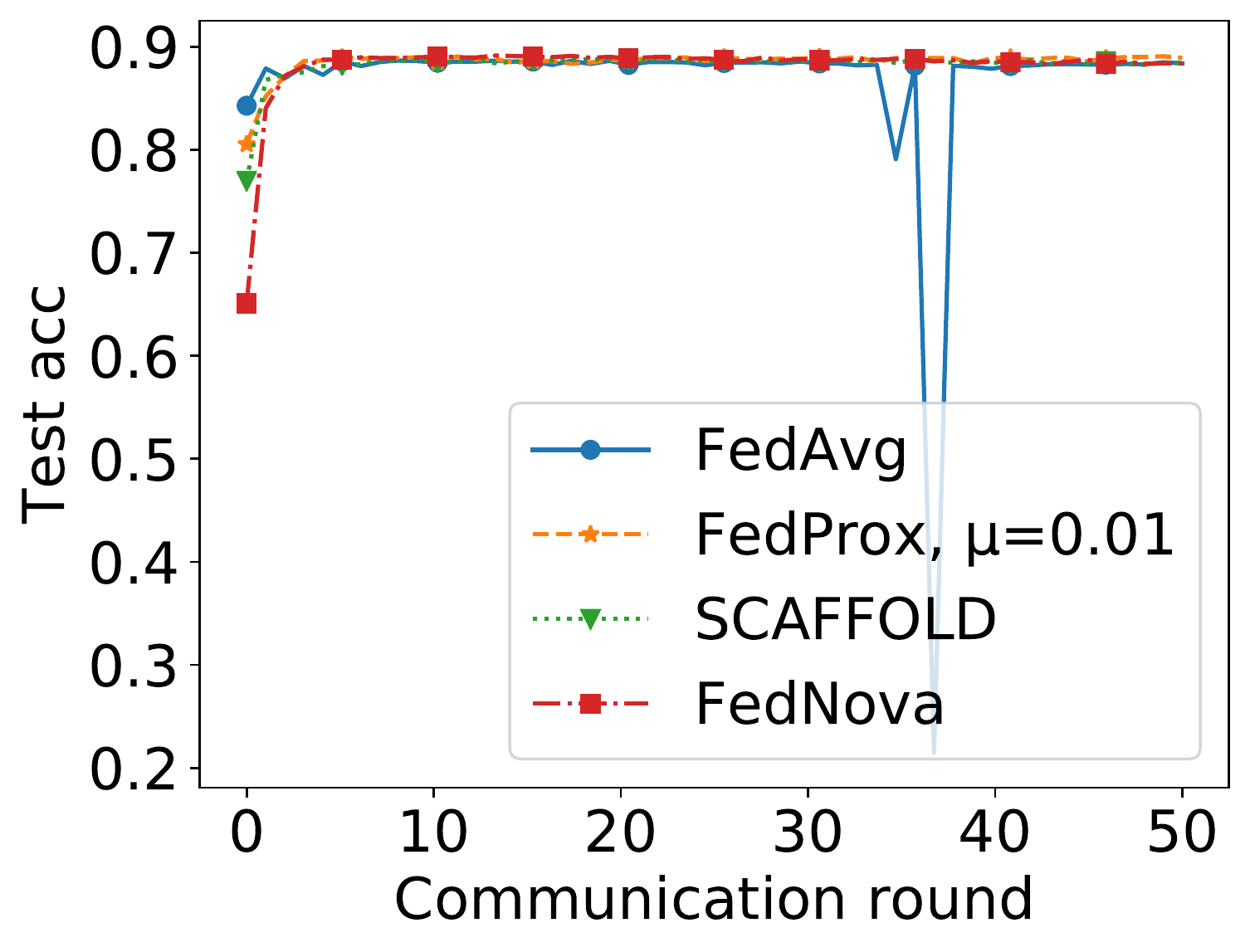}}
    \caption{The training curves of different approaches on SVHN.}
    \label{fig:svhn_curve}
\end{figure*}

\begin{figure*}
    \centering
    \subfloat[FCUBE]{\includegraphics[width=0.33\textwidth]{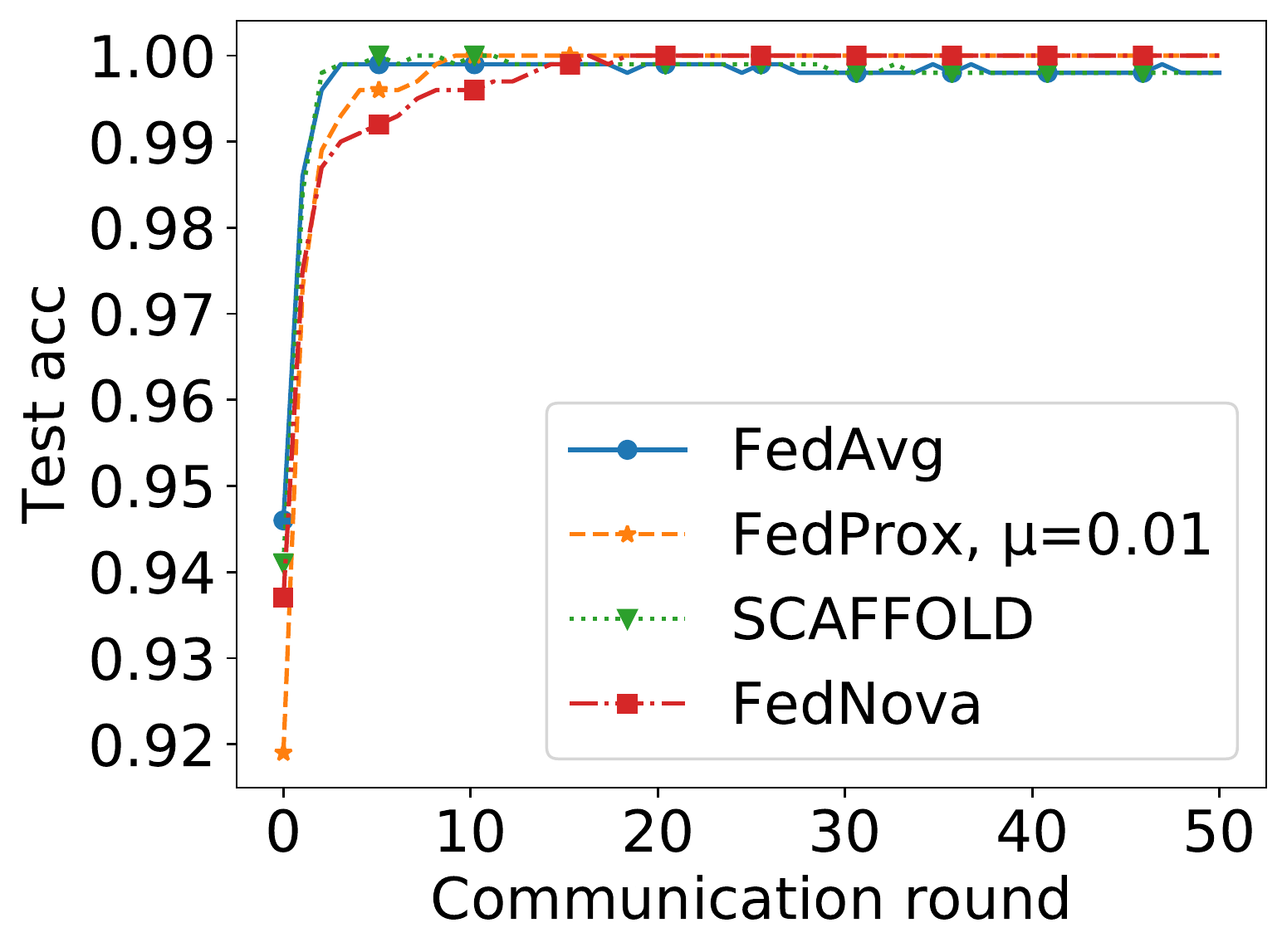}}
    \subfloat[FEMNIST]{\includegraphics[width=0.33\textwidth]{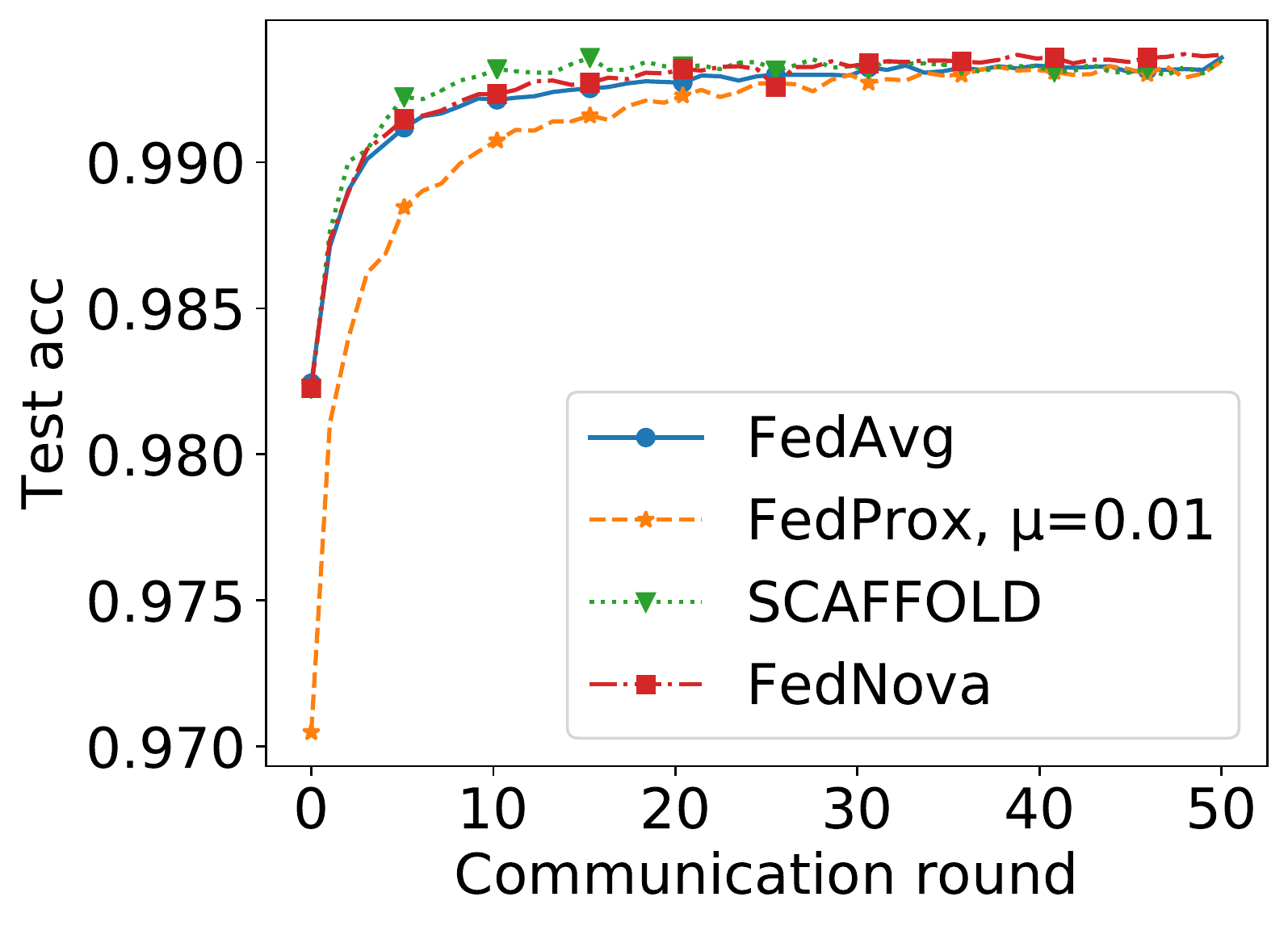}}
    \caption{The training curves of different approaches on FCUBE and FEMNIST.}
    \label{fig:cube_femnist_curve}
\end{figure*}

Figures \ref{fig:cifar10_curve_otherfour}, \ref{fig:mnist_curve}, \ref{fig:fmnist_curve}, \ref{fig:svhn_curve}, and \ref{fig:cube_femnist_curve} show the training curves of different approaches on the studied datasets except CIFAR-10.

\subsection{Number of Local Epochs}
\label{app:epoch}
\begin{figure*}
    \centering
    \subfloat[$p_k \sim Dir(0.5)$]{\includegraphics[width=0.33\textwidth]{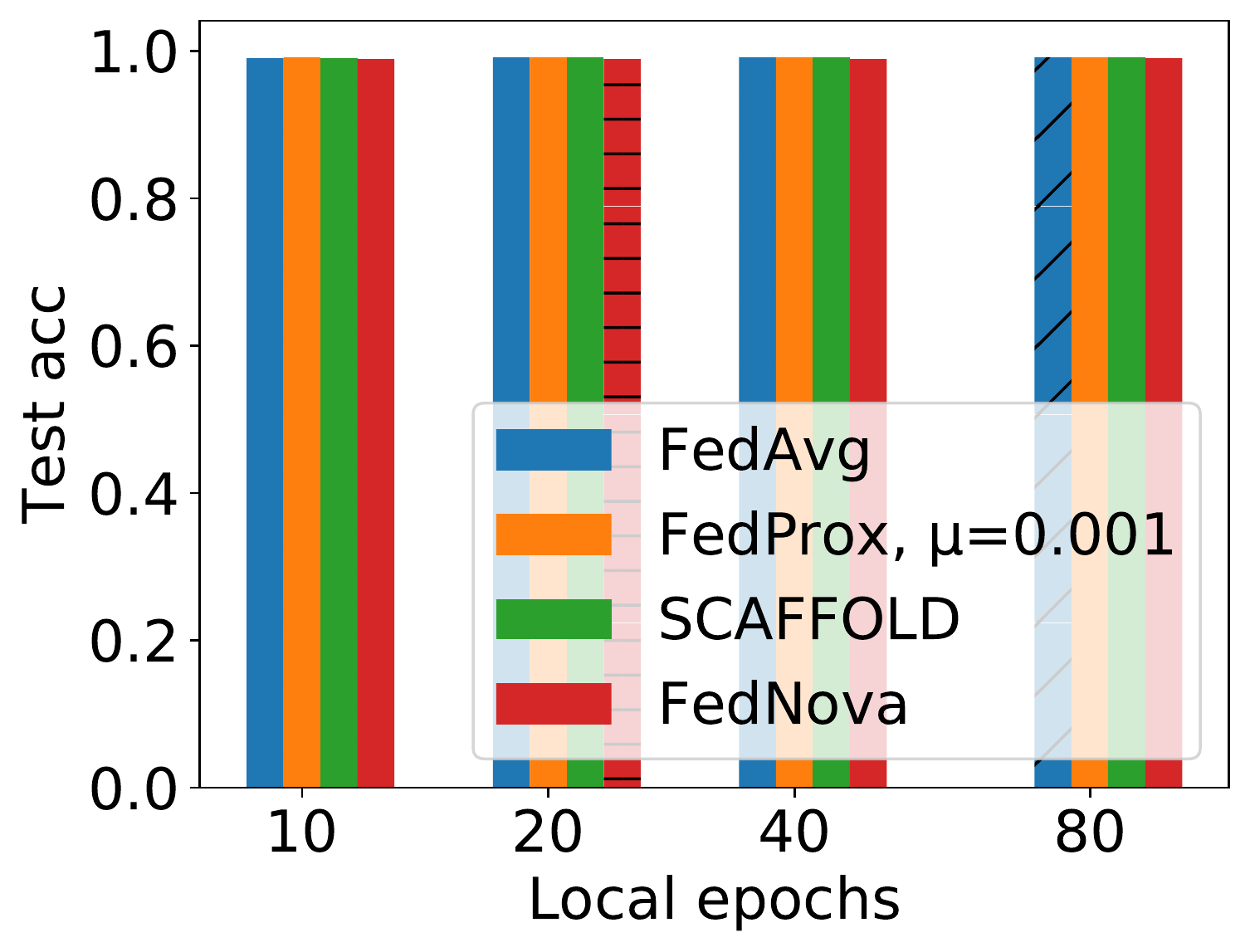}}
    \subfloat[$\#C=1$]{\includegraphics[width=0.33\textwidth]{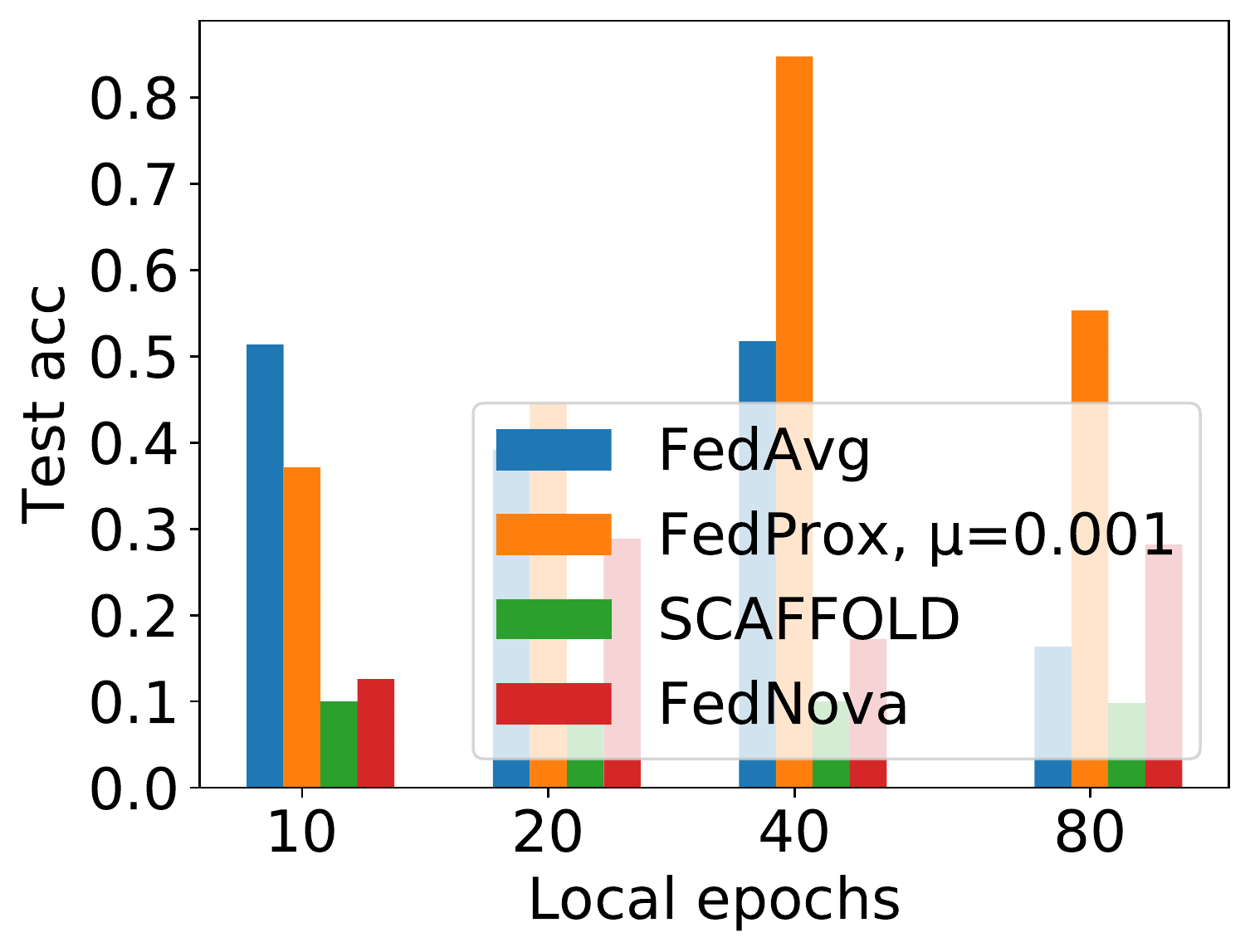}}
    \subfloat[$\#C=2$]{\includegraphics[width=0.33\textwidth]{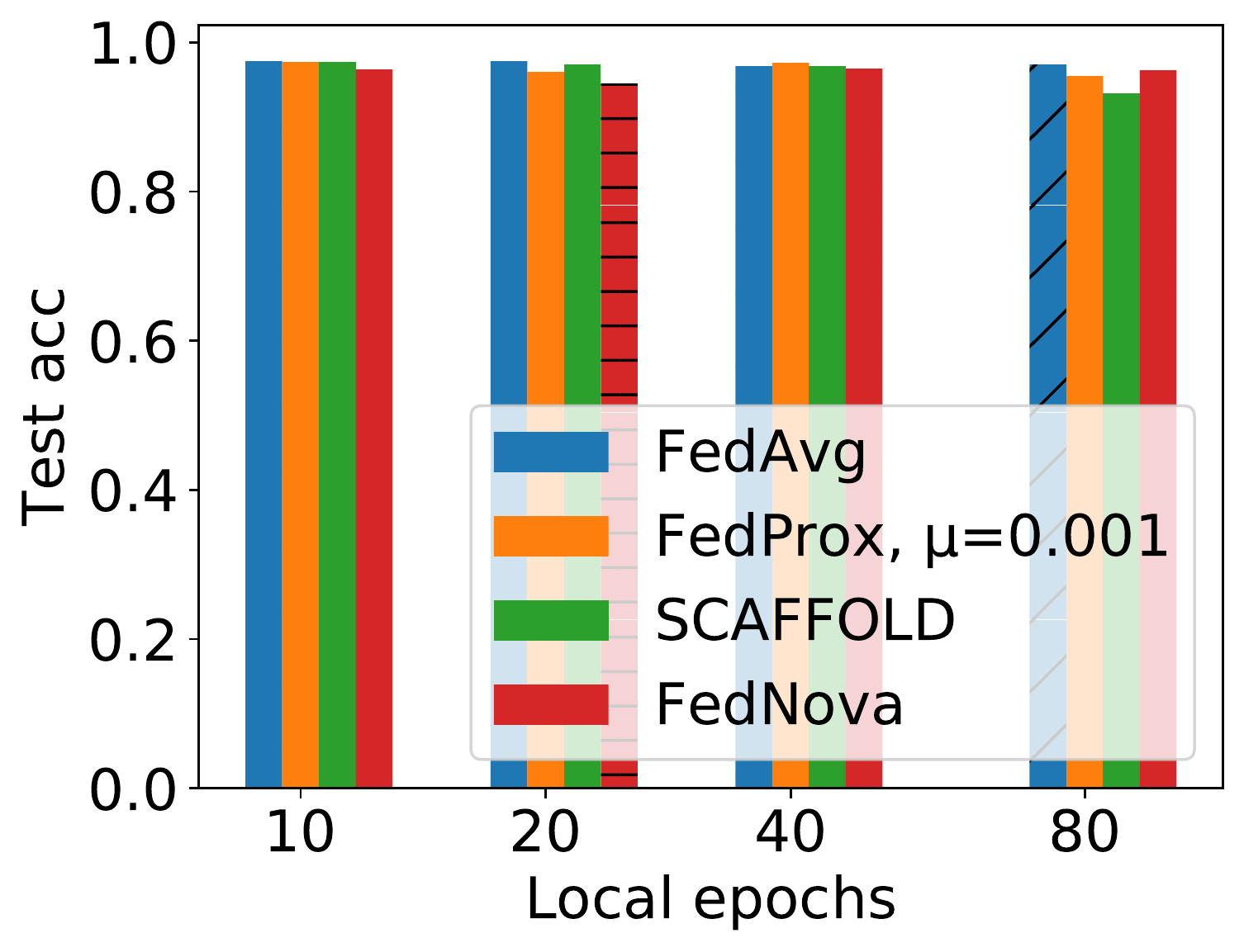}}
    \hfill
    \subfloat[$\#C=3$]{\includegraphics[width=0.33\textwidth]{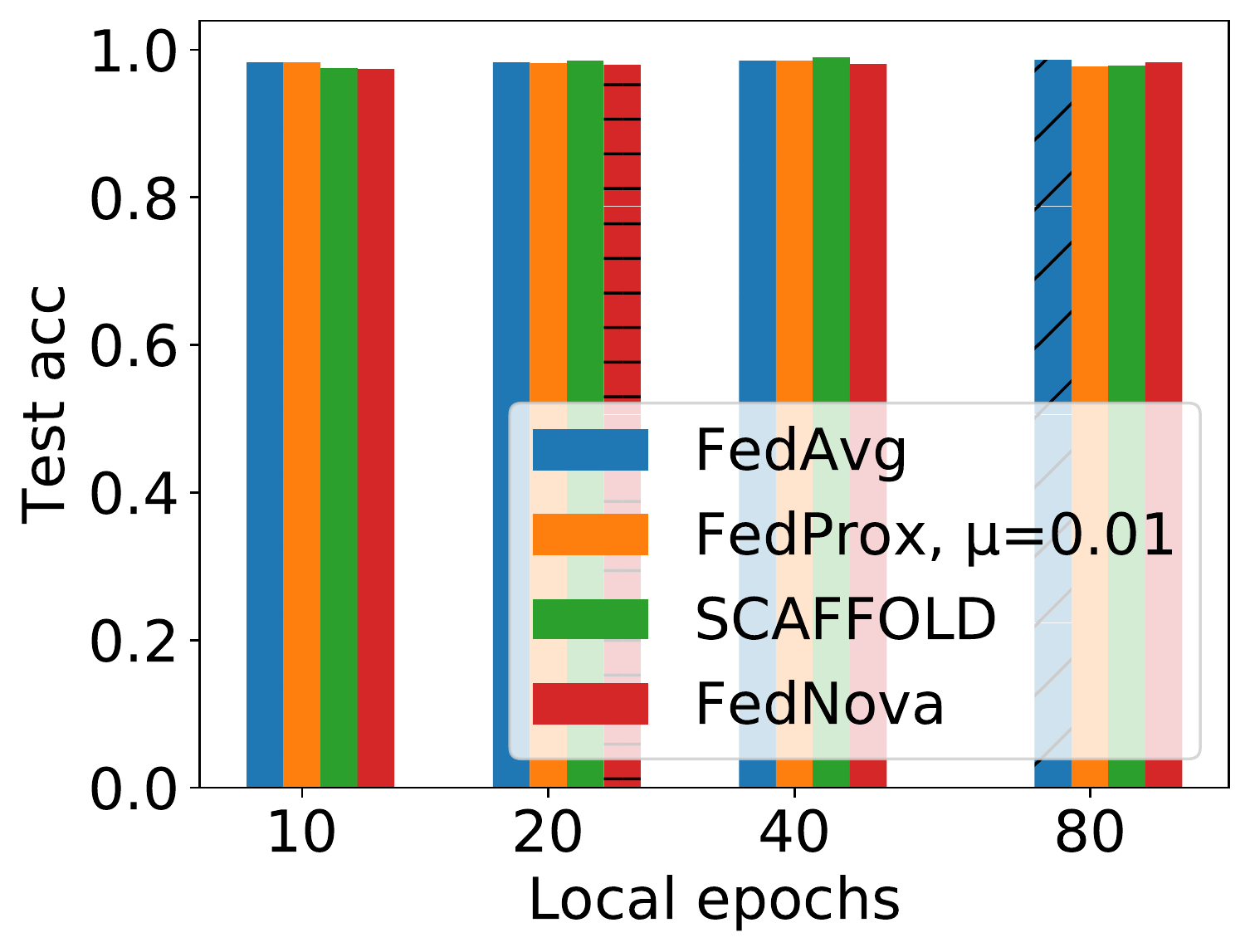}}
    \subfloat[$\hat{\mb{x}} \sim Gau(0.1)$]{\includegraphics[width=0.33\textwidth]{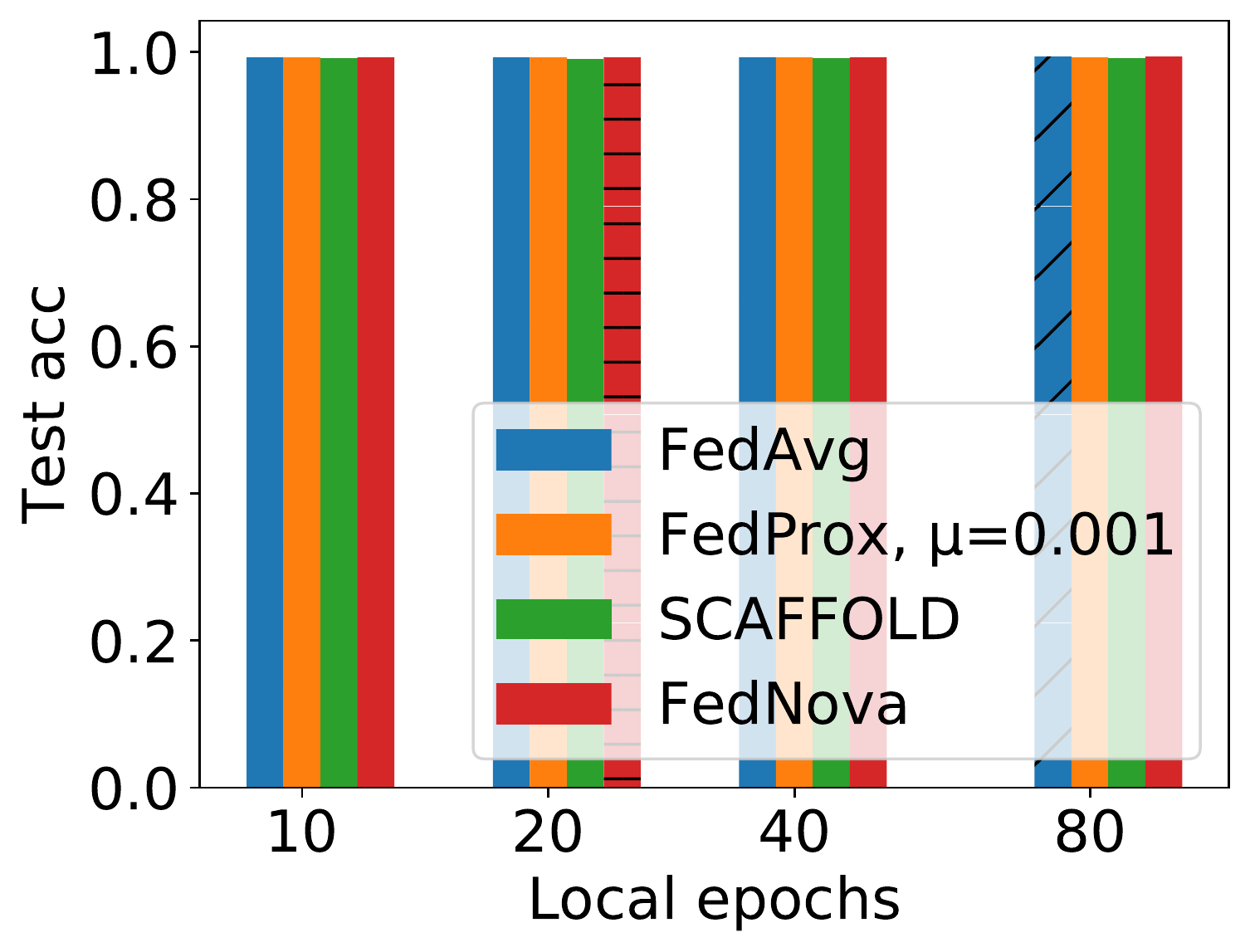}}
    \subfloat[$p \sim Dir(0.5)$]{\includegraphics[width=0.33\textwidth]{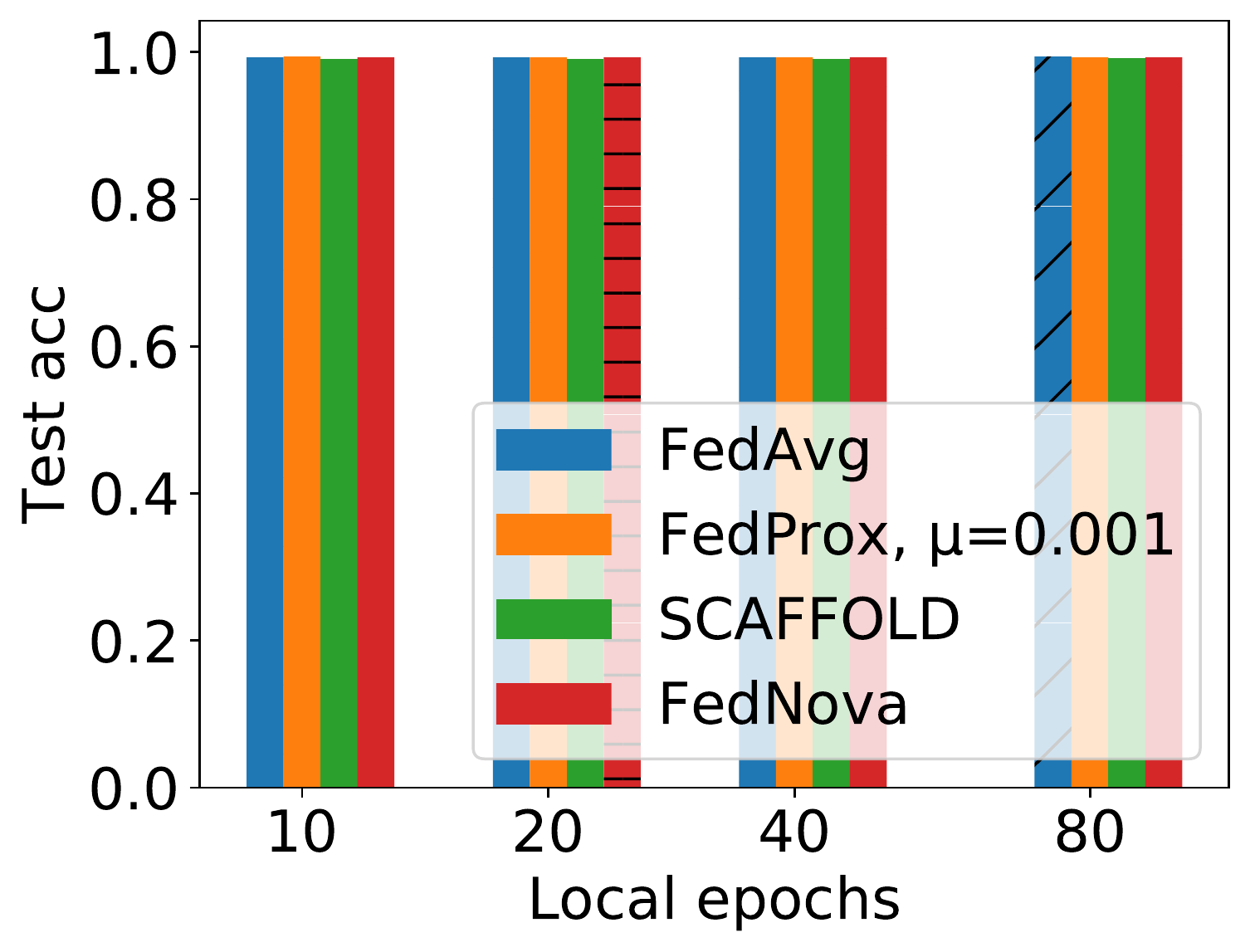}}
    \caption{The test accuracy with different number of local epochs on MNIST.}
    \label{fig:mnist_epoch}
\end{figure*}

\begin{figure*}
    \centering
    \subfloat[$p_k \sim Dir(0.5)$]{\includegraphics[width=0.33\textwidth]{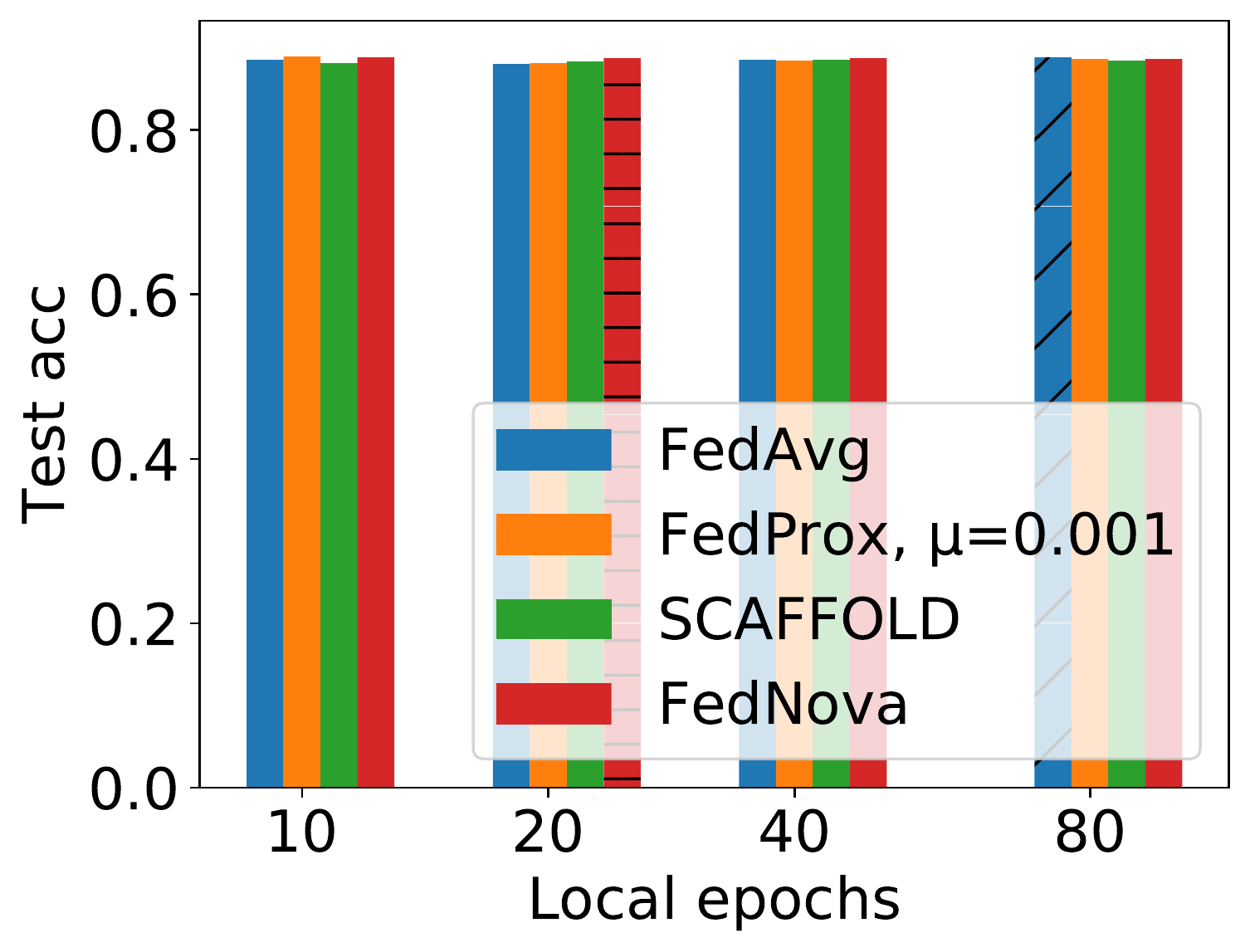}}
    \subfloat[$\#C=1$]{\includegraphics[width=0.33\textwidth]{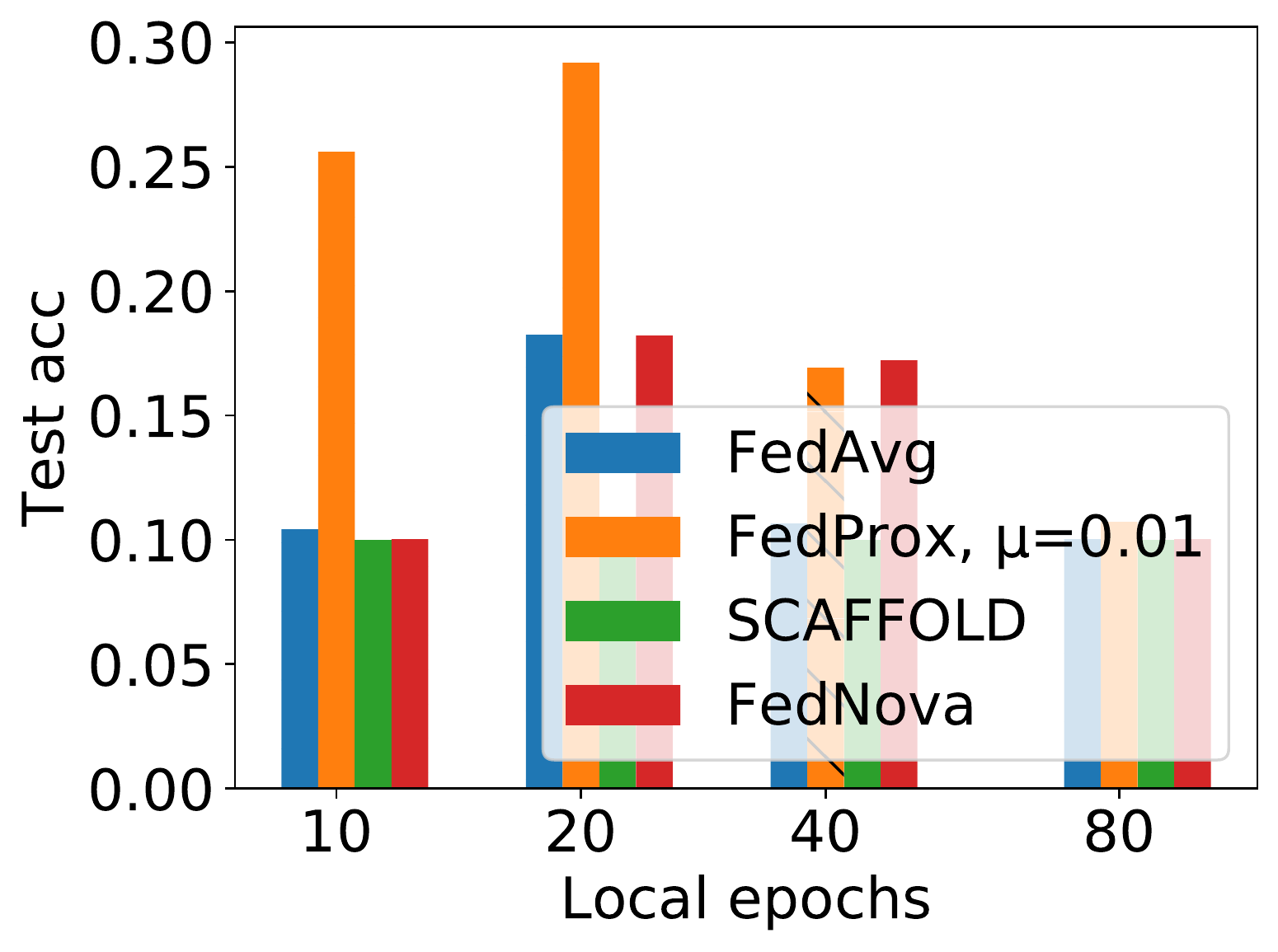}}
    \subfloat[$\#C=2$]{\includegraphics[width=0.33\textwidth]{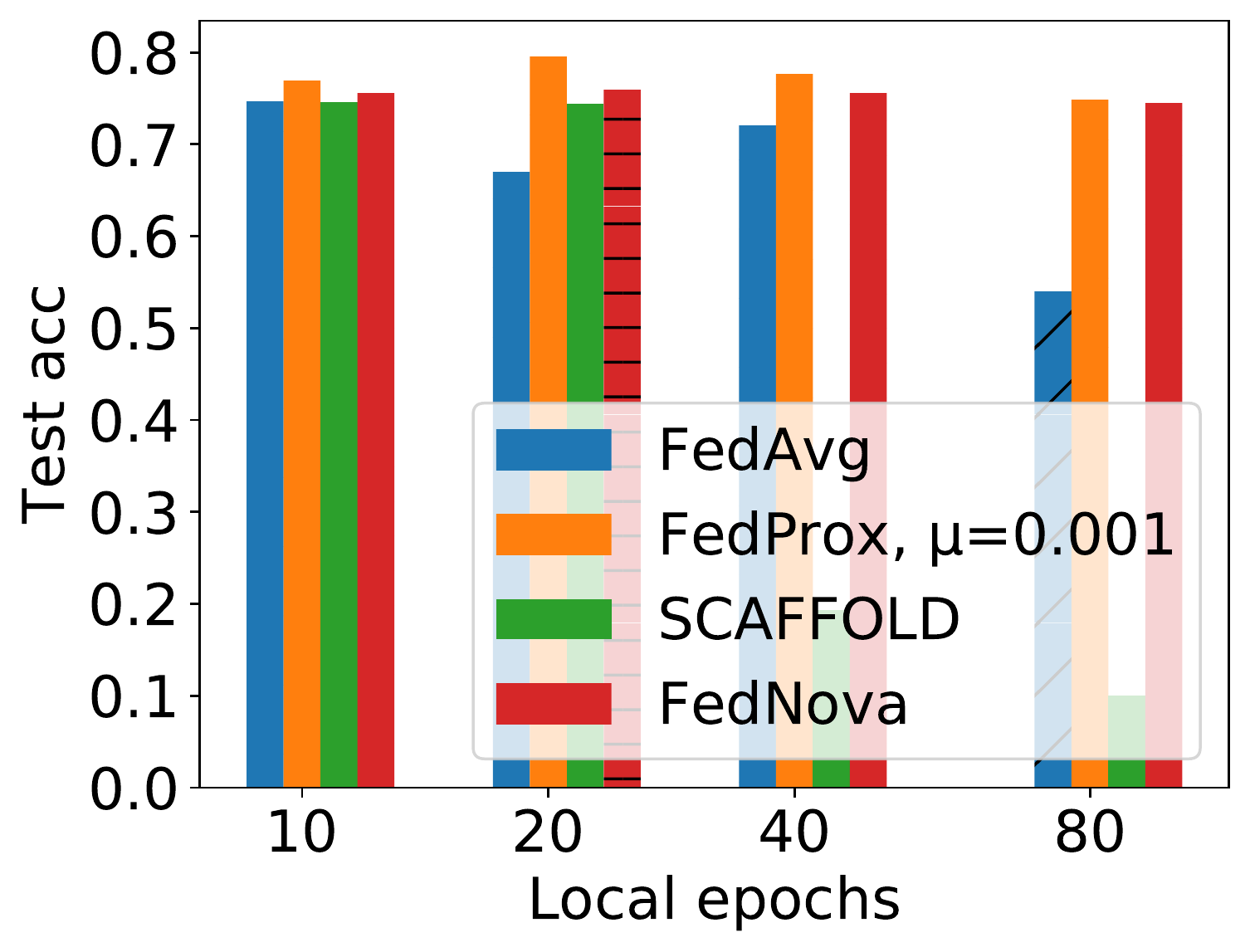}}
    \hfill
    \subfloat[$\#C=3$]{\includegraphics[width=0.33\textwidth]{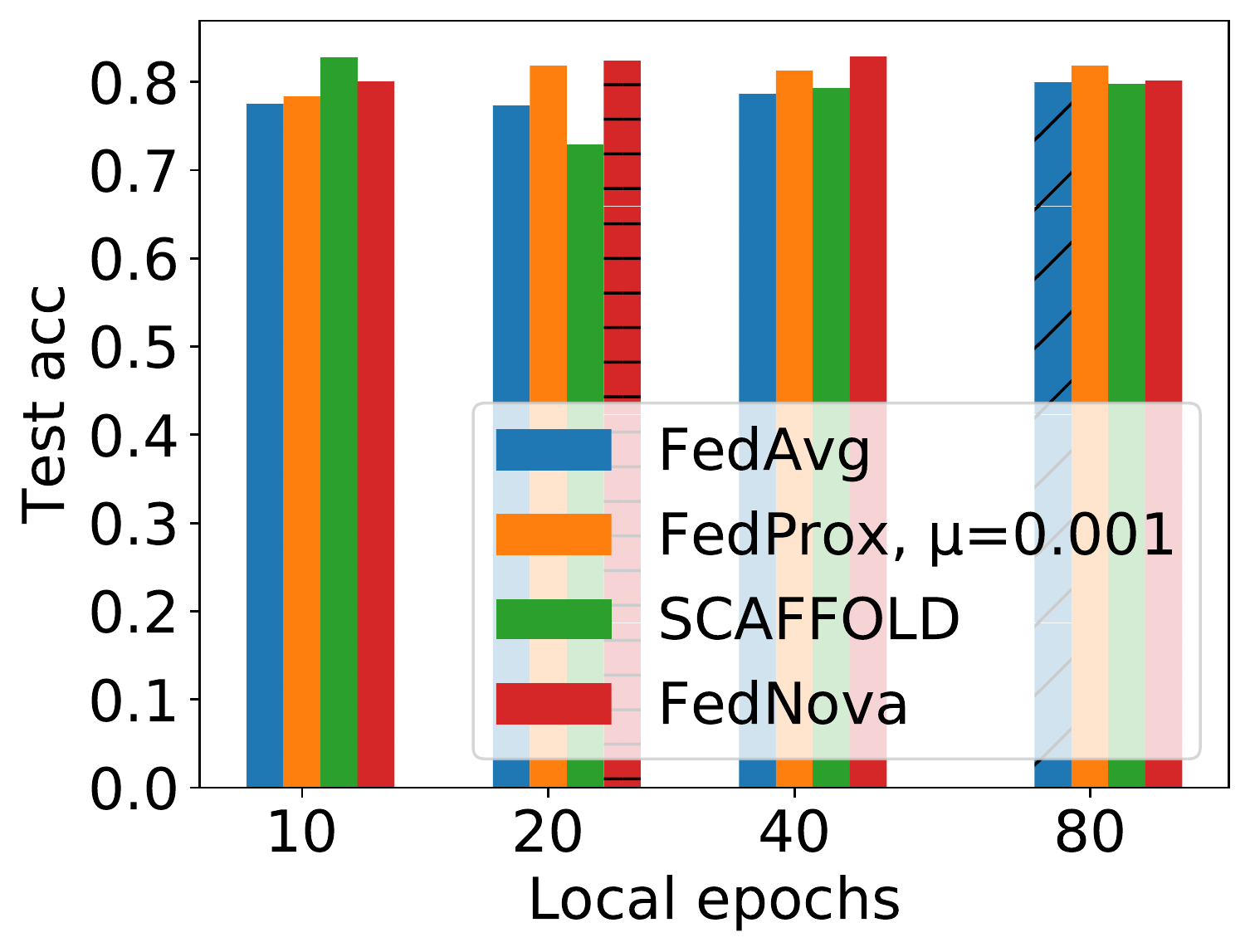}}
    \subfloat[$\hat{\mb{x}} \sim Gau(0.1)$]{\includegraphics[width=0.33\textwidth]{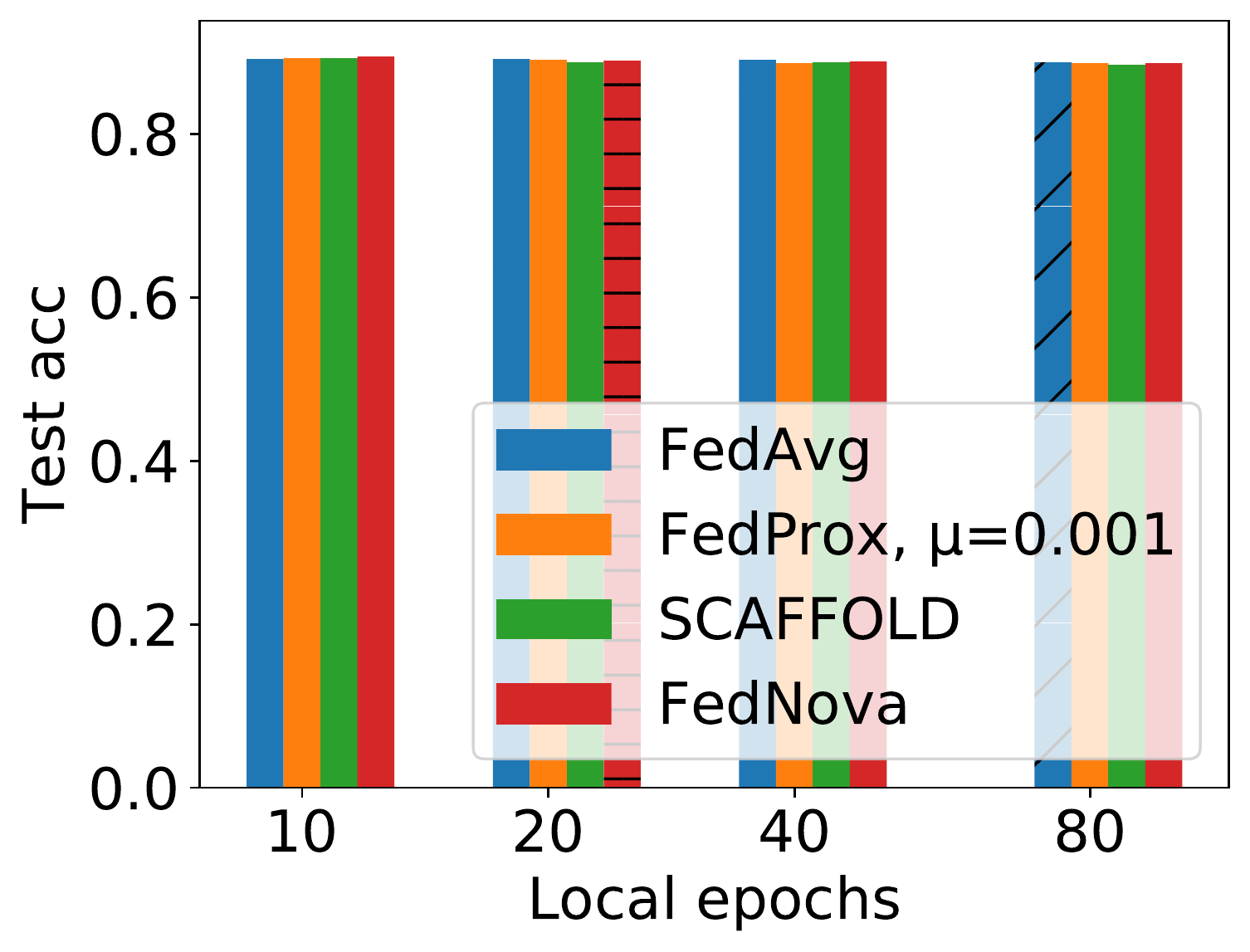}}
    \subfloat[$p \sim Dir(0.5)$]{\includegraphics[width=0.33\textwidth]{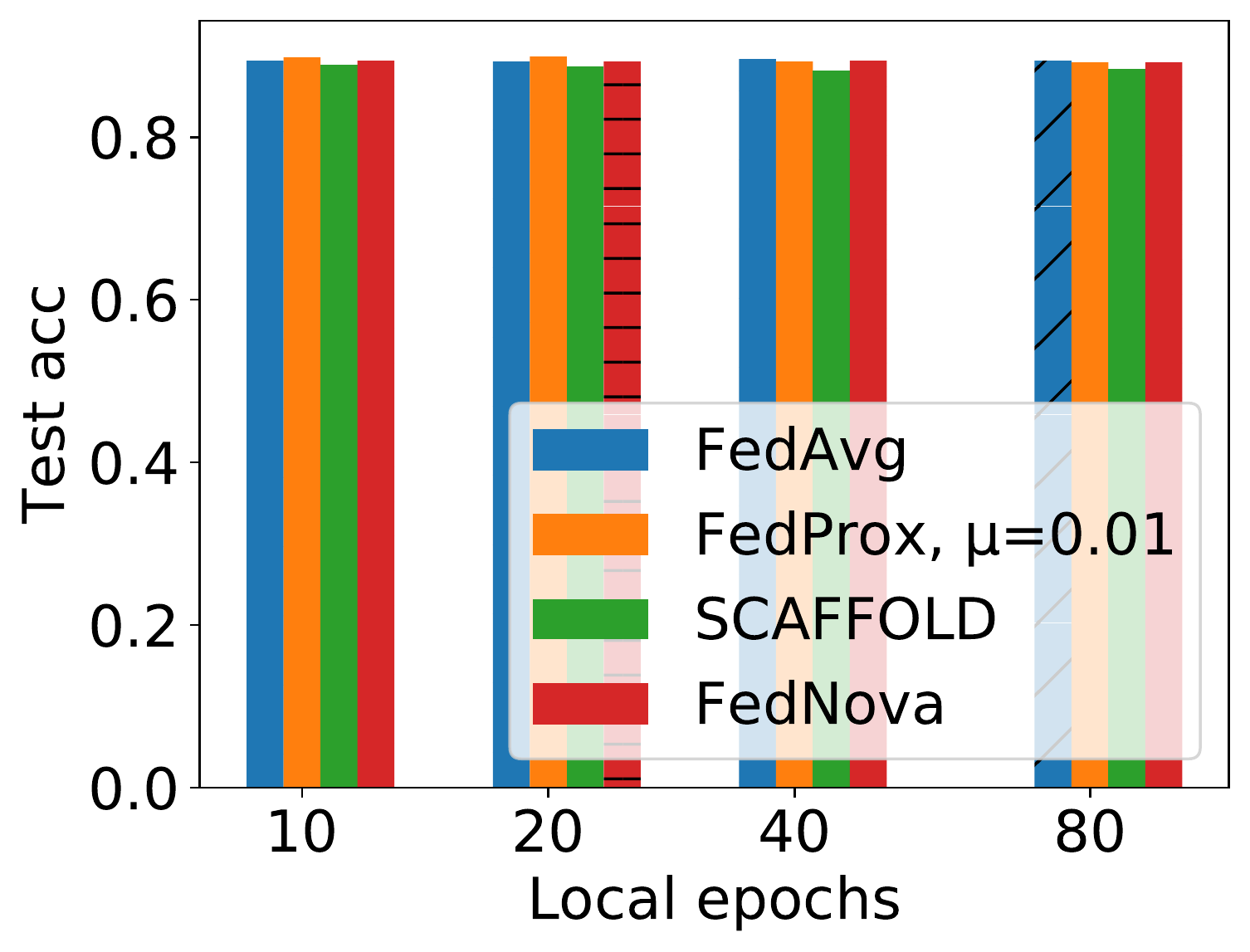}}
    \caption{The test accuracy with different number of local epochs on FMNIST.}
    \label{fig:fmnist_epoch}
\end{figure*}

\begin{figure*}
    \centering
    \subfloat[$p_k \sim Dir(0.5)$]{\includegraphics[width=0.33\textwidth]{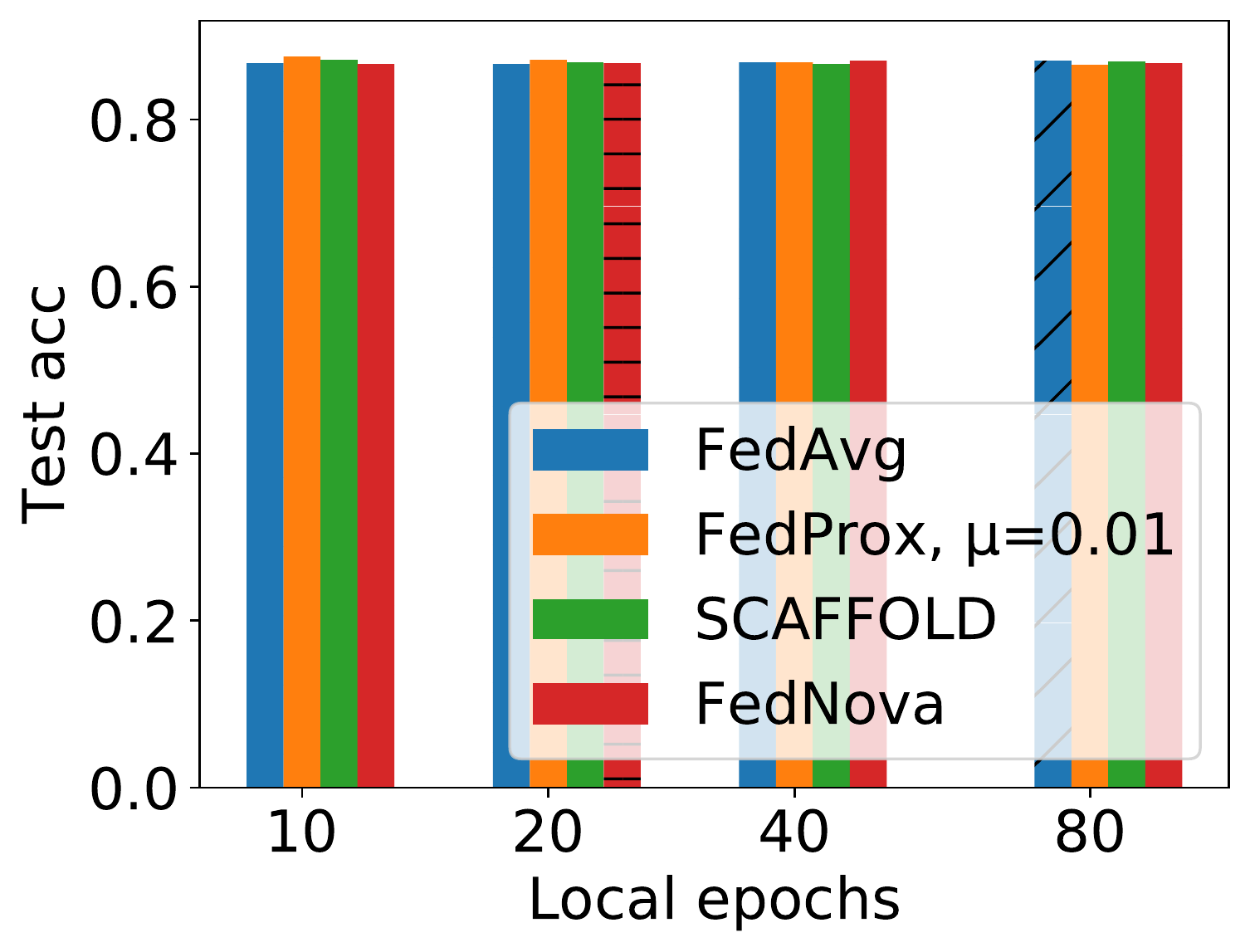}}
    \subfloat[$\#C=1$]{\includegraphics[width=0.33\textwidth]{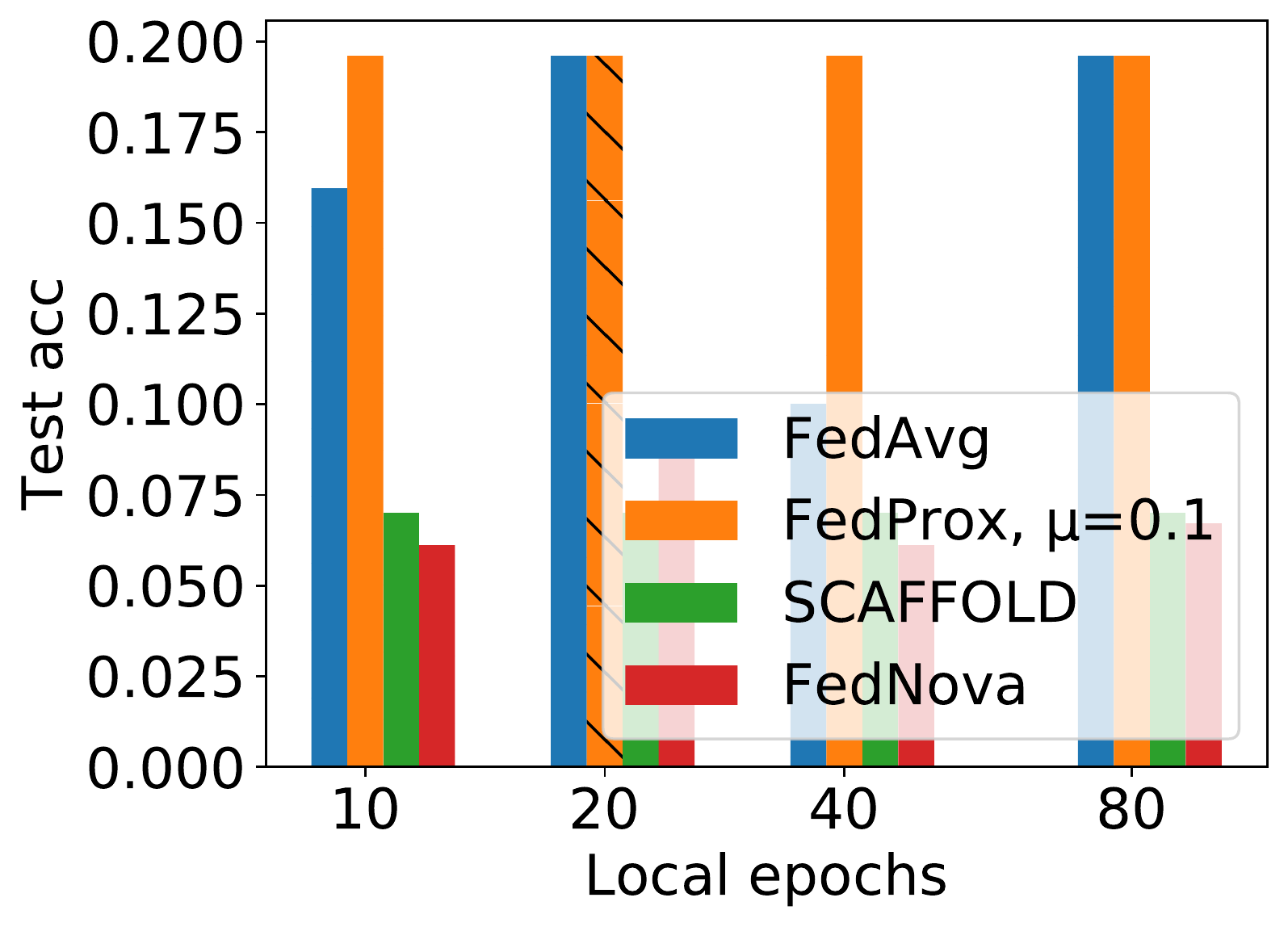}}
    \subfloat[$\#C=2$]{\includegraphics[width=0.33\textwidth]{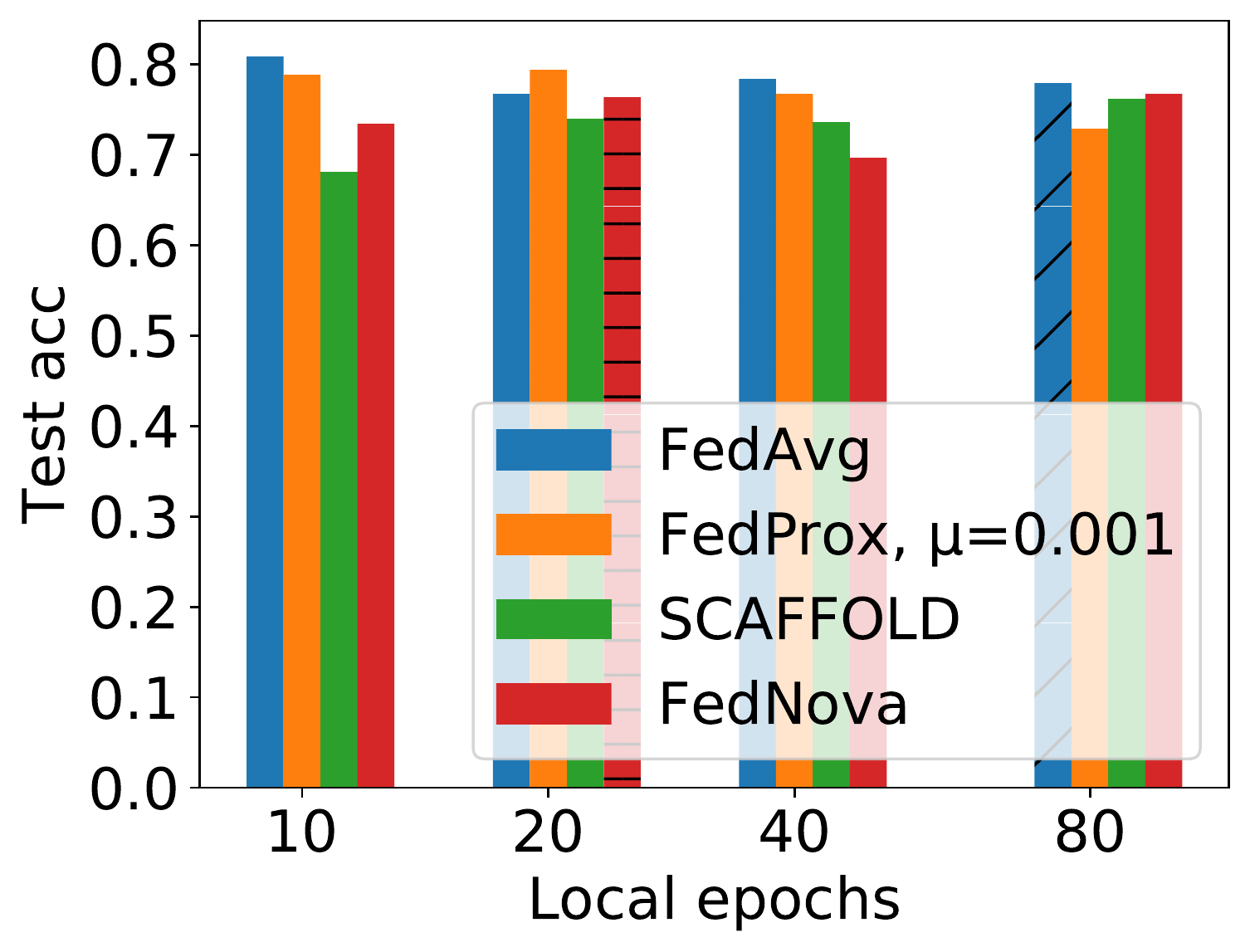}}
    \hfill
    \subfloat[$\#C=3$]{\includegraphics[width=0.33\textwidth]{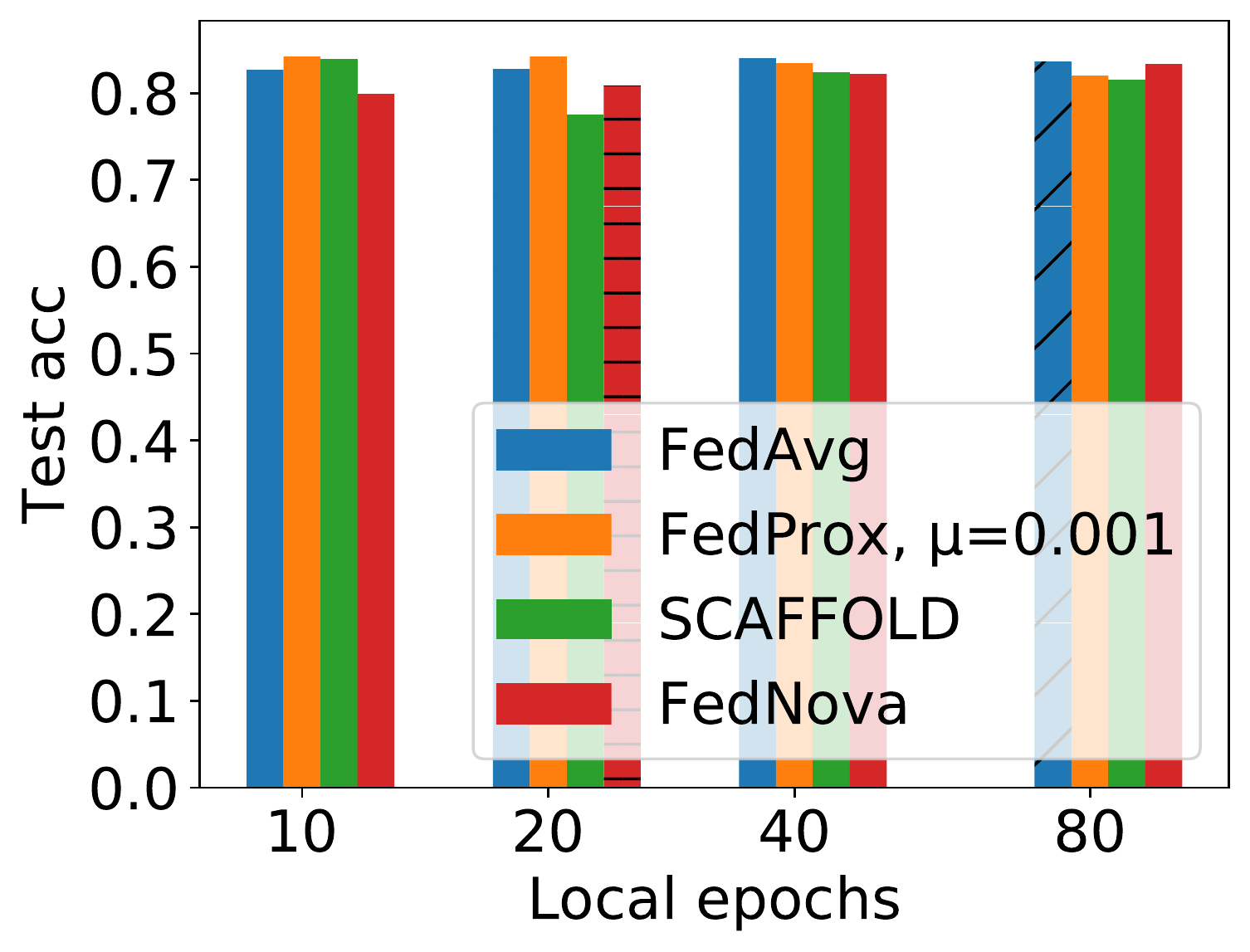}}
    \subfloat[$\hat{\mb{x}} \sim Gau(0.1)$]{\includegraphics[width=0.33\textwidth]{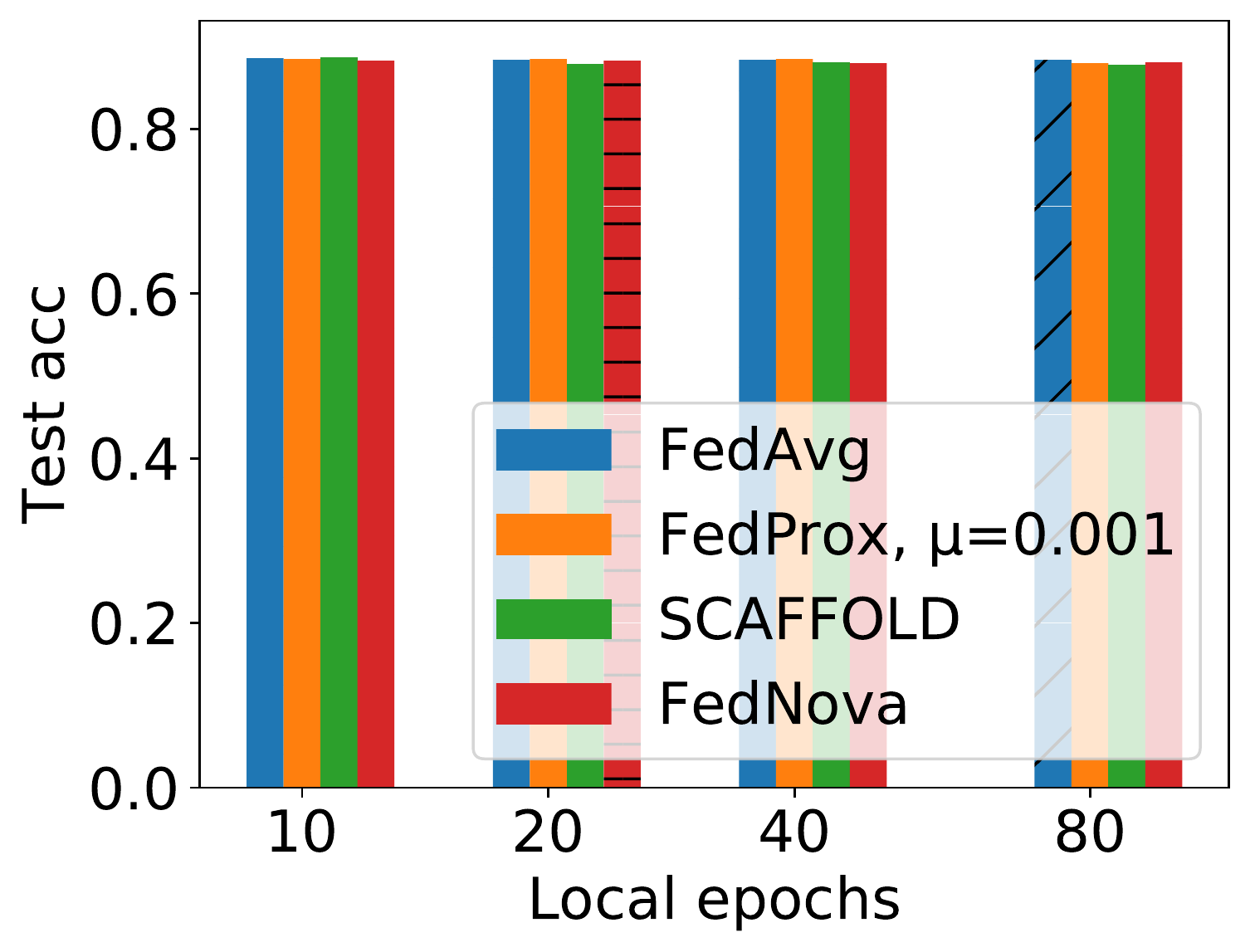}}
    \subfloat[$p \sim Dir(0.5)$]{\includegraphics[width=0.33\textwidth]{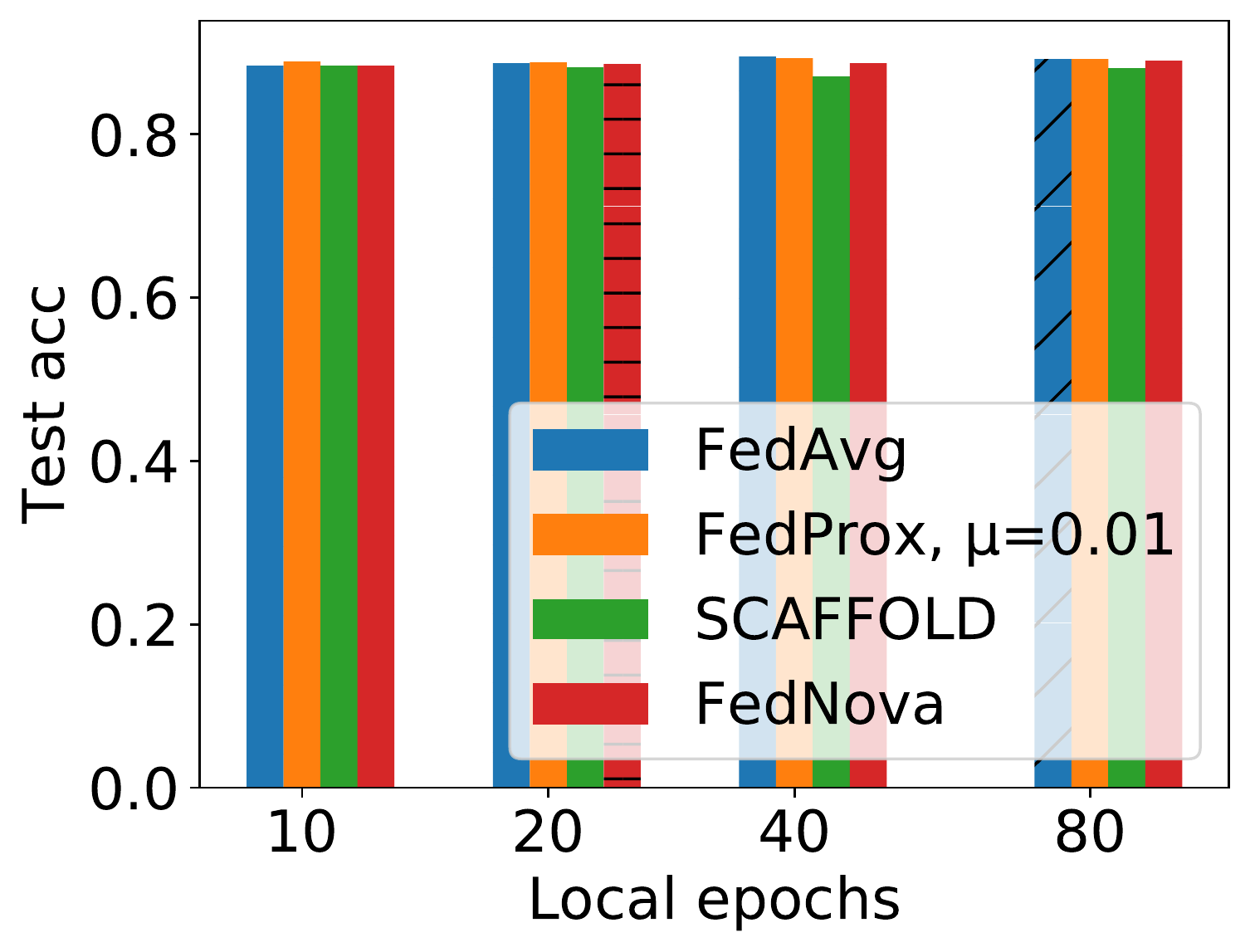}}
    \caption{The test accuracy with different number of local epochs on SVHN.}
    \label{fig:svhn_epoch}
\end{figure*}

\begin{figure*}
    \centering
    \subfloat[FCUBE]{\includegraphics[width=0.33\textwidth]{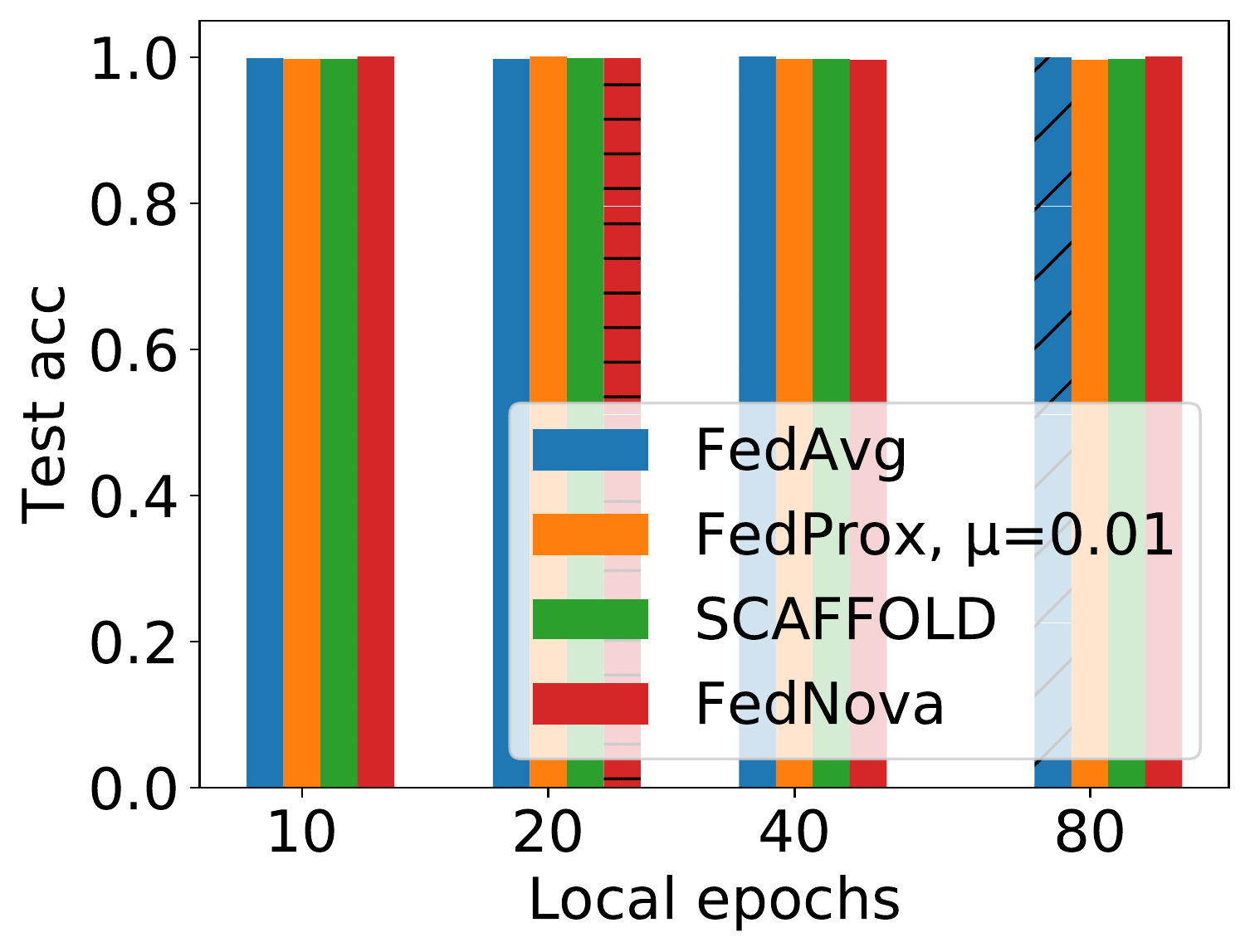}}
    \subfloat[FEMNIST]{\includegraphics[width=0.33\textwidth]{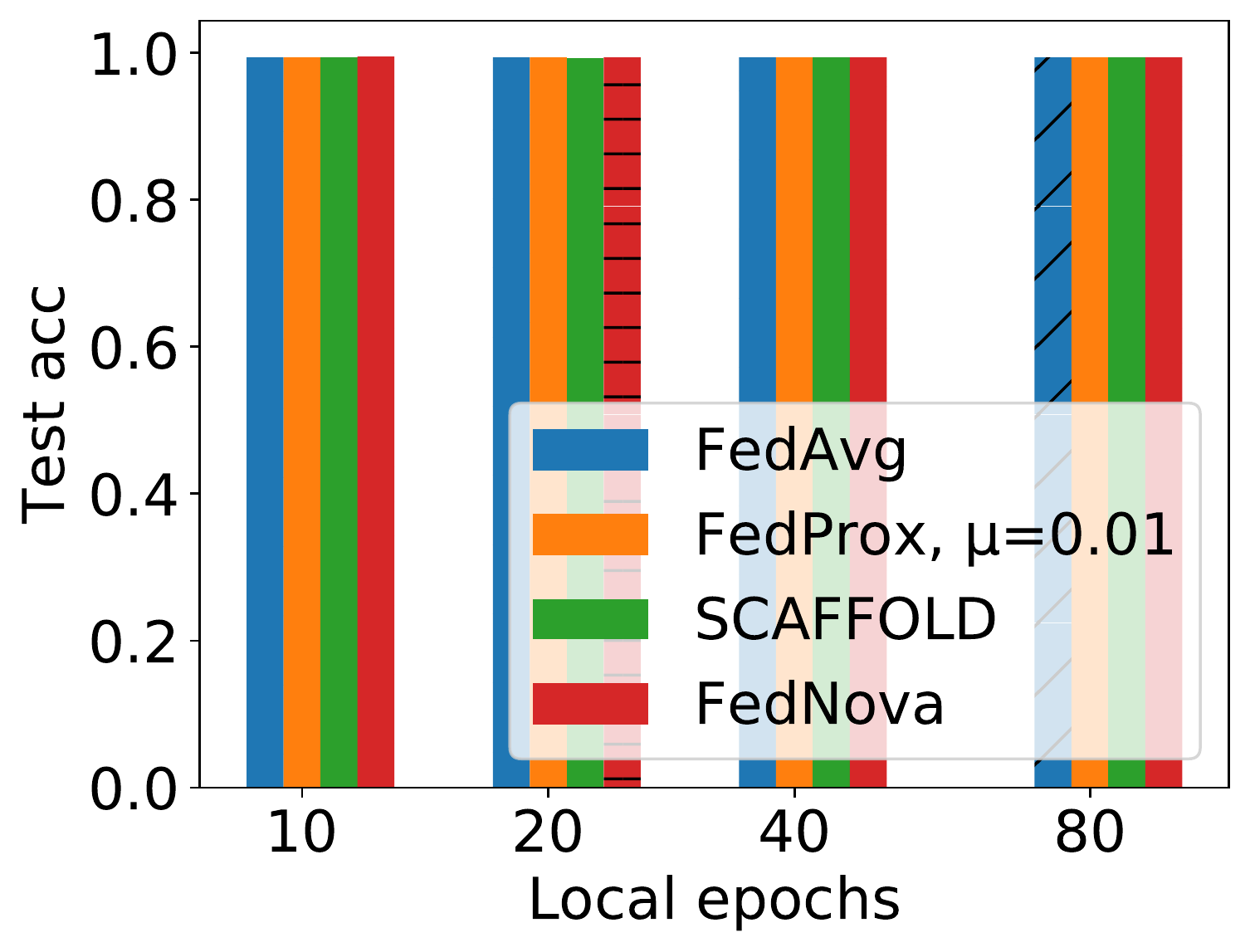}}
    \caption{The test accuracy with different number of local epochs on FCUBE and FEMNIST.}
    \label{fig:cube_femnist_epoch}
\end{figure*}

Figures \ref{fig:cifar10_curve_otherfour}, \ref{fig:mnist_epoch}, \ref{fig:fmnist_epoch}, \ref{fig:svhn_epoch}, and \ref{fig:cube_femnist_epoch} show the accuracy with different number of local epochs on the studied datasets except CIFAR-10.

\begin{figure*}[!]
    \centering
    \subfloat[$\#C=1$]{\includegraphics[width=0.24\textwidth]{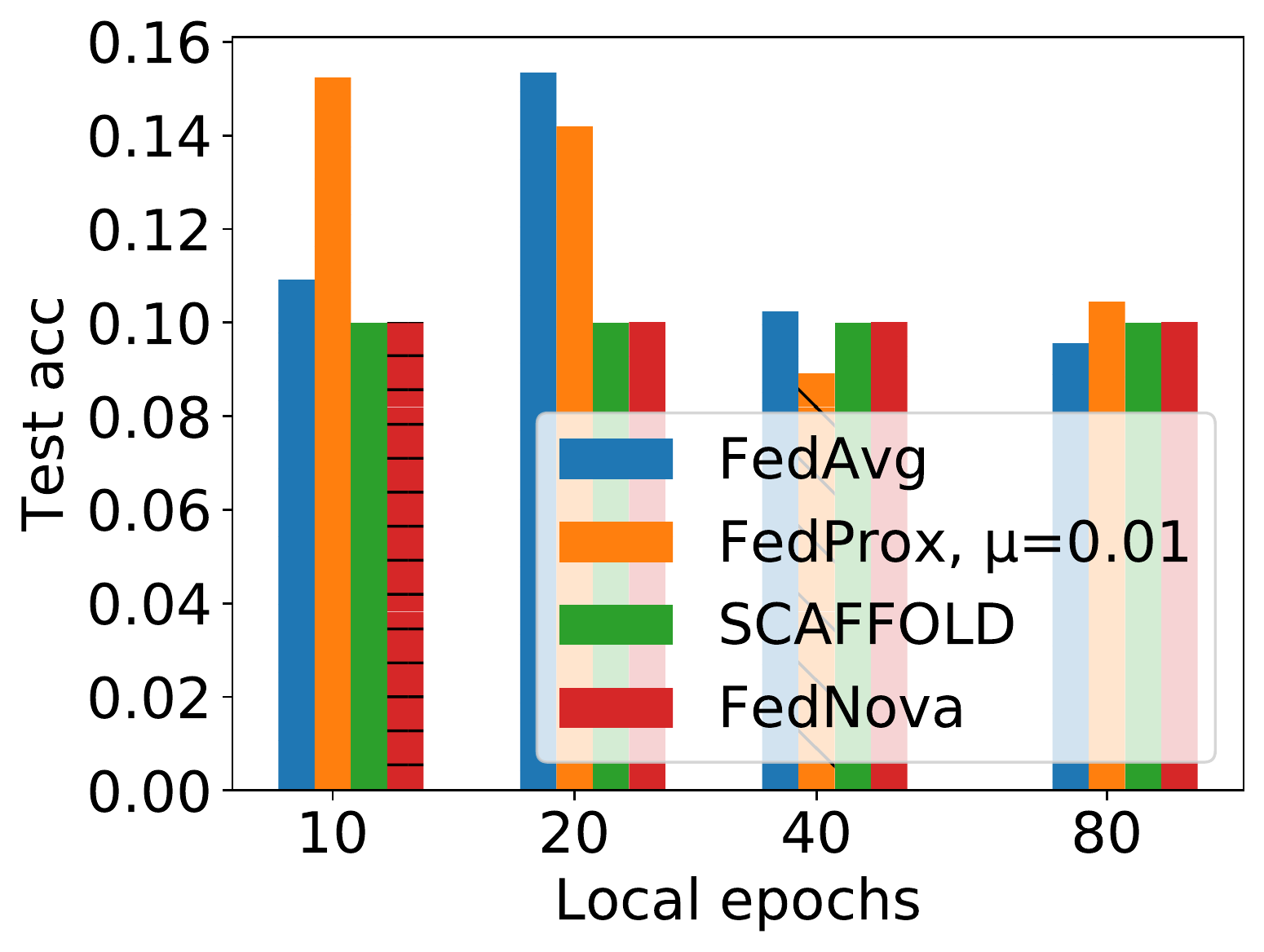}}
    \subfloat[$\#C=2$]{\includegraphics[width=0.24\textwidth]{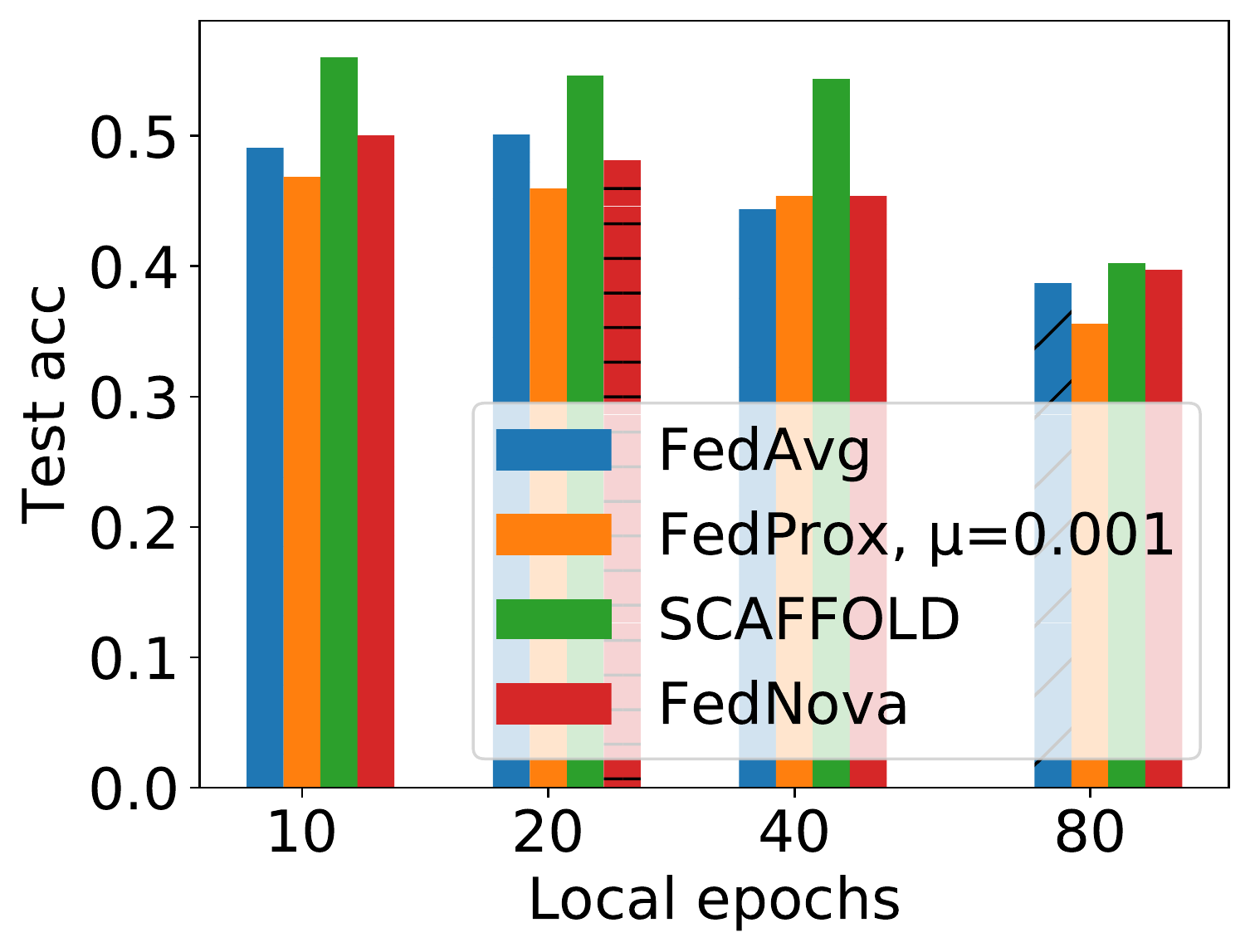}}
    \hfill
    \subfloat[$\#C=3$]{\includegraphics[width=0.24\textwidth]{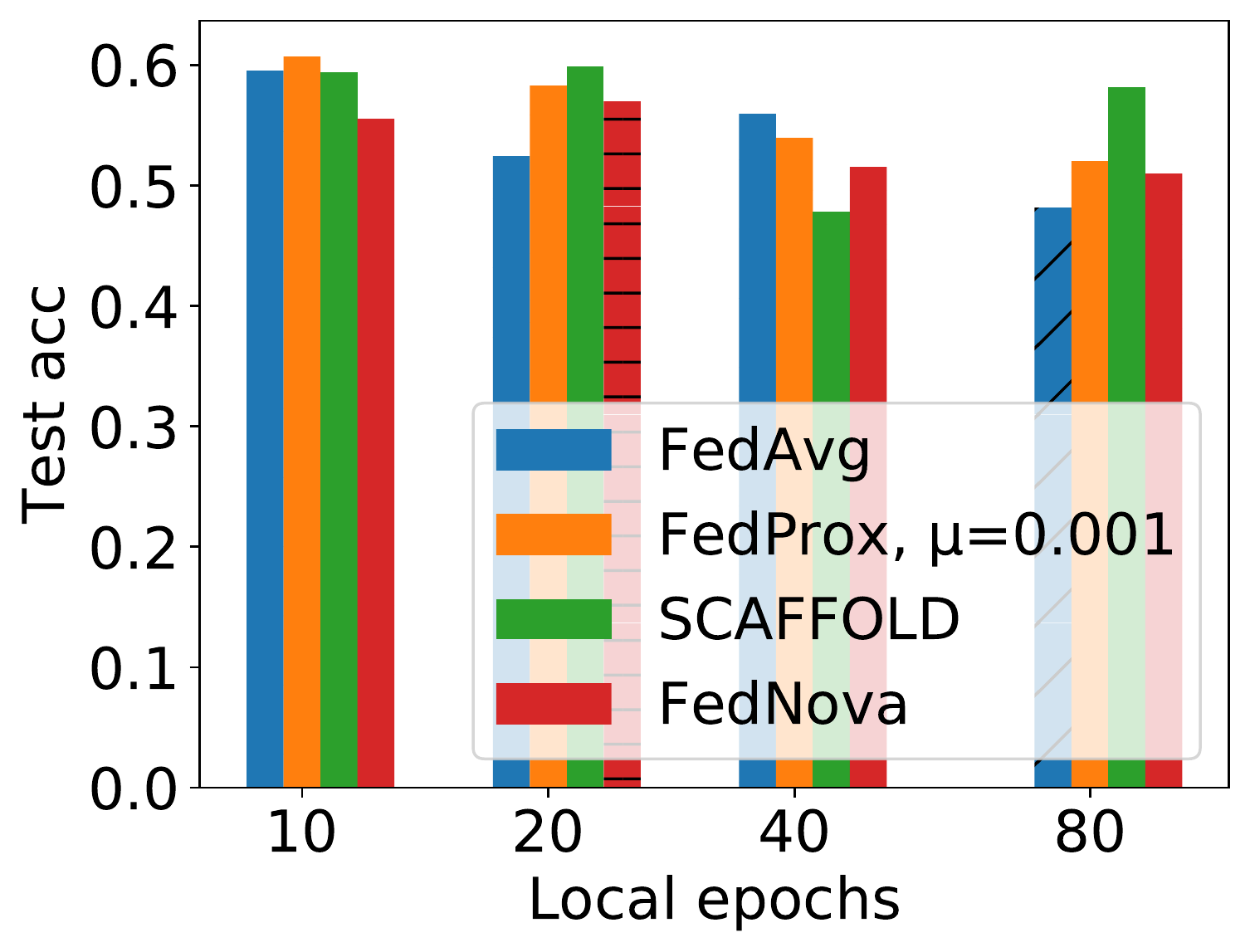}}
    \subfloat[$q \sim Dir(0.5)$]{\includegraphics[width=0.24\textwidth]{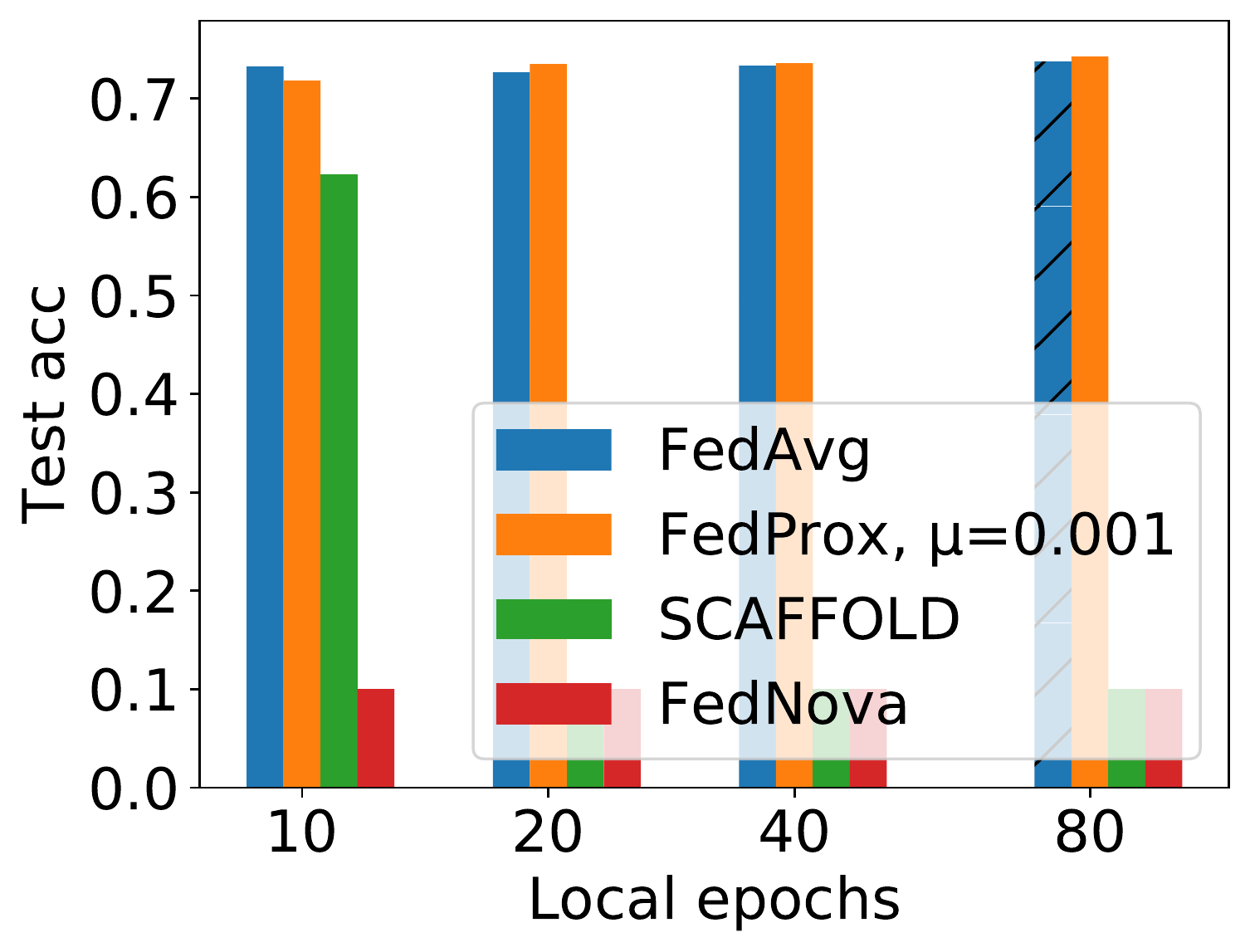}}
    \caption{The test accuracy with different numbers of local epochs on CIFAR-10.}
    \label{fig:cifar10_epoch_otherfour}
\end{figure*}

\subsection{Party Sampling}
\begin{figure*}[t]
    \centering
    \subfloat[$\#C=1$]{\includegraphics[width=0.24\textwidth]{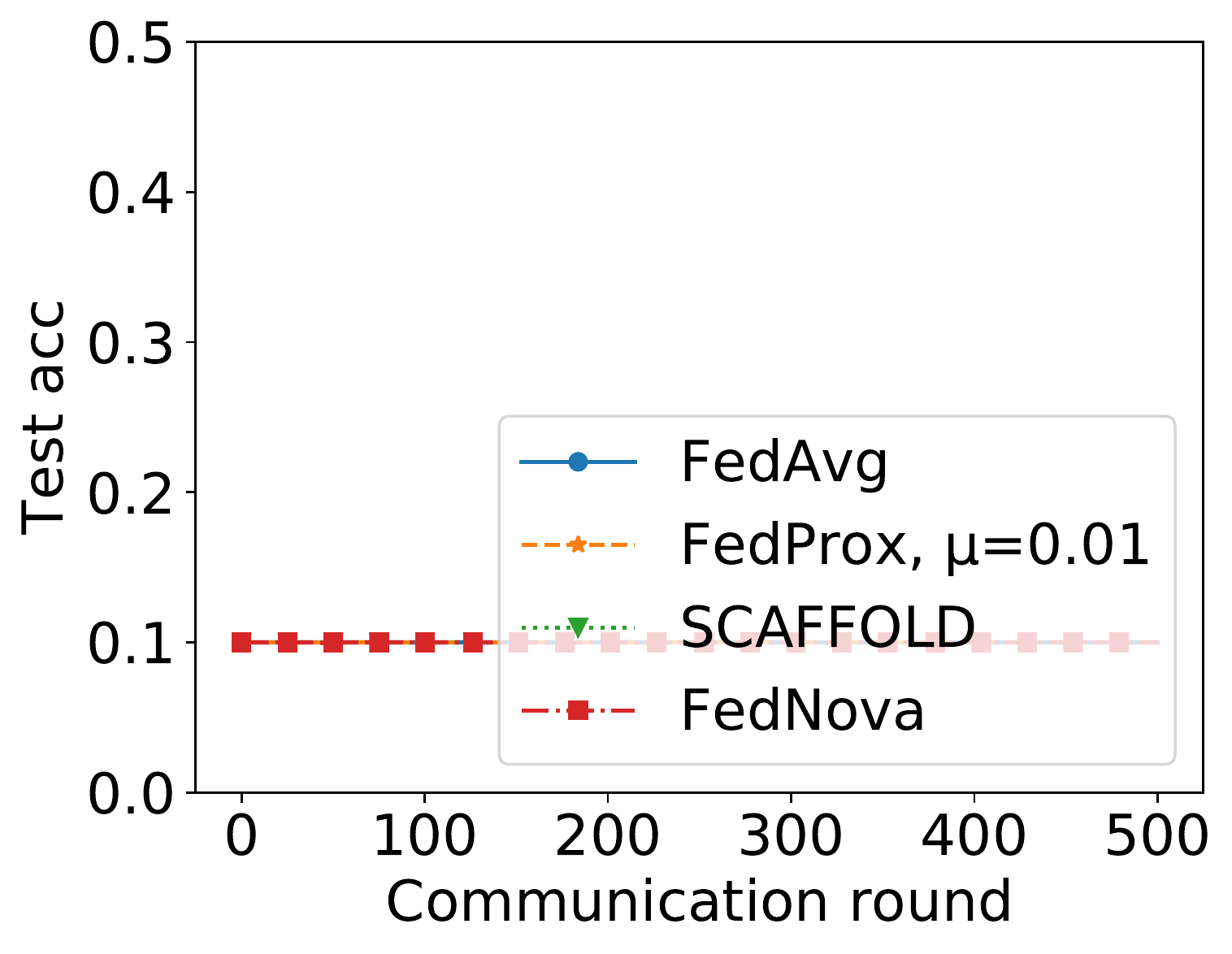}}
    \subfloat[$\#C=2$]{\includegraphics[width=0.24\textwidth]{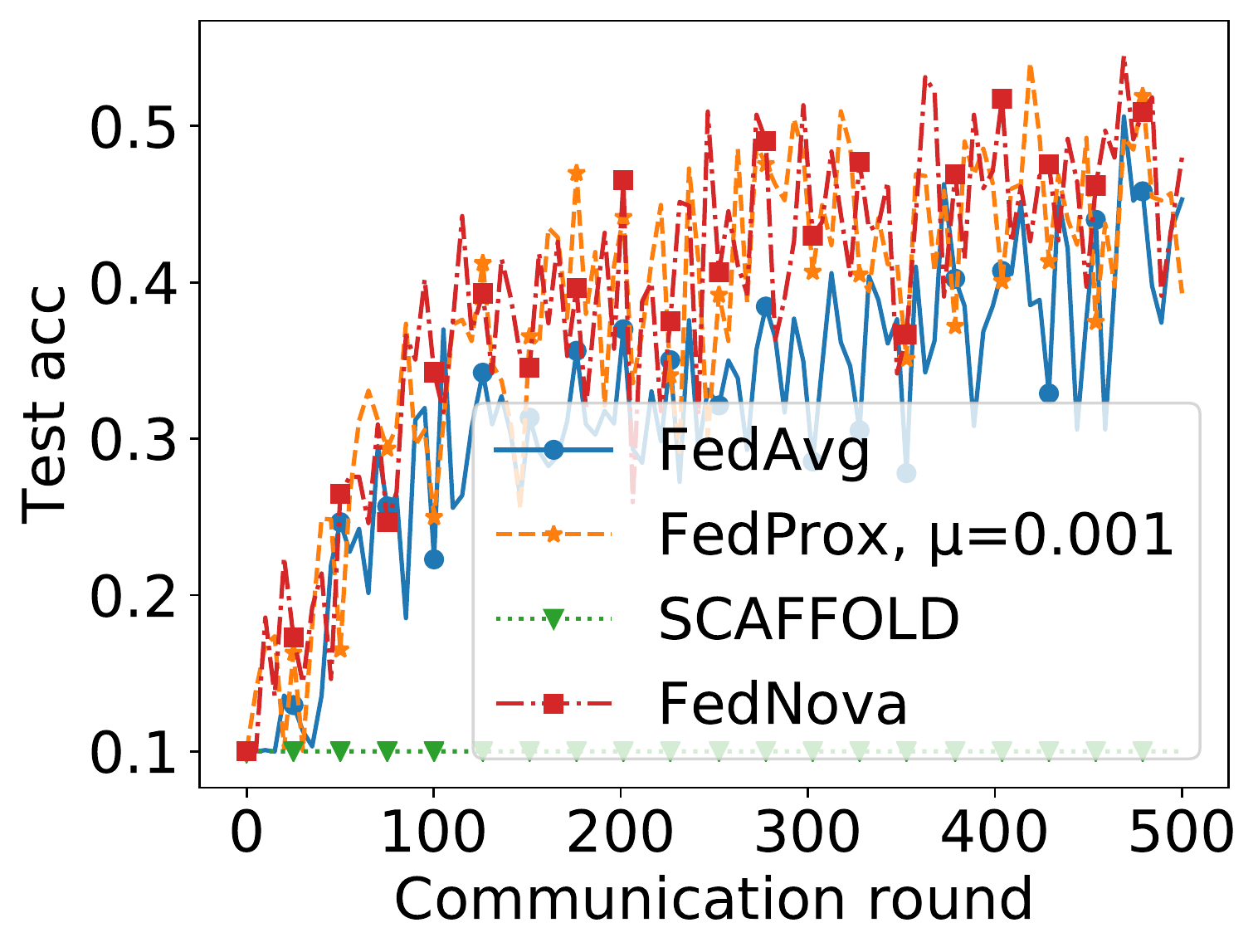}}
    \subfloat[$\#C=3$]{\includegraphics[width=0.24\textwidth]{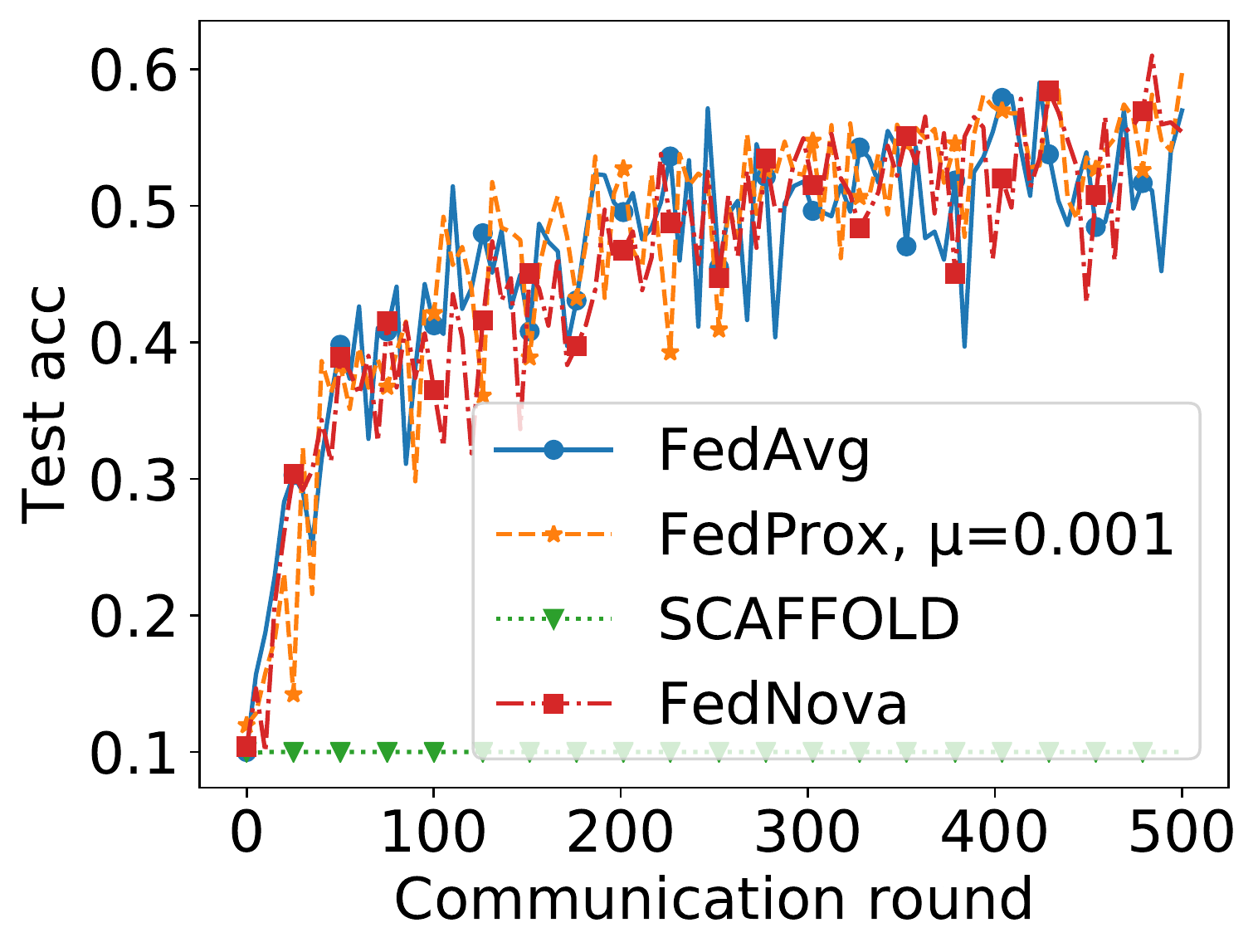}}
    \subfloat[IID\label{fig:100partyiid}]{\includegraphics[width=0.24\textwidth]{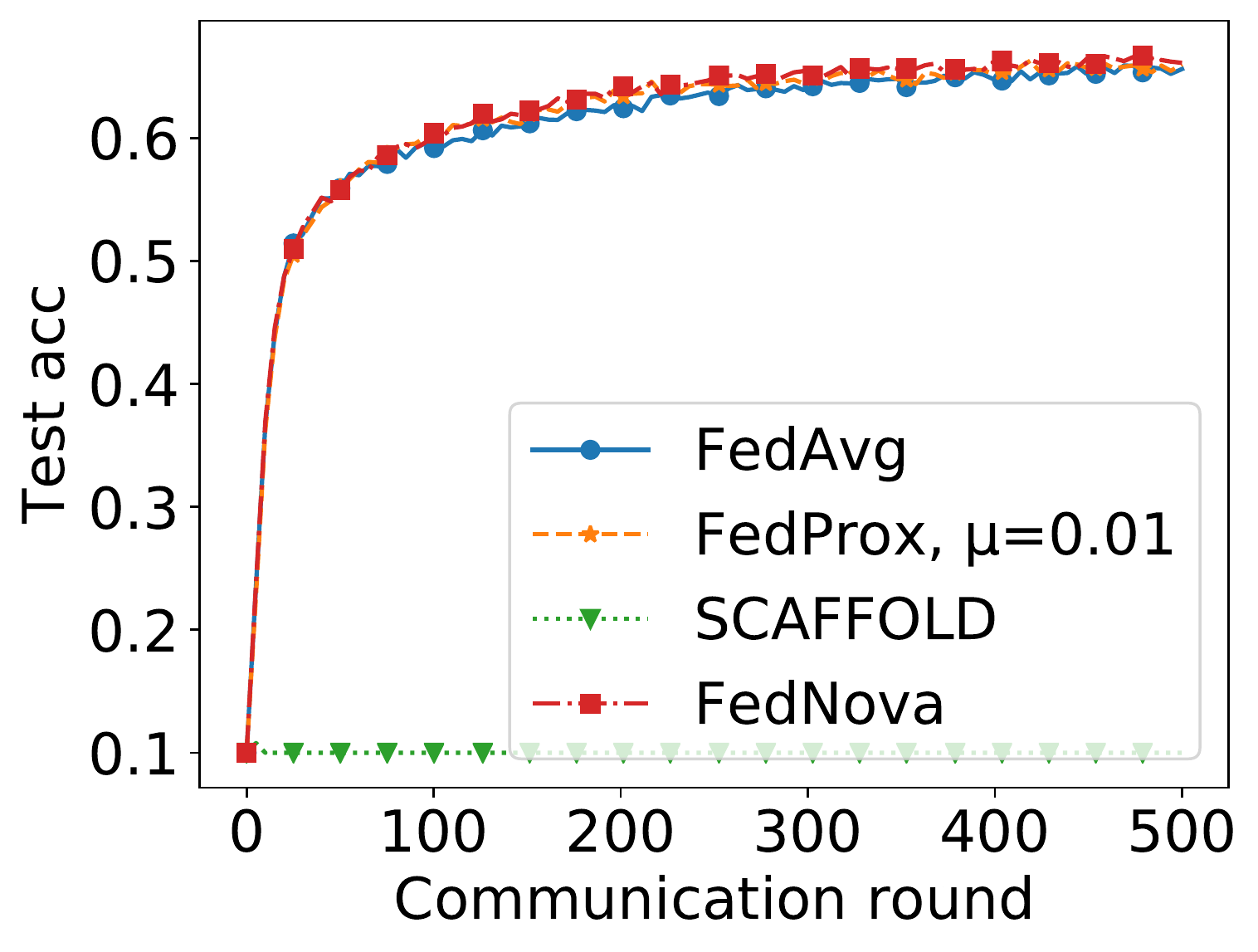}}
    \caption{The training curves of different approaches on CIFAR-10 with 100 parties and sample fraction 0.1.}
    \label{fig:100_parties_otherfour}
\end{figure*}

Figure \ref{fig:100_parties_otherfour} shows the training curves of the studied approaches on CIFAR-10 under the party sampling setting.

\begin{figure*}[!]
    \centering
    \subfloat[FedAvg]{\includegraphics[width=0.24\textwidth]{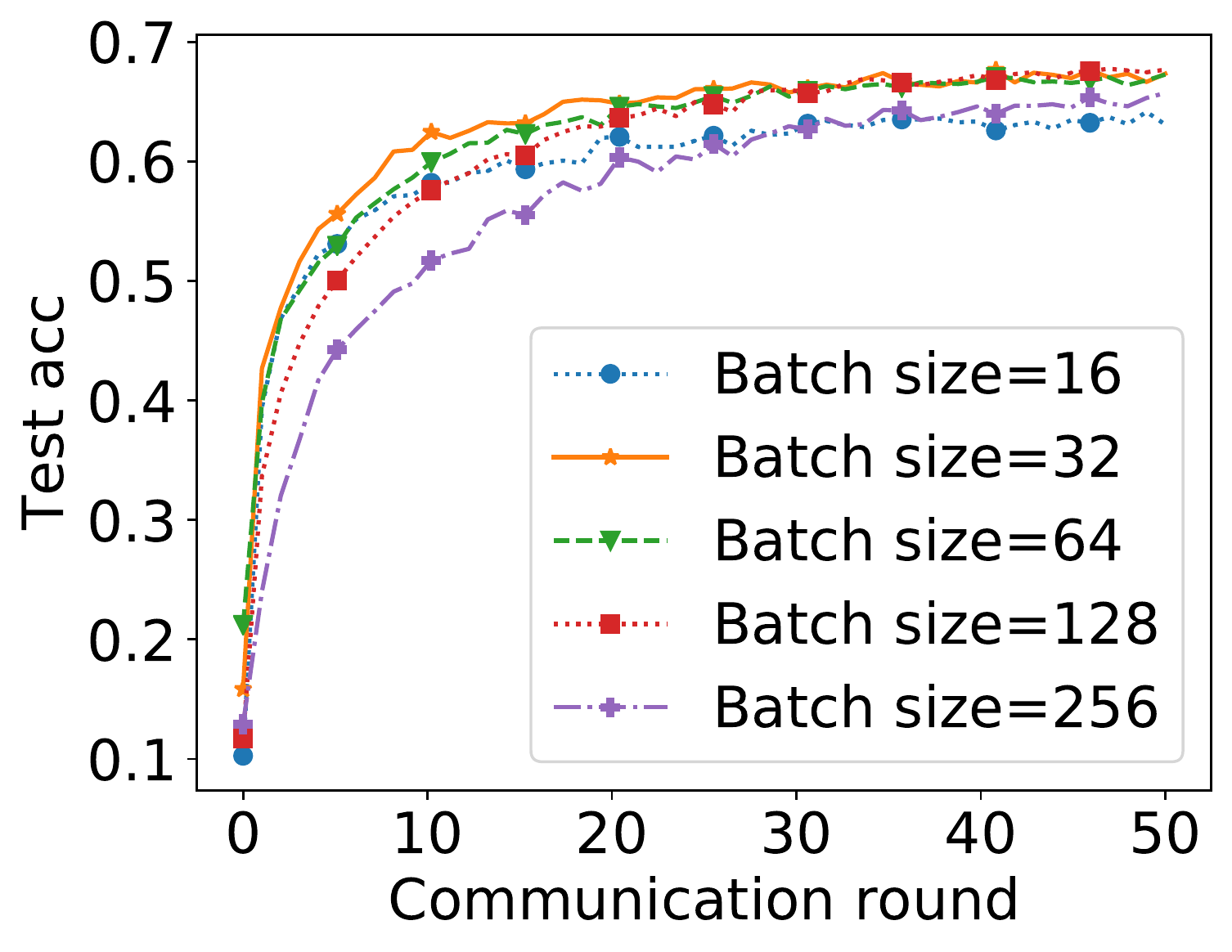}}
    \subfloat[FedProx ($\mu=0.01$)]{\includegraphics[width=0.24\textwidth]{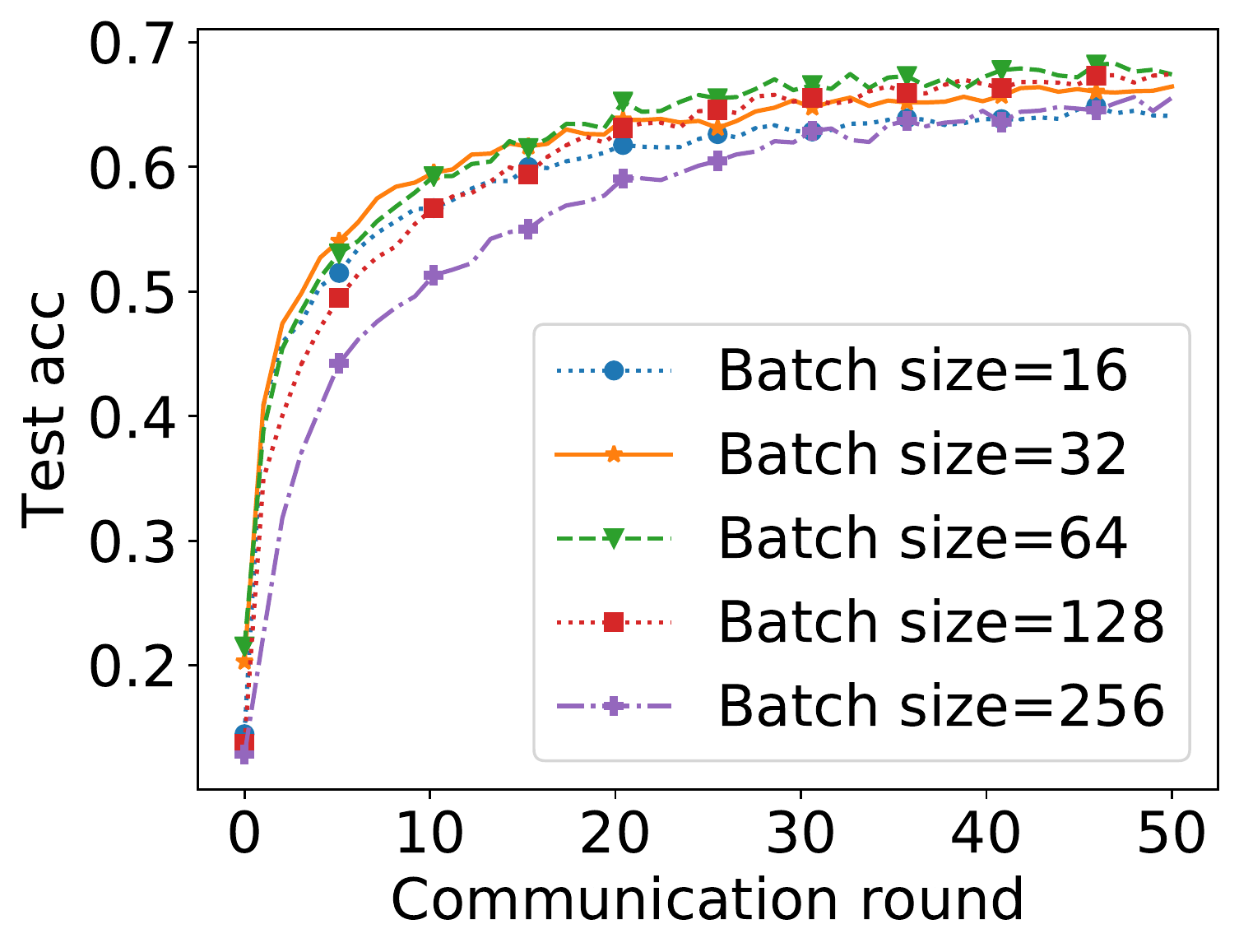}}
    \subfloat[SCAFFOLD]{\includegraphics[width=0.24\textwidth]{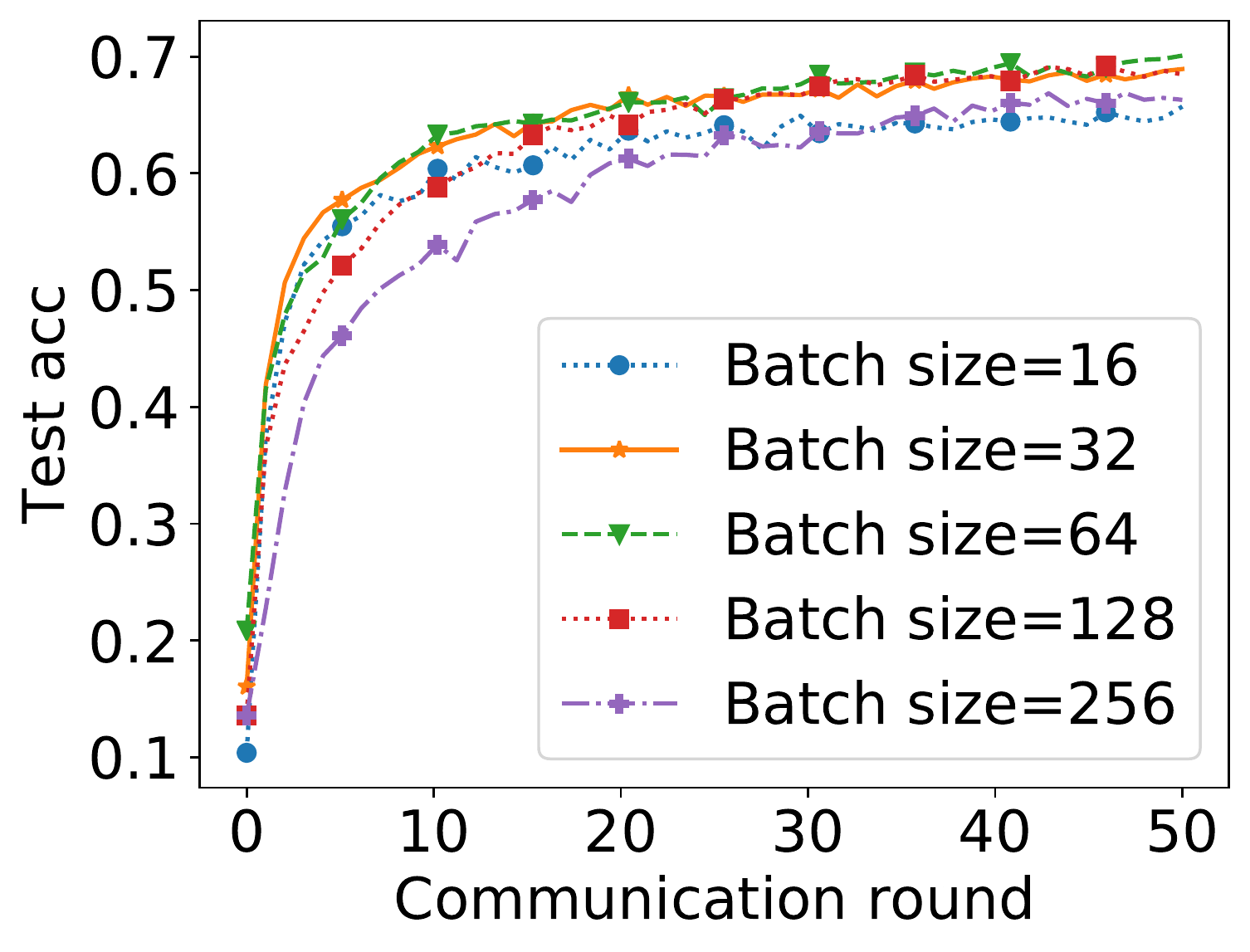}}
    \subfloat[FedNova]{\includegraphics[width=0.24\textwidth]{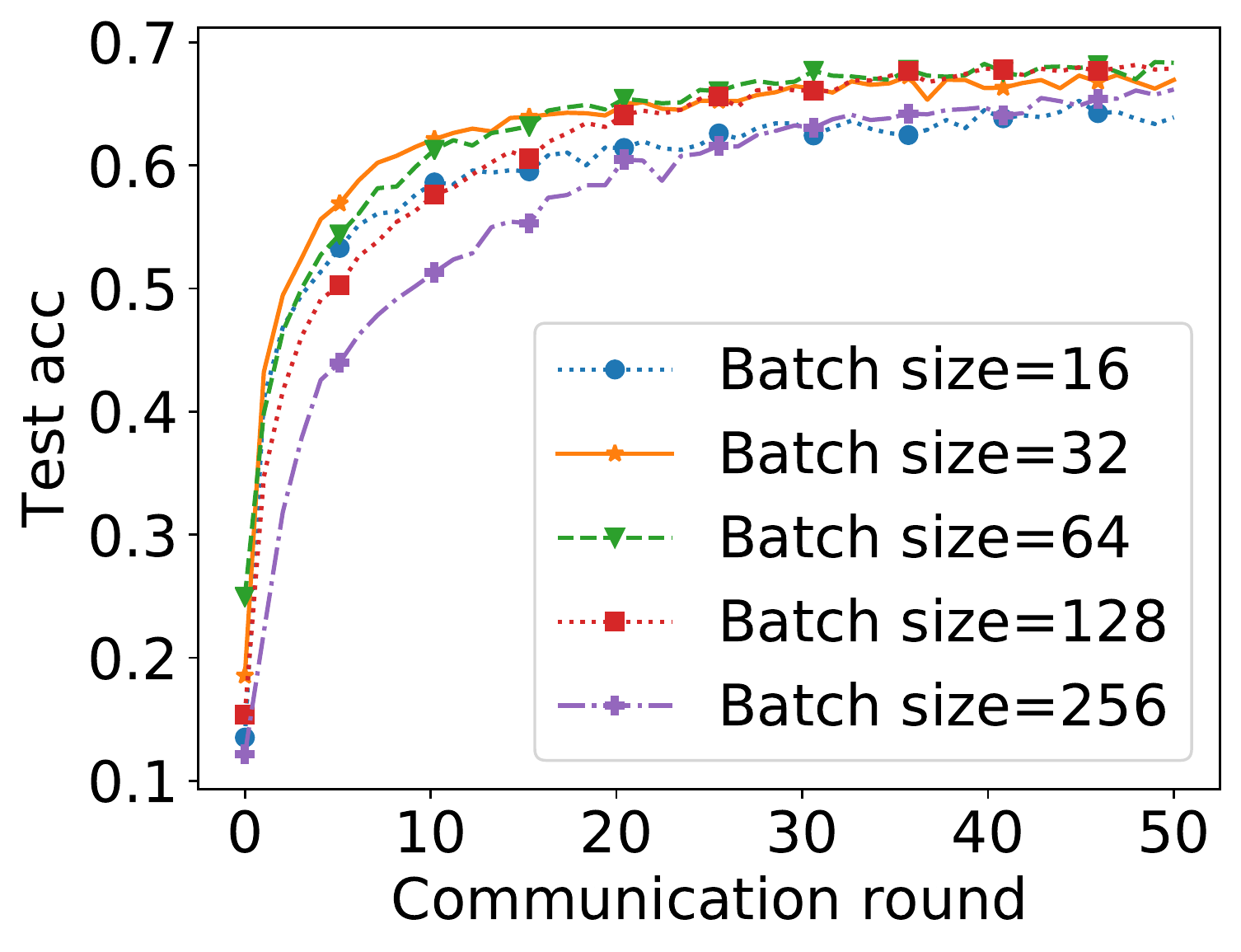}}
    \caption{The training curves of different batch sizes on CIFAR-10 under $p_k \sim Dir(0.5)$ partition.}
    \label{fig:batch-size}
\end{figure*}

\subsection{Batch Size}
\label{sec:batchsize}
\noindent \fbox{\parbox{0.98\linewidth}{
\tb{Finding (10):} The heterogeneity of local data does not appear to influence the behaviors of different choices of batch sizes.
}}

Batch size is an important hyper-parameter in deep learning. We choose FedAvg and FedProx as the representative algorithms and study the effect of batch size in FL by varying it from 16 to 256 as shown in Figure \ref{fig:batch-size}. Like centralized training, a large batch size slows down the learning process. Moreover, four studied algorithms have similar behaviours given different batch sizes. The results demonstrate that there is no clear relationship between the setting of batch size and the heterogeneity of local data. The knowledge of the behaviors of different batch sizes still applies in the non-IID federated setting.

\begin{figure*}[!]
    \centering
    \subfloat[VGG-9 under $p_k \sim Dir(0.1)$ partition ]{\includegraphics[width=0.33\textwidth]{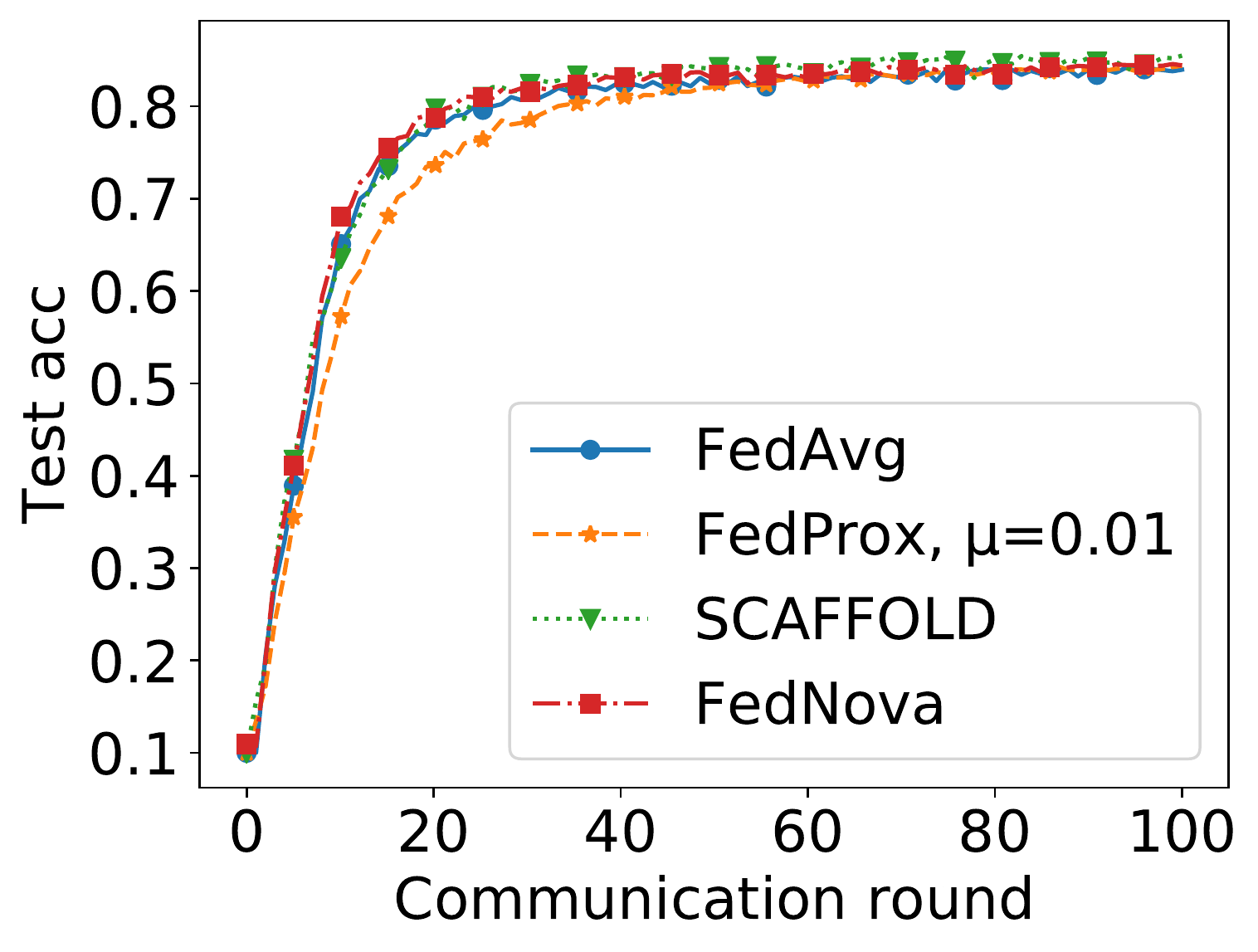}}
    \subfloat[VGG-9 under $\hat{\mb{x}} \sim Gau(0.1)$ partition]{\includegraphics[width=0.33\textwidth]{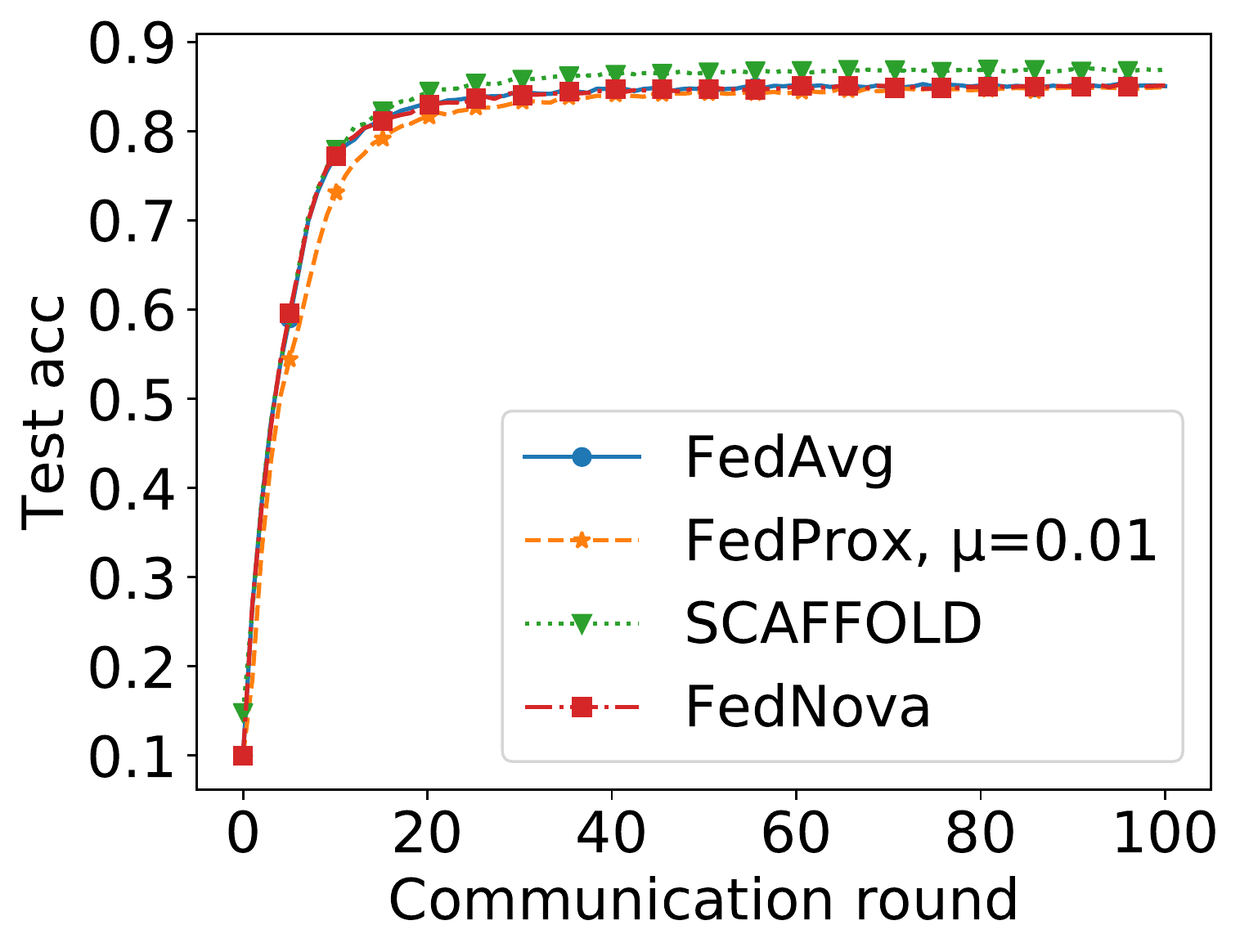}}
    \subfloat[VGG-9 under $q \sim Dir(0.1)$ partition]{\includegraphics[width=0.33\textwidth]{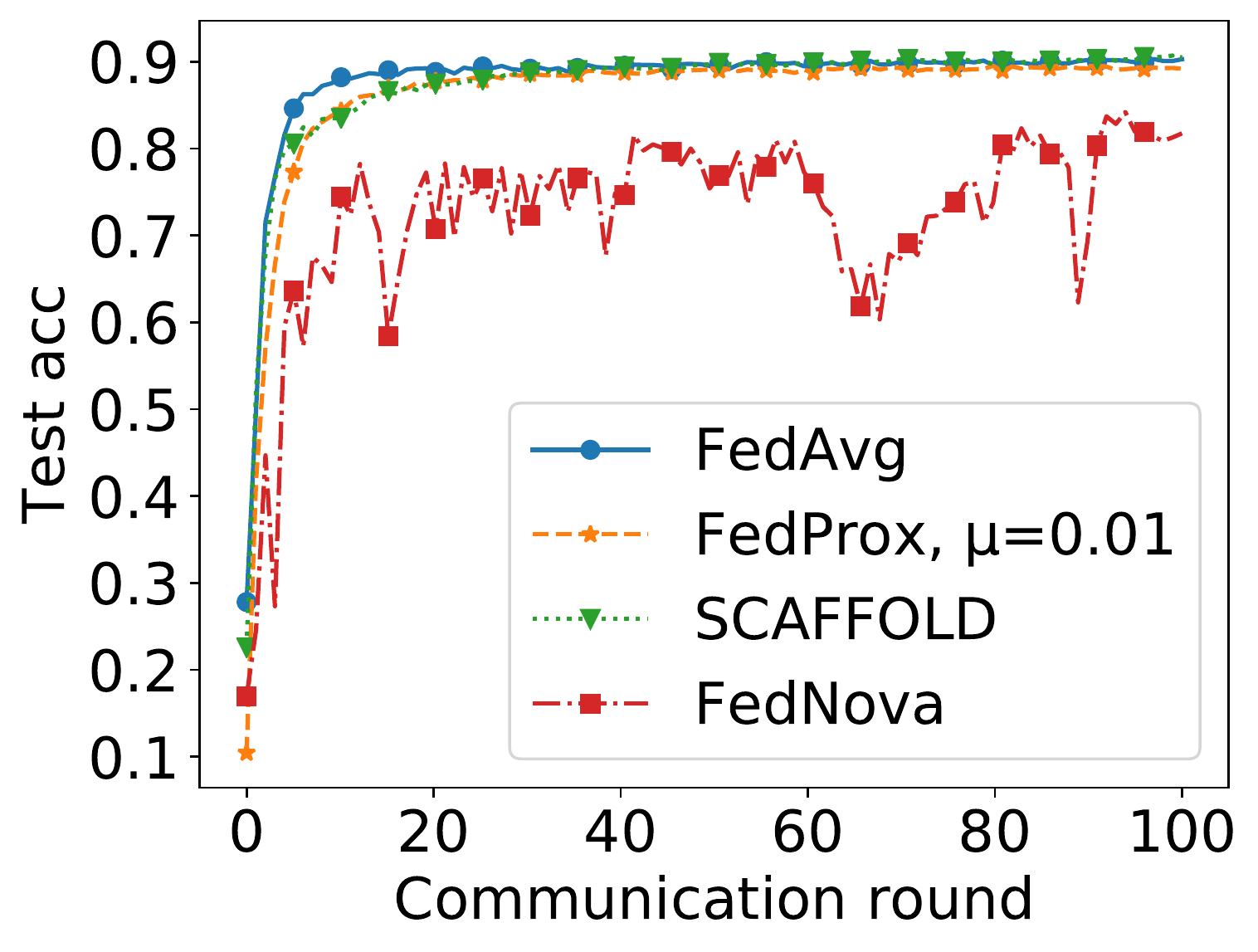}}
    \hfill
    \subfloat[ResNet-50 under $p_k \sim Dir(0.1)$ partition ]{\includegraphics[width=0.33\textwidth]{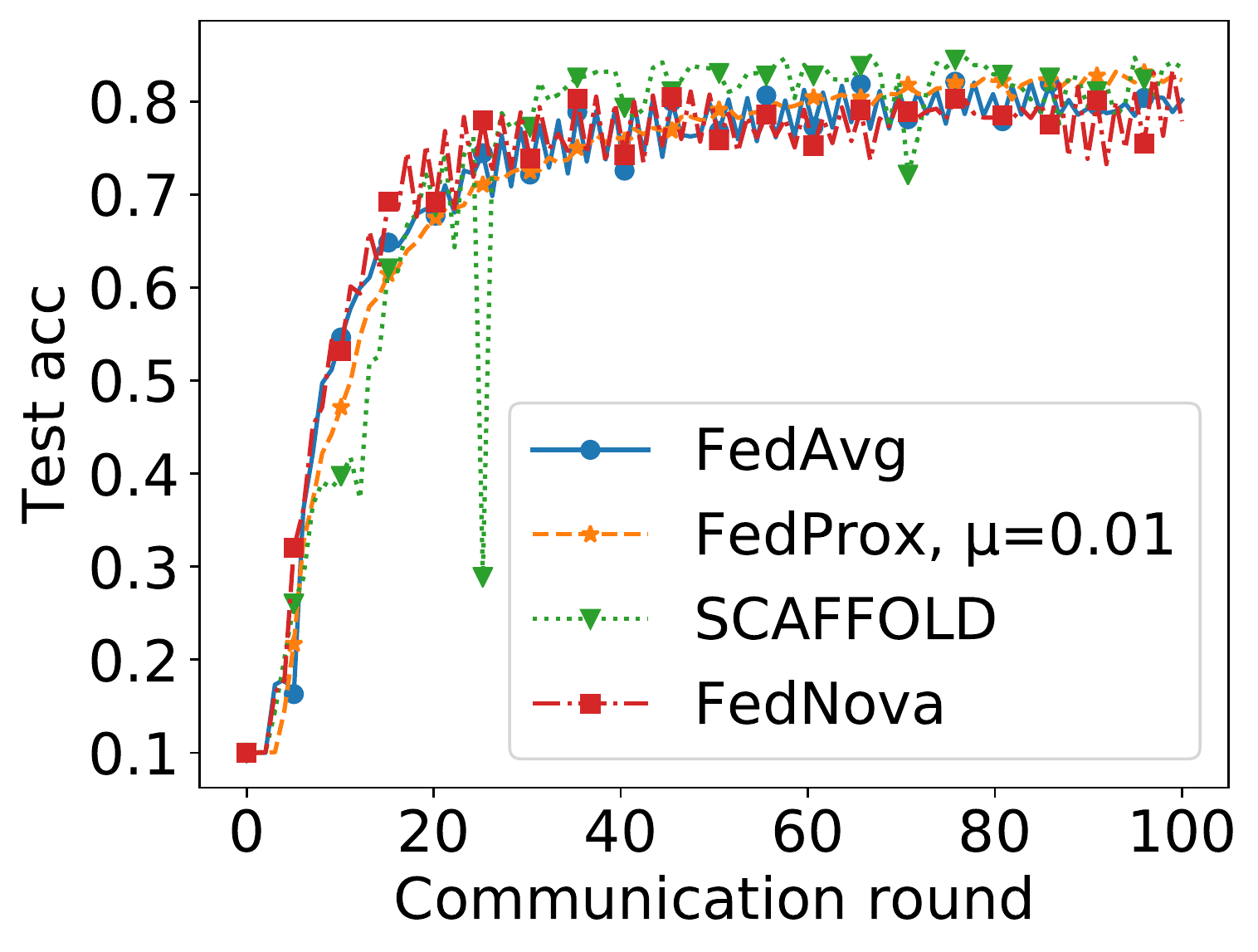}}
    \subfloat[ResNet-50 under $\hat{\mb{x}} \sim Gau(0.1)$ partition]{\includegraphics[width=0.33\textwidth]{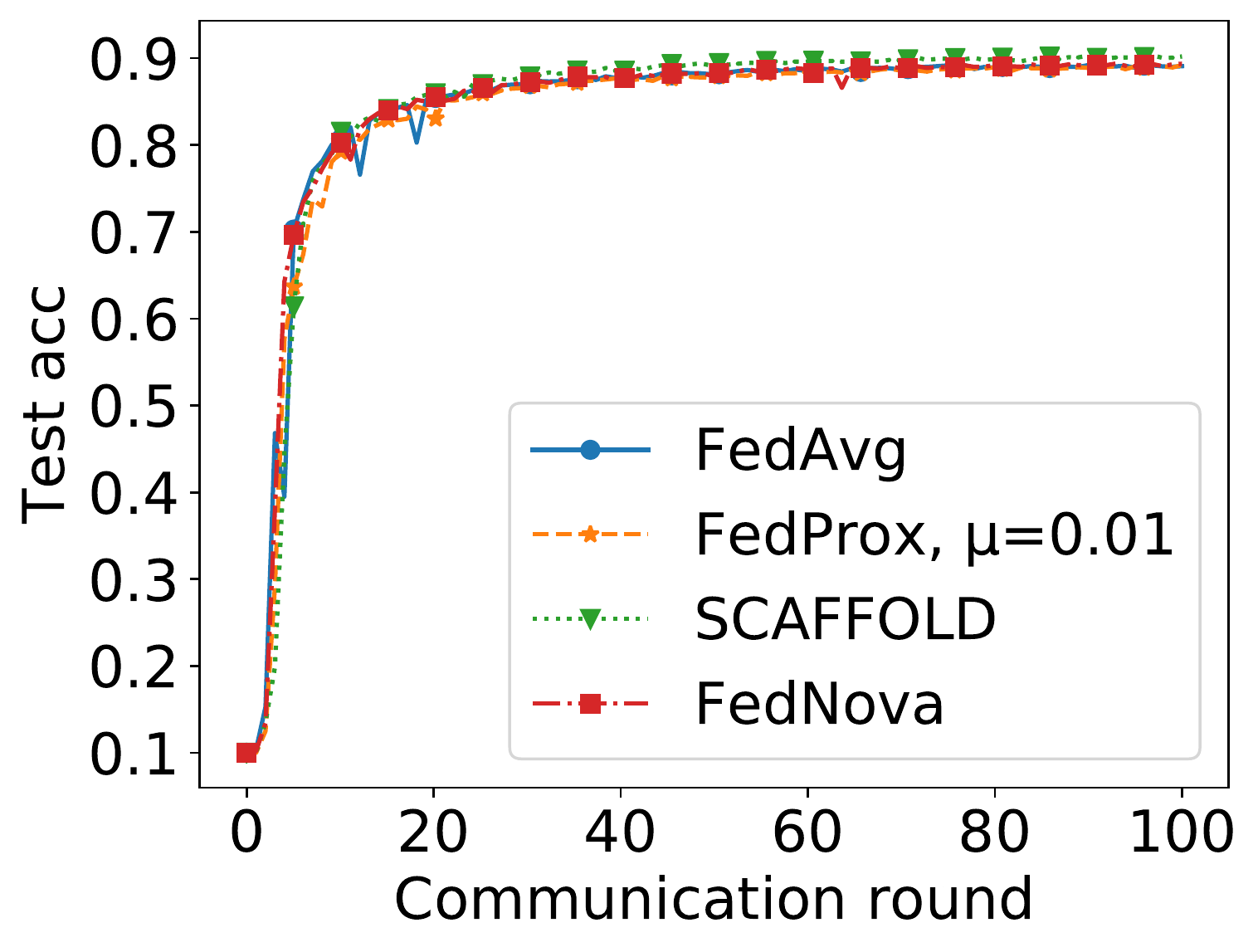}}
    \subfloat[ResNet-50 under $q \sim Dir(0.1)$ partition]{\includegraphics[width=0.33\textwidth]{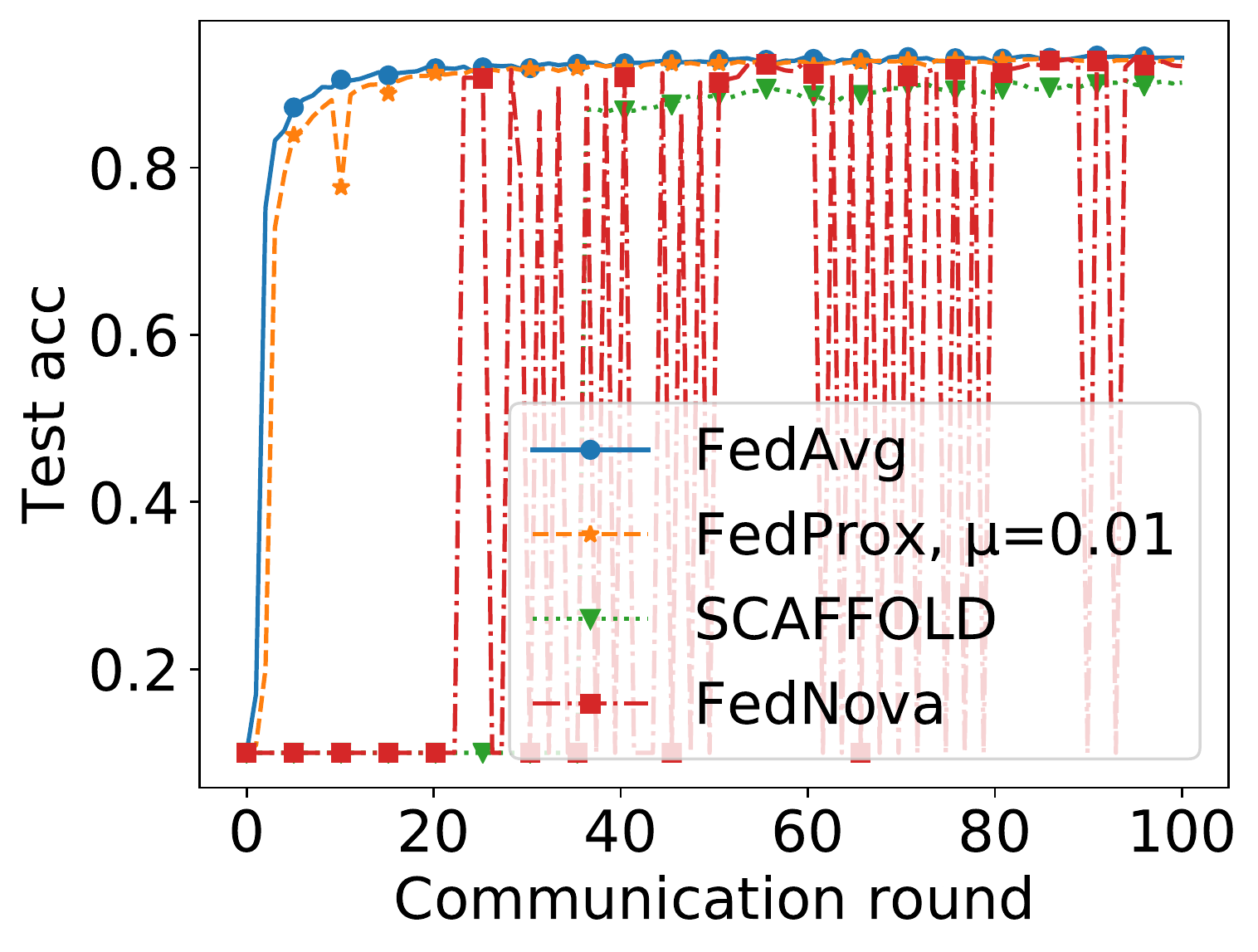}}
    \caption{The training curves of VGG-9/ResNet-50 on CIFAR-10 under different partitions.}
    \label{fig:heavy-model}
\end{figure*}


\subsection{Model Architectures}
\label{sec:model}
\noindent \fbox{\parbox{0.98\linewidth}{
\tb{Finding (11):} A simple averaging of batch normalization layers introduces instability in non-IID setting.
}}

In the previous experiments, the models we use are simple CNNs and MLPs. Here we try more complex models including VGG-9 and ResNet-50 \cite{he2016deep}. The experimental results on CIFAR-10 are shown in Figure \ref{fig:heavy-model}. Overall, while the final accuracies of using VGG-9 and ResNet-50 are usually close, training a ResNet-50 appears to more unstable. ResNet-50 uses batch normalization to standardize the inputs to a layer. A challenge in training ResNet-50 is how to aggregate the batch normalization layers. While the local batch normalization layers can handle the local distribution well, a simple averaging of these layers may not be able to catch the statistics of global distribution and introduces more instability. 

\end{document}